%% file: latex/acl_latex.tex
\pdfoutput=1

\documentclass[11pt]{article}

\usepackage[final]{acl}

\usepackage{times}
\usepackage{latexsym}

\usepackage[T1]{fontenc}

\usepackage[utf8]{inputenc}

\usepackage{microtype}

\usepackage{inconsolata}

\usepackage{graphicx}

\usepackage{algorithm}
\usepackage{algorithmic}
\usepackage{xcolor}
\usepackage{array}
\usepackage{makecell}
\usepackage{afterpage}
\usepackage[table]{xcolor}

\usepackage{comment}
\usepackage{booktabs}
\usepackage{multirow}
\usepackage{amsmath}
\usepackage{caption}

\usepackage{hyperref}
\definecolor{darkblue}{rgb}{0,0,0.6}
\hypersetup{
    colorlinks=true,
    linkcolor=darkblue,
    citecolor=darkblue,
    urlcolor=darkblue,
    pdfborder={0 0 0}
}
\usepackage[utf8]{inputenc}
\usepackage{amsfonts}
\usepackage{dsfont} 

\usepackage{enumitem}
\usepackage[normalem]{ulem} 
\graphicspath{{figure/}}
\usepackage{subcaption}

\newif\ifcolordiff
\colordifffalse   

\ifcolordiff

  \newcommand{\deleted}[1]{\textcolor{red}{\sout{#1}}}
\else

  \newcommand{\deleted}[1]{}
\fi

\setlist[itemize]{topsep=2pt,partopsep=0pt,itemsep=2pt,parsep=0pt}
\setlist[enumerate]{topsep=4pt,partopsep=0pt,itemsep=4pt,parsep=0pt}

\begin{document}

%
%

\title{Aggregation-Aware Synthetic Text Generation Against Authorship Re-Identification}


\author{
  \textbf{Qian Ma},
  \textbf{Anna Squicciarini},
  \textbf{Sarah Rajtmajer} \\
  Information Sciences and Technology\\The Pennsylvania State University \\
  \texttt{\{qfm5033, acs20, smr48\}@psu.edu}
}


\maketitle
\begin{abstract}
Online users often release multiple texts under the same identity, giving attackers an author profile that can reveal more than any single text. Existing authorship obfuscation methods optimize privacy independently for each document, leaving them blind to cross-document correlations that make aggregation dangerous. We propose Aggregation-Aware Synthetic Text Generation (AAST), a framework that addresses this gap by jointly selecting synthetic texts at the bundle level rather than optimizing each text in isolation. AAST targets attribution and verification attacks, including cross-genre settings where attacker references come from a genre not observed during generation or selection. Experiments across same-genre, cross-genre, neural, and independent non-neural stylometric attacks show that AAST lowers account-level linkability as bundle size grows, while preserving semantic quality, linguistic acceptability, and sentiment alignment.

\end{abstract}

\input{latex/section/1_introduction}

\input{latex/section/2_related_work}

\input{latex/section/3_thread_and_method}

\input{latex/section/4_experimental_setting}

\input{latex/section/5_results}

\input{latex/section/6_conclusion}

\newpage

\input{latex/section/10_limitation_ethical}

\bibliography{custom}

\newpage

\input{latex/section/appendix}

\end{document}

%% file: latex/section/1_introduction.tex
\section{Introduction}

Pseudonymous writing is central to online communities, blogs, and social media platforms~\cite{leavitt2015throwaway}. It lets users participate publicly while separating posts from legal identity, which matters for sensitive communities and for users managing identity boundaries online~\cite{triggs2021context,bao2024keep,xie2024differentially}. Yet pseudonyms hide legal names, but don't protect against linkability. Authorship analysis has long shown that texts carry stable signals through word choice, syntax, punctuation, function words, and other habits of expression~\cite{stamatatos2009survey,neal2017surveying}. Neural authorship representations make this risk more practical by learning embeddings for attribution and verification~\cite{rivera2021learning}. Recent work further shows that large language models (LLMs) can expose identity related cues from writing style~\cite{nguyen2025unraveling}, infer authors in open world settings~\cite{tan2025open}, and deanonymize pseudonymous online profiles from unstructured text at scale~\cite{lermen2026large}. 

A growing body of work studies authorship obfuscation and privacy-preserving rewriting. Recent approaches use LLMs, decoding algorithms, privacy-aware abstraction, and anonymization objectives~\cite{fisher2024styleremix,huang2024nap,dou2024reducing,huang2025zero}. A central challenge is balancing privacy with utility; minor edits may still reveal authorial style, while aggressive rewriting can degrade meaning, fluency, or downstream usefulness~\cite{yang2025robust}. 

Notably, nearly all existing authorship obfuscation methods optimize privacy independently for each document, despite the fact that real adversaries aggregate multiple posts into account-level author profiles. Even individually well-obfuscated texts can remain linkable when released together, because consistent stylistic features across a released collection leave account-level signals that no per-document method is designed to suppress. This mismatch undermines the performance of existing methods. 
In a Reddit authorship attribution setting, for example, top-eight author identification rises from 2\% with one unmodified comment to 93\% with 16 comments~\cite{bao2024keep}. Recent large scale deanonymization work further shows that LLMs can link pseudonymous profiles by extracting and matching signals across user histories~\cite{lermen2026large}. Cross-genre authorship analysis adds another concern, since one person may write in different genres, such as social media posts and news articles~\cite{rivera2021learning,ma2025crossnews}. Direct identifier masking is useful, but masked documents can remain vulnerable when attackers use retrieval and infilling~\cite{pilan2022text,charpentier2025re}. These findings motivate defenses that consider the released collection at the author level, rather than each text in isolation.
 
We propose Aggregation-Aware Synthetic Text Generation (AAST), a framework for reducing account-level authorship re-identification by weakening linkability across released \emph{bundles}, where each bundle is a collection of texts attributed to the same account. Unlike prior approaches that rewrite texts independently, AAST jointly selects synthetic texts at the bundle level, targeting the aggregated author signals used in attribution and verification attacks. This shifts the method from optimizing isolated rewrites to reducing linkability across a released account history.

We evaluate AAST on same-genre and cross-genre datasets under authorship attribution and verification attacks. Same-genre evaluation uses the benchmark Blog Authorship Corpus \cite{schler2006effects} and our constructed Reddit authorship corpus. The complete dataset and protocol details are provided in \S\ref{subsec:dataset}. Cross-genre evaluation uses the CROSSNEWS authorship benchmark linking news articles and Twitter/X posts by the same authors \cite{ma2025crossnews}. Compared with authorship obfuscation  baselines~\cite{bao2024keep,fisher2024jamdec}, AAST reduces attribution and verification risk as bundle size grows while maintaining semantic and linguistic quality, and sentiment alignment. 

\noindent Our contributions are:

\begin{itemize}
   \item We formulate and study account-level authorship privacy under aggregation, where multiple released texts from the same account form an author profile. This shifts the privacy target from isolated text rewriting to reducing linkability across a released account history.
    \item We propose AAST, an aggregation-aware synthetic text generation framework that jointly selects released texts at the bundle level to reduce linkability within a user's own account history. 
    \item We evaluate AAST across same-genre, cross-genre, neural, and independent non-neural stylometric attacks. Results show that AAST reduces account-level linkability as bundle size grows while maintaining semantic, linguistic, and sentiment utility.
\end{itemize}

\noindent All code is available at \url{https://github.com/masonmq/aast}.

%% file: latex/section/2_related_work.tex
\section{Related Work}
\label{sec:related_work}

\paragraph{Authorship analysis and author re-identification.}
Authorship analysis is commonly studied through attribution and verification~\cite{mosteller1963inference,holmes1998evolution,stamatatos2009survey}. Two common tasks are \emph{authorship attribution}, which identifies the author of a text from a candidate set, and \emph{authorship verification}, which decides whether two texts were written by the same author~\cite{keswani2016author,stamatatos2016clustering,karadzhov2017case,217499}. Recent work argues that the distinction between these tasks matters for realistic evaluation, especially when the candidate set, domain, or genre changes~\cite{bevendorff2025two,ma2025crossnews}.

Classical authorship models rely on stylometric features such as character \(n\)-grams, function words, punctuation, and compression based scores~\cite{teahan2003using,koppel2004authorship,seidman2013authorship}. Neural models later learned author representations directly from text, with LUAR using contrastive learning to produce author embeddings for verification and attribution~\cite{rivera2021learning}. PAN shared tasks provide widely used verification settings and metrics~\cite{stamatatos2022overview,bevendorff2025two}. CROSSNEWS shows that author signals do not always transfer cleanly across genres~\cite{ma2025crossnews}.

\paragraph{Authorship obfuscation and text privatization.}
Authorship obfuscation rewrites text so that the original author becomes harder to identify while the rewritten text remains useful~\cite{potthast2016author,mahmood2019girl,altakrori2022multifaceted}. It differs from ordinary paraphrasing, which may preserve author style, and from style transfer, which usually assumes a target style~\cite{mireshghallah2021style,fisher2024jamdec}. Recent work studies LLMs for privacy preserving rewriting, including text abstraction, text anonymization, authorship privacy, and privacy attack modeling~\cite{dou2024reducing,nguyen2025unraveling,staab25lmanon}. LLM based paraphrasing and style imitation can reduce the robustness of authorship verification systems~\cite{alperin2025masks}. KiP fine tunes a language model with reinforcement learning to balance privacy, semantic soundness, and fluency, and evaluates against LUAR attribution and PAN style verification attacks~\cite{bao2024keep}. JAMDEC uses small language models with constrained diverse decoding, over generation, and filtering~\cite{fisher2024jamdec}. These methods are closest to AAST because they target author style obfuscation.

\begin{figure*}[t]
    \centering
    \includegraphics[width=\linewidth]{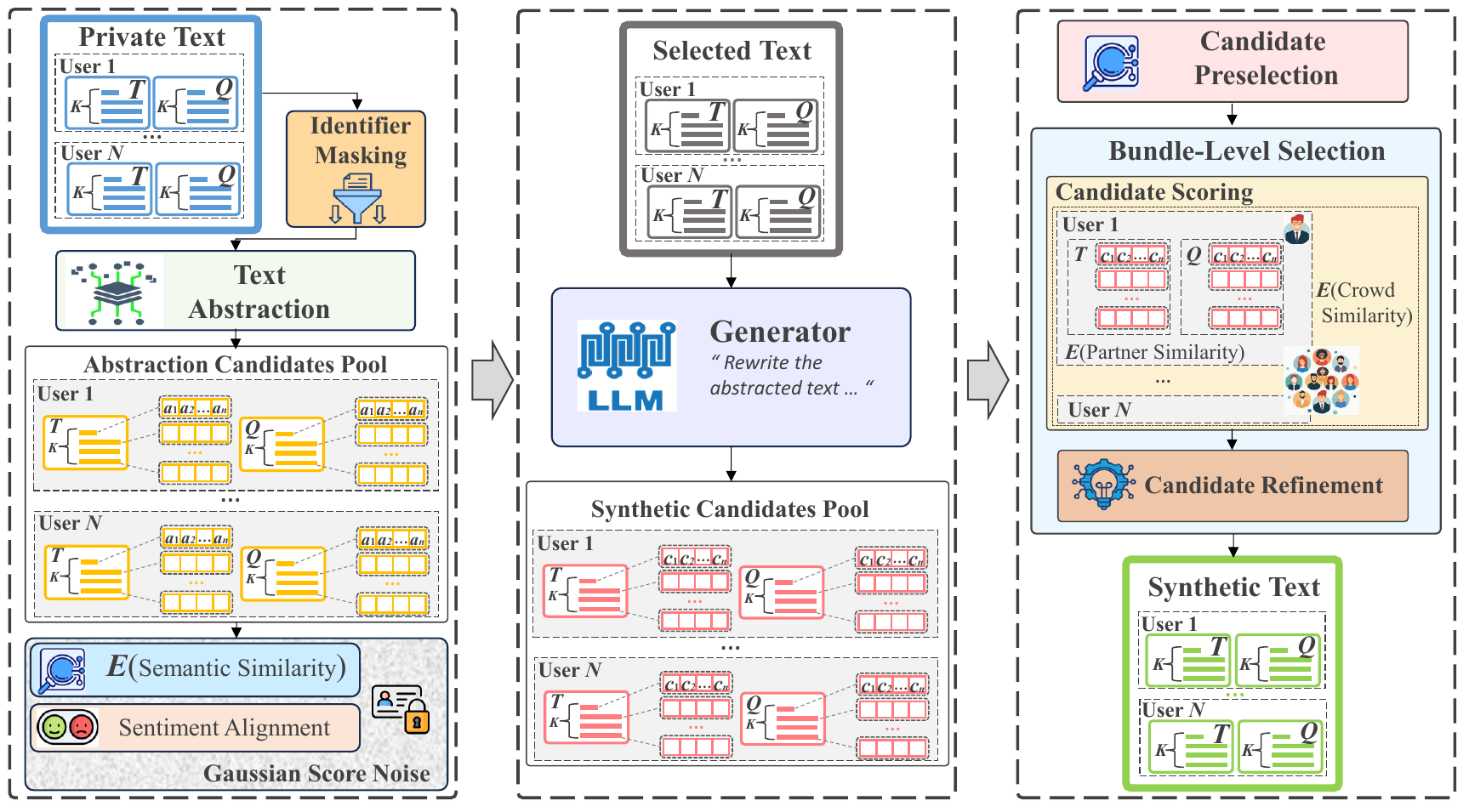}
    \caption{Overview of the AAST pipeline for aggregation-aware synthetic text generation.} 
    \label{fig:AAST}
\end{figure*}

\paragraph{Linkage risk and aggregation.}
Privacy risk arises not only from explicit identifiers, but also from linking released data with other information. Recent work systematizes dataset privacy attacks through linkage, where re-identification and attribute inference are treated as different forms of information gain~\cite{powar2023sok}. Text de-identification work also shows that masking direct identifiers is not enough, since adversaries can use background knowledge, retrieval, and infilling to recover masked information~\cite{pilan2022text,charpentier2025re}. Differential privacy has also been studied for text generation, including synthetic data from foundation model APIs~\cite{yue2023synthetic,xie2024differentially}. AAST includes a limited noisy abstraction selection analysis, but addresses aggregation by treating released account histories as the privacy target.

%% file: latex/section/3_thread_and_method.tex
\section{AAST Methodology}

\subsection{Threat Model}
\label{subsec:threat_model}

Each private text \(x\) is associated with a pseudonymous account \(u \in \mathcal{U}\), and \(\mathcal{S}\) contains synthetic texts \(\tilde{x}\) generated for the same accounts. Even after privatization, synthetic texts may preserve writing patterns that can re-identify the account author when multiple texts are observed together.

We consider an adversary that links released synthetic texts at the account level. Our primary setting is same-genre aggregation, where the released synthetic corpus contains multiple texts generated from each pseudonymous account. The adversary compares synthetic bundles and performs either an \emph{attribution attack}, ranking candidate accounts for a synthetic bundle, or a \emph{verification attack}, deciding whether two synthetic bundles come from the same account. Larger $K$ represents a stronger aggregation threat because the adversary observes a larger released account history.

We also evaluate a cross-genre reference setting, where the adversary has access to external reference texts from another genre written by candidate authors and attempts to match them against the released synthetic corpus. AAST generates and selects synthetic texts from one genre only, and does not observe cross-genre reference texts during generation or selection. This setting tests whether a released synthetic text remains linkable to other writings by the same author. 

Our goal is to lower attribution and verification success while preserving text utility.

\subsection{System Overview}

AAST reduces account-level authorship re-identification by weakening linkability at the bundle level (Figure~\ref{fig:AAST}). Following, we describe preprocessing and four primary steps: \emph{private text abstraction}; \emph{synthetic text generation}; \emph{candidate preselection}; and, \emph{bundle-level selection}. 
 
\subsection{Preprocessing}
\label{subsec:preprocessing}
Let the private corpus \(\mathcal{X}\) be grouped by pseudonymous account \(u \in \mathcal{U}\). AAST includes an optional \textsc{Identifier\_Masking} module before abstraction, which replaces structured identifiers and recognizable surface cues with placeholders such as \texttt{[USER]} and \texttt{[EMAIL]}. We enable it for cross-genre settings, where tweets often contain handles, URLs, hashtags, and other surface identifiers, but disable it for same-genre Reddit and Blog settings because masking can remove useful semantic content.

The attacker aggregates $K$ texts at a time, where $K$ denotes the number of texts from the same user grouped into each bundle. We organize each user's private texts into a \emph{query} bundle \(\mathcal{X}^{Q}_u = (x^{Q}_{u,1},\dots,x^{Q}_{u,K})\) and a \emph{target} bundle \(\mathcal{X}^{T}_u = (x^{T}_{u,1},\dots,x^{T}_{u,K})\), and the synthetic corpus keeps the same structure. We use an episode embedding function \(E(\cdot)\) to map each bundle to a single vector, matching the attacker's aggregation behavior by scoring bundles instead of isolated texts. The AAST algorithm is provided in Algorithm~\ref{alg:aast}.

\begin{algorithm}[t]
\small
\caption{AAST}
\label{alg:aast}
\textbf{Input:} private bundles \(\{X^{Q}_u, X^{T}_u\}_{u\in\mathcal{U}}\), abstraction model \(\Psi\), abstraction selector \(S\), generator \(G\), episode embedder \(E\), bundle size \(K\) \\
\textbf{Parameters:} candidates per text \(C\), weight \(\alpha\), margin \(\tau\) \\
\textbf{Output:} synthetic bundles \(\{\tilde{S}^{Q}_u, \tilde{S}^{T}_u\}_{u\in\mathcal{U}}\)

\begin{algorithmic}[1]

\FORALL{$u \in \mathcal{U}$, $b \in \{Q,T\}$, and $k \in \{1,\dots,K\}$}
    \STATE \(\mathcal{A}(x^{b}_{u,k}) \leftarrow \Psi(x^{b}_{u,k})\)
    \STATE \(a^{b}_{u,k} \leftarrow S\!\left(x^{b}_{u,k}, \mathcal{A}(x^{b}_{u,k})\right)\)
    \STATE \(\mathcal{C}(a^{b}_{u,k}) \leftarrow G(a^{b}_{u,k})\)
    \STATE \(\mathcal{C}'(a^{b}_{u,k}) \leftarrow \text{Can\_Preselection}\!\left(x^{b}_{u,k}, \mathcal{C}(a^{b}_{u,k})\right)\)
    \STATE Initialize \(\tilde{x}^{b}_{u,k}\) with one candidate from \(\mathcal{C}'(a^{b}_{u,k})\)
    
\ENDFOR

\FORALL{$u \in \mathcal{U}$}
    \STATE Compute current bundle embeddings \(v^{Q}_{u} \leftarrow E(\tilde{S}^{Q}_{u})\) and \(v^{T}_{u} \leftarrow E(\tilde{S}^{T}_{u})\)
\ENDFOR

\FORALL{$u \in \mathcal{U}$}
    \STATE Build \(\mathcal{V}_{-u}\) from the current bundle embeddings of other users
    \FORALL{$b \in \{Q,T\}$ and $k \in \{1,\dots,K\}$}
        \STATE Let \(\bar{b}\) denote the partner side of the same user
        \STATE For each \(c \in \mathcal{C}'(a^{b}_{u,k})\), form the candidate bundle \(\mathcal{S}^{b}_{u}(c)\), compute its episode embedding \(v^{b}_{u}(c)\), and evaluate the bundle objective \(\mathcal{L}(c)\)
        \STATE \(\mathcal{R} \leftarrow \{c \in \mathcal{C}'(a^{b}_{u,k}) : \mathcal{L}(c) \le \min_{c' \in \mathcal{C}'(a^{b}_{u,k})} \mathcal{L}(c') + \tau\}\)
        \STATE Compute \(s_{\mathrm{text}}(c)\) for each \(c \in \mathcal{R}\)
        \STATE \(c^{*} \leftarrow \arg\min_{c \in \mathcal{R}} s_{\mathrm{text}}(c)\)
        \STATE Update \(\tilde{x}^{b}_{u,k} \leftarrow c^{*}\) and \(v^{b}_{u} \leftarrow v^{b}_{u}(c^{*})\)
    \ENDFOR
\ENDFOR

\STATE \textbf{return} \(\{\tilde{S}^{Q}_u, \tilde{S}^{T}_u\}_{u\in\mathcal{U}}\)

\end{algorithmic}
\end{algorithm}

\subsection{Private Text Abstraction}
\label{subsec:data_abstraction}

\subsubsection{Abstraction candidate generation}Prior work shows that rephrasing sensitive disclosures into less specific terms can reduce privacy risk while preserving utility~\cite{dou2024reducing}. However, using private text for synthetic data generation can still preserve private and authorship related signals. To address this risk, we employ an abstraction model \(\Psi\) to transform each private sample \(x\in\{x^{(1)},\dots,x^{(N)}\}\) into a set of \(h\) abstracted candidates, \(\mathcal{A}(x)=\{a_1,\dots,a_h\}\), where \(N\) is the number of private samples~\cite{ma-rajtmajer-2026-private}. The abstraction candidates preserve the core meaning and sentiment of the private text while reducing direct correspondence with its surface form. Further implementation details are provided in Appendix~\ref{appendix-abstraction-alignment}.

\subsubsection{Noisy abstraction selection}
\label{subsec:dp_selection}

We use noisy abstraction selection as a randomized scoring step before candidate generation. The formal privacy statement is conditional and limited to the noisy score selection step; it does not cover candidate generation, the selected abstraction text, or the final synthetic corpus. The algorithm and proof are provided in Appendix~\ref{appendix-dp-aast} and~\ref{appendix-dp-privacy-guarantee}.

\subsection{Synthetic Text Generation}
\label{subsec:syn_generation}

For each selected abstraction \(a(x)\), we prompt an LLM to generate a pool of \(C\) synthetic candidates:
$\mathcal{C}(a(x))=\{c^{(1)},\dots,c^{(C)}\}$.
Candidate generation is performed for each slot \(x^{b}_{u,k}\), where \(k \in \{1,\dots,K\}\) and \(b \in \{Q,T\}\). The prompt asks the LLM to preserve meaning while varying writing style; details are provided in Appendix~\ref{appendix-prompt-design}.

\subsection{Candidate Preselection}
\label{subsec:candidate_preselect}
We then apply Candidate Preselection to each slot's generated pool by using the private source text for that slot as a reference. AAST encodes the source text and its generated candidates in a shared stylistic embedding space, ranks candidates by similarity to the source, and retains the half with the lowest similarity. The remaining candidates enter \emph{Bundle-Level Selection}.

\subsection{Bundle-Level Selection}
\label{subsec:bundle_level_selection}

\subsubsection{Bundle-level candidate scoring}
\label{subsubsec:candidate_scoring}
We select exactly one candidate per slot, but score candidates at the bundle level. Fix a user \(u\), a bundle side \(b \in \{Q,T\}\), and a position $k$. After Candidate Preselection, let \(\mathcal{C}'(a(x^{b}_{u,k})) \subseteq \mathcal{C}(a(x^{b}_{u,k}))\) denote the reduced candidate pool for slot $k$. During selection, all slots have temporary synthetic choices, and candidates for slot $k$ are scored by replacing the current choice at that slot. Let
$\tilde{S}^{b}_u=(\tilde{s}^{b}_{u,1},\dots,\tilde{s}^{b}_{u,K})$
denote the current temporary synthetic bundle for user \(u\) on side \(b\).

For any candidate \(c \in \mathcal{C}'(a(x^{b}_{u,k}))\), we form a candidate bundle by replacing only slot $k$:
$\tilde{S}^{b}_u(c) = (\tilde{s}^{b}_{u,1},\dots,\tilde{s}^{b}_{u,k-1}, c, \tilde{s}^{b}_{u,k+1},\dots,\tilde{s}^{b}_{u,K})$,
and compute its episode embedding \(v^{b}_u(c)=E(\tilde{S}^{b}_u(c))\).

We score \(c\) using two bundle-level terms.

\noindent\textbf{(1) Partner similarity.}
Let \(\bar{b}\) denote the other side of the same user, with \(\bar{Q}=T\) and \(\bar{T}=Q\). We measure similarity between the candidate bundle and the partner bundle from the same user:
\begingroup
\begin{equation}
\small
s_{\mathrm{partner}}(c)=\cos\!\big(v^{b}_u(c),\,v^{\bar{b}}_u\big).
\label{eq:spartner}
\end{equation}
\endgroup

\noindent\textbf{(2) Crowd similarity.}
Let \(\mathcal{V}_{-u}\) be a pool of current bundle embeddings from other users, and let \(m\) be the number of nearest other user bundles:
\begingroup
\begin{equation}
\small
s_{\mathrm{crowd}}(c)=
\mathrm{mean}\Big(\mathrm{Top}\text{-}m\big\{\cos(v^{b}_u(c),v): v\in\mathcal{V}_{-u}\big\}\Big).
\label{eq:sothers}
\end{equation}
\endgroup

\noindent These two terms define the bundle-level objective
\begingroup
\begin{equation}
\small
\mathcal{L}(c)=
\alpha\, s_{\mathrm{partner}}(c)
-
(1-\alpha)\, s_{\mathrm{crowd}}(c),
\label{eq:slot_select_stage1}
\end{equation}
\endgroup
where lower is better. This objective favors candidates that reduce similarity between the two bundles of the same user while keeping the selected bundle close to nearby bundles from other users.

\subsubsection{Candidate refinement}
\label{subsubsec:candidate_refinement}
We then refine among candidates that remain competitive under the bundle embedding objective. Let
\begingroup
\begin{equation}
\small
\mathcal{R} = \Big\{c \in \mathcal{C}'\big(a(x^{b}_{u,k})\big)
:\mathcal{L}(c) \le \min_{c' \in \mathcal{C}'(a(x^{b}_{u,k}))} \mathcal{L}(c') + \tau\Big\},
\label{eq:safe_pool}
\end{equation}
\endgroup
where \(\tau\) is a small margin that defines a near optimal candidate set under the objective in \S\ref{subsubsec:candidate_scoring}.

Within \(\mathcal{R}\), we use a character \(n\)-gram bundle similarity surrogate \(g(\cdot,\cdot)\) to compare each candidate bundle \(\tilde{S}^{b}_u(c)\) with the partner bundle \(\tilde{S}^{\bar{b}}_u\). We choose the candidate with the lowest \(g\) score, update slot \(k\), and refresh the corresponding bundle embedding. The dominant episode embedding cost of AAST is 
$
\mathcal{O}\big(|\mathcal{U}| \cdot 2K \cdot C\big).
$
Further complexity analysis is provided in Appendix~\ref{appendix_complexity_analysis}.

%% file: latex/section/4_experimental_setting.tex
\section{Experimental Methodology}
\subsection{Datasets}
\label{subsec:dataset}
\subsubsection{Same-genre}
\label{subsubsec:sd_dataset}

\paragraph{Reddit Authorship Corpus.}
We build a Reddit authorship corpus from self-disclosure posts about financial hardship and poverty. We select subreddits related to economic difficulty and use keyword filtering to identify posts about financial struggle. The subreddit and keyword lists are provided in Appendix Table~\ref{tab:reddit_sub_key}. The collected posts span January 1, 2011 to March 31, 2026 and contain 103,587 posts. This corpus supports aggregation study with up to 268 users, where each user has two bundles of up to $K$=16 posts.

\paragraph{Blog Authorship Corpus.}
The Blog Authorship Corpus~\cite{schler2006effects} is a standard authorship benchmark. It contains 681,288 posts from 19,320 bloggers collected from Blogger.com in August 2004, with about 35 posts per user on average~\cite{blog_authorship_corpus}. We exclude posts shorter than 60 words and set the largest bundle size to $K$=16, so each user can provide 16 posts for the query bundle and 16 posts for the target bundle. 

For comparability across conditions and bundle sizes, we use 250 users in same-genre Reddit and Blog experiments, ensuring each user can form two bundles up to $K$=16, or 2$K$=32 texts per user.

\subsubsection{Cross-genre}
\label{subsubsec:crossnews_dataset}

\paragraph{CROSSNEWS.} CROSSNEWS~\cite{ma2025crossnews} is a cross-genre authorship benchmark linking news articles and Twitter/X posts by the same authors. We use its manually verified gold set, which contains 500 journalists from the New York Times, the Guardian, and the Times of India, with 100 articles and 100 tweets per author.

We evaluate two cross-genre settings, Article--Tweet and Tweet--Article. In Article--Tweet, AAST generates synthetic tweets from tweet inputs, while articles are used only as attacker references and are never observed by AAST during generation or selection. Tweet--Article is constructed analogously. This tests whether synthetic text in one genre remains linkable to reference text from another genre. Article--Tweet is expected to be harder because tweets are often short and metadata heavy.

\begin{figure*}[t]
    \centering

    \begin{minipage}[c]{0.87\linewidth}
        \centering

        \begin{subfigure}[t]{0.32\linewidth}
            \centering
            \includegraphics[width=\linewidth]{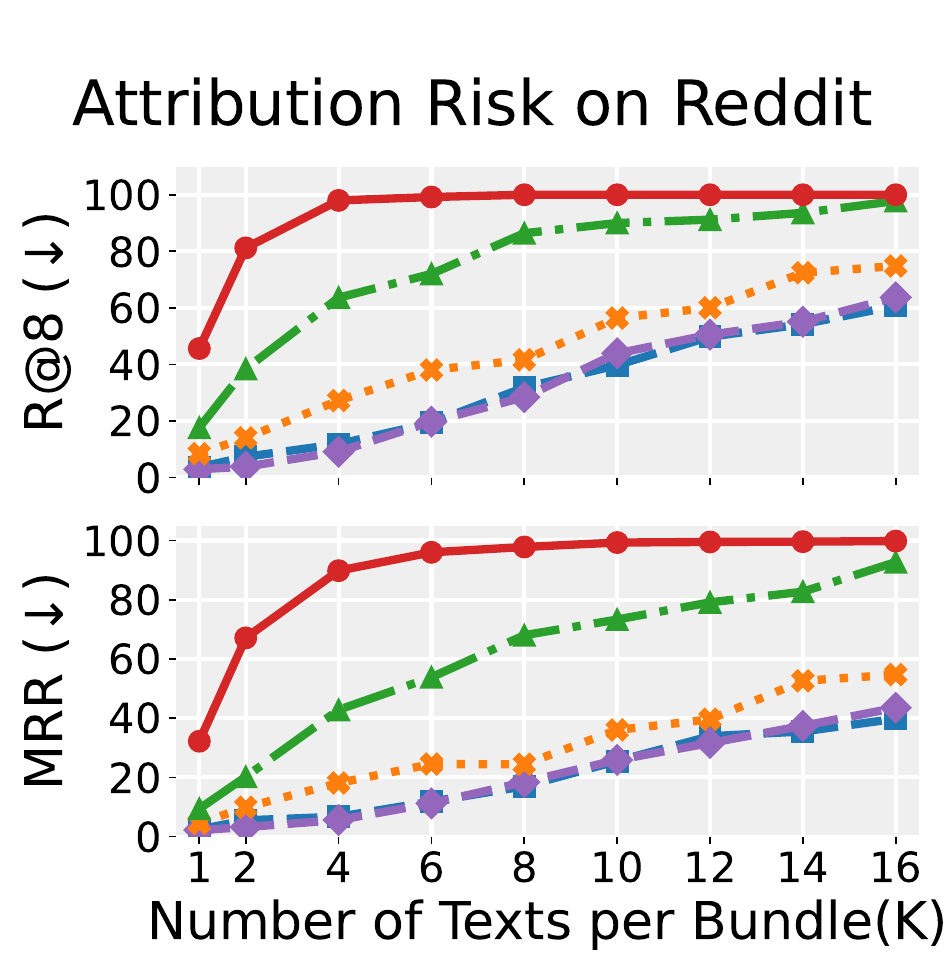}
            \caption{Attribution risk}
            \label{fig:reddit_attr}
        \end{subfigure}
        \hfill
        \begin{subfigure}[t]{0.32\linewidth}
            \centering
            \includegraphics[width=\linewidth]{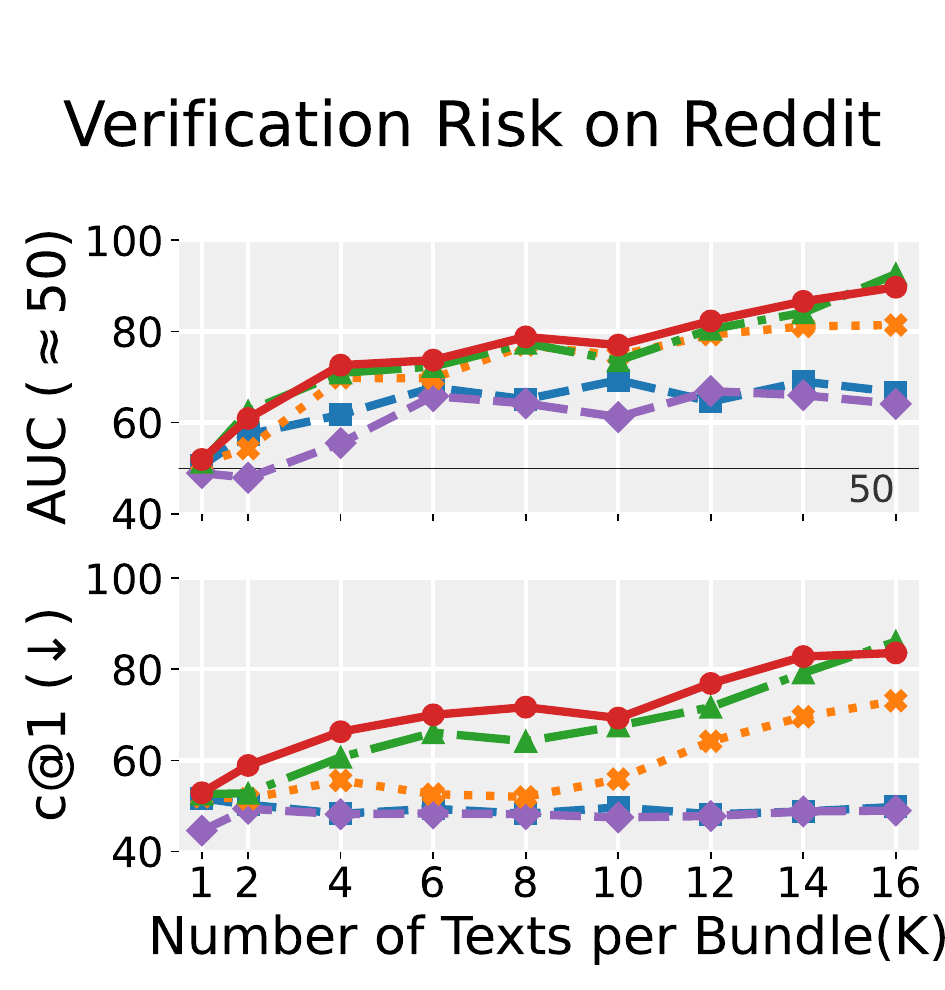}
            \caption{Verification risk}
            \label{fig:reddit_veri}
        \end{subfigure}
        \hfill
        \begin{subfigure}[t]{0.32\linewidth}
            \centering
            \includegraphics[width=\linewidth]{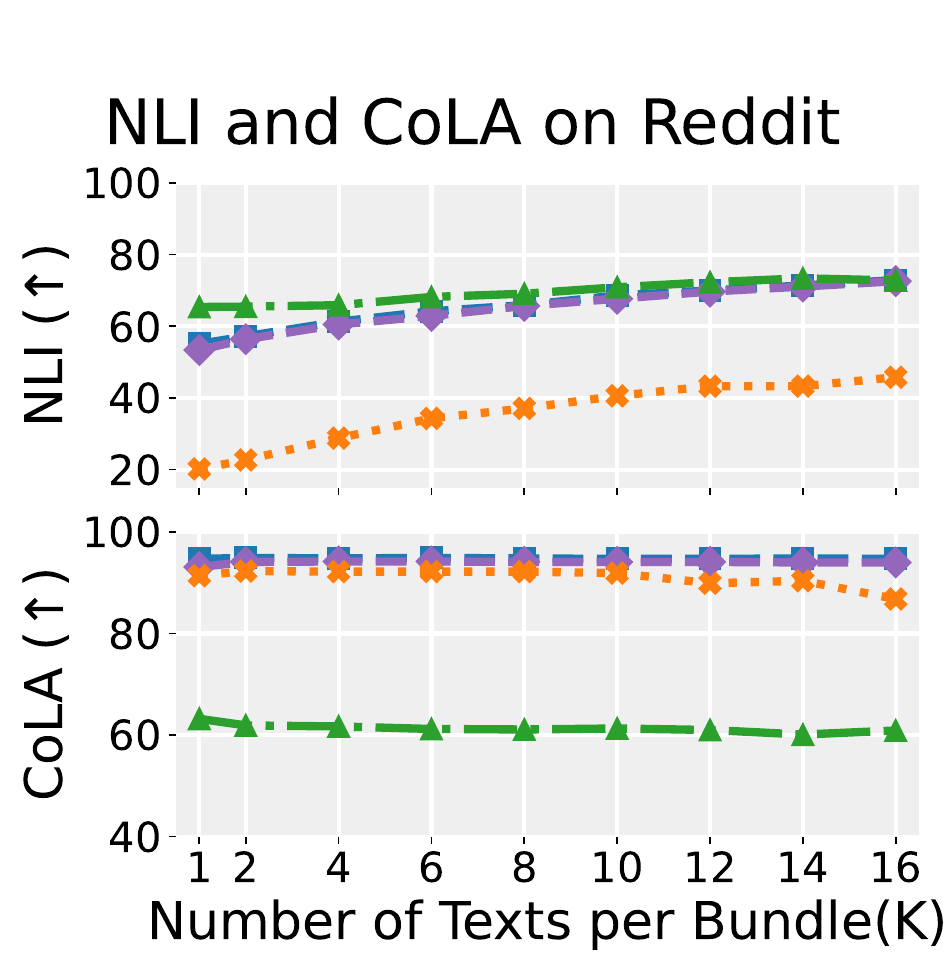}
            \caption{NLI and CoLA}
            \label{fig:reddit_utility}
        \end{subfigure}

    \end{minipage}
    \hfill
    \begin{minipage}[c]{0.11\linewidth}
        \centering
        \raisebox{1.6cm}{\includegraphics[width=\linewidth]{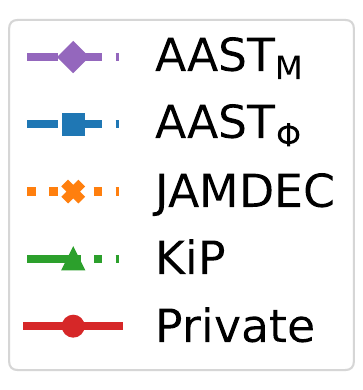}}
    \end{minipage}

    \caption{Attribution risk, verification risk, NLI, and CoLA across bundle sizes on Reddit. In (c), each synthetic output is evaluated against its corresponding private text, so the private text serves as the reference and is not plotted.}
    \label{fig:reddit_cross_k}
\end{figure*}

\subsection{Baselines}
\label{subsec:baselines}
We compare AAST with KiP~\cite{bao2024keep} and JAMDEC~\cite{fisher2024jamdec}, two recent strong authorship obfuscation baselines. KiP fine tunes an LLM with reinforcement learning to hide authorial style while preserving text quality. JAMDEC uses keyword extraction, constrained generation, over generation, and filtering to produce obfuscated rewrites with smaller language models. We also include RUPTA~\cite{yang2025robust} as an additional LLM-based rewriting baseline to test whether LLM rewriting alone is sufficient for aggregation-level authorship privacy. 

Baseline selection and settings are provided in Appendix~\ref{appendix-baseline-setup}. The RUPTA comparison in Appendix~\ref{appendix-llm-baseline-comparison} tests whether strong LLM-based rewriting alone is sufficient for aggregation-level authorship privacy.

\subsection{Metrics}
\label{subsec:metrics}

\noindent{\bf Privacy Metrics.}
We evaluate same-genre aggregation risk with authorship attribution and verification attacks. For the main same-genre attribution evaluation, we use LUAR~\cite{rivera2021learning} as a strong neural attribution attacker and report Mean Reciprocal Rank (MRR) and Recall at 8 (R@8). For verification, we use PAN22 authorship verification~\cite{stamatatos2022overview} and report Area Under the Curve (AUC) and Correctness at One (c@1). To test whether the privacy gains transfer beyond the neural models used in the main evaluation, we also include independent classical stylometric attribution and verification attacks. These stylometric attacks provide non-neural transfer evaluations for both same-genre attribution and verification.

For cross-genre evaluation, we follow CROSSNEWS and use SELMA, its authorship embedding model, to test whether synthetic text remains linkable across Article and Tweet genres. We use the strongest SELMA prompt variants for each task. For attribution, we use SELMA+TaskOnly and SELMA+LIP; for verification, we use SELMA+TaskOnly and SELMA+PromptAV. We report the CROSSNEWS metrics: top-one Accuracy, R@8, and Average Rank for attribution, and Accuracy and F1 for verification. Privacy metric details are provided in Appendix~\ref{appendix-privacy-metrics}.

\noindent{\bf Utility Metrics.}
We evaluate synthetic text quality with semantic quality, linguistic acceptability, and sentiment alignment. Semantic quality is measured by NLI, linguistic acceptability by CoLA, and sentiment alignment by comparing sentiment labels between private and synthetic texts. Utility metric details are provided in Appendix~\ref{appendix-utility-metrics}.

\subsection{Models and Implementation Settings}
\label{subsec:models}

We use bart-large-cnn for abstraction, sentiment-roberta-large-en for sentiment alignment, sentence-t5-base for semantic scoring, Mistral-Small and Phi-4 for generation. AAST$_{\mathrm{M}}$ denotes AAST with Mistral-Small as the generator, and AAST$_{\Phi}$ denotes AAST with Phi-4. Models and implementation settings are provided in Appendix~\ref{appendix-implementation-settings}.

%% file: latex/section/5_results.tex
\section{Results}
\label{sec:results}

\subsection{Privacy Results}
\label{subsec:privacy_results}

\subsubsection{Same-genre authorship attribution}

\paragraph{Reddit authorship attribution.}
\label{subsubsec:reddit_attribution}

Using the main LUAR attribution attacker, Figure~\ref{fig:reddit_attr} shows that aggregation sharply increases attribution risk on Reddit (details reported in Appendix Table~\ref{tab:attribution-reddit-blog}). The private setting rises from moderate linkability at $K$=1 to near perfect linkability at larger $K$. By $K$=8, private text already reaches 100 R@8, showing that only a small number of texts from the same account can make attribution highly reliable. In result tables, the best value is shown in \textbf{bold}, and the second best value is \uline{underlined}.

AAST consistently reduces this attribution risk across all bundle sizes. At small $K$, AAST$_{\mathrm{M}}$ gives the lowest risk (e.g., 2.2 MRR and 2.9 R@8 at $K$=1). As $K$ grows, AAST$_{\Phi}$ becomes stronger on most larger bundle sizes. At $K$=16, AAST$_{\Phi}$ obtains the lowest risk among all methods, with 39.8 MRR compared with 54.7 for JAMDEC and 92.7 for KiP. AAST$_{\mathrm{M}}$ also remains below both rewriting baselines. These results show that AAST reduces authorship attribution under aggregation on Reddit, although the attack still becomes harder to suppress as $K$ increases.

\paragraph{Blog authorship attribution.}
\label{subsubsec:blog_attribution}

AAST shows a clear advantage on Blog. As shown in Appendix Figure~\ref{fig:blog_attr} and Table~\ref{tab:attribution-reddit-blog}, AAST gives the lowest attribution risk across all bundle sizes. AAST$_{\Phi}$ is the strongest setting throughout, at $K$=16, AAST$_{\Phi}$ keeps R@8 at 61.4, compared with 83.6 for JAMDEC and 94.0 for KiP. AAST$_{\mathrm{M}}$ is consistently the next strongest method. The gap is most visible at larger $K$, where AAST remains much harder to attribute under stronger aggregation pressure.

\paragraph{Stylometric attribution.}
\label{subsubsec:stylo_attribution}
The LUAR attribution results above provide an attacker-aware neural evaluation because AAST uses an authorship episode encoder during bundle-level selection. To test whether the gains transfer beyond LUAR, Appendix Table~\ref{tab:stylometric-attribution} evaluates the same-genre setting with an independent classical n-gram stylometric attribution attacker. AAST remains effective under this unseen attacker. At $K$=16, the two AAST variants obtain the lowest and second-lowest MRR and R@8 on both Reddit and Blog. These results provide direct transfer evidence that AAST's attribution gains are not confined to the LUAR-based neural evaluator, but persist under an independent non-neural stylometric attacker. Full results are in Appendix~\ref{appendix-same-genre-stylo-attri}.

\subsubsection{Same-genre authorship verification}

\paragraph{Reddit authorship verification.}
\label{subsubsec:reddit_verification}

In Figure~\ref{fig:reddit_veri}, the private curve shows that aggregation also makes authorship verification much easier. AAST keeps verification risk much lower across bundle sizes. AAST$_{\mathrm{M}}$ is usually the strongest setting, especially at the largest bundle size, where it keeps AUC at 64.1, while JAMDEC and KiP rise to 81.4 and 92.6. AAST$_{\Phi}$ also improves over both rewriting baselines for most $K$. These results show that AAST reduces verification risk under aggregation, with the clearest gains on AUC, even when larger bundles make the attack stronger. The corresponding values are reported in Appendix Table~\ref{tab:verification-reddit-blog}.

\begin{figure}[t]
    \centering
\includegraphics[width=\linewidth]{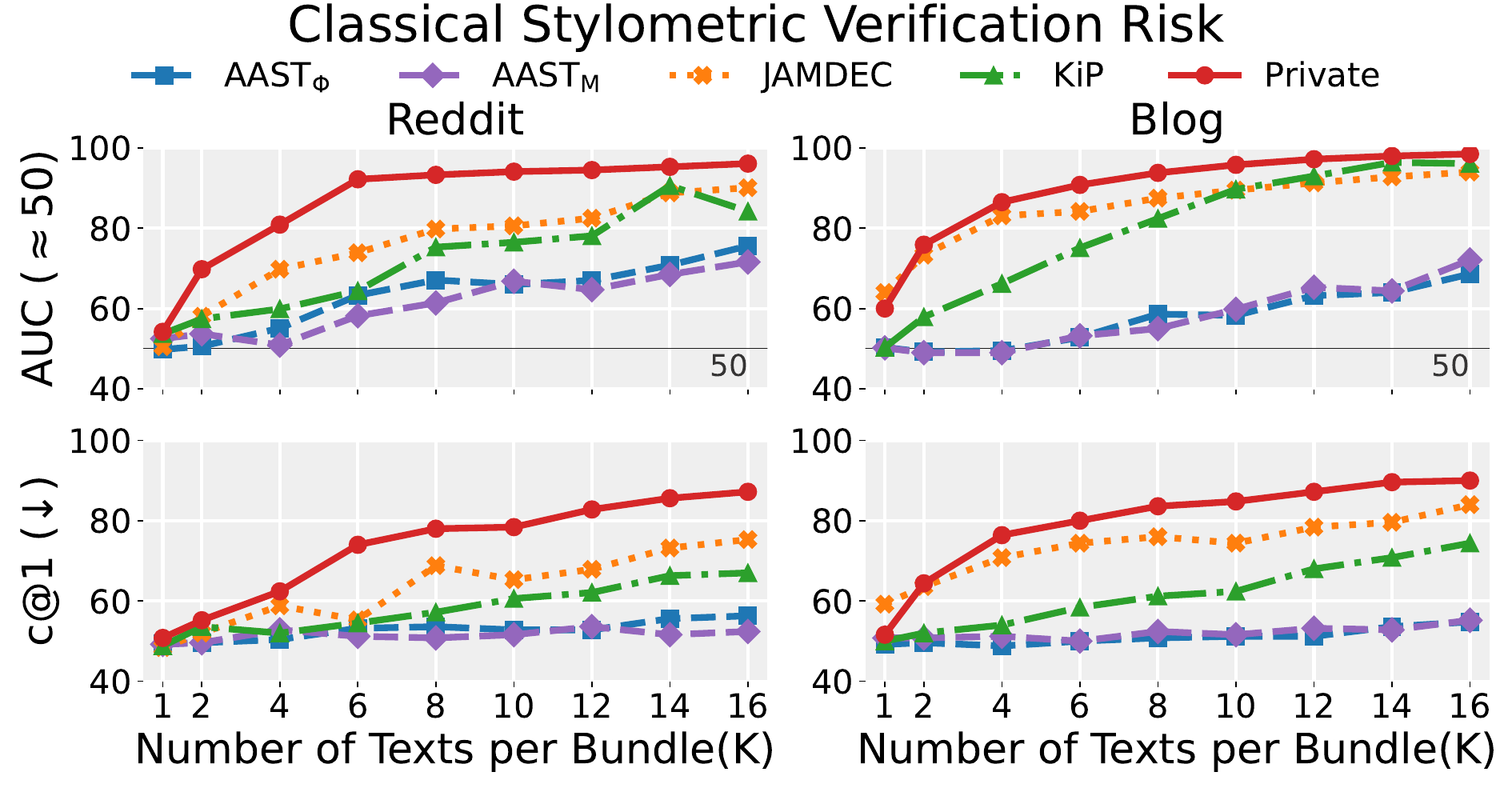}
    \caption{Verification results under a stylometric transfer attack on Reddit and Blog.} 
    \label{fig:stylo_verifi}
\end{figure}

\paragraph{Blog authorship verification.}
\label{subsubsec:blog_verification}

Appendix Figure~\ref{fig:blog_veri} and Table~\ref{tab:verification-reddit-blog} show that AAST remains stable against verification risk on Blog. AAST$_{\mathrm{M}}$ gives the strongest protection at larger bundle sizes. At $K$=16, it keeps AUC at 68.6 and c@1 at 50.3, while JAMDEC reaches 73.3 AUC and 61.4 c@1, and KiP reaches 85.3 AUC and 68.1 c@1. The Blog verification results show that AAST weakens verification of whether two texts come from the same author under aggregation. 

\paragraph{Stylometric verification.}
\label{subsubsec:stylometric_verification}

We also evaluate AAST with the classical stylometric transfer attack. Figure~\ref{fig:stylo_verifi} and Appendix Table~\ref{tab:stylometric-verifi} show the same trend. At $K$=16, AAST$_{\mathrm{M}}$ gives lower AUC than JAMDEC and KiP on Reddit, 71.6 versus 90.1 and 84.2, respectively. On Blog, AAST$_{\Phi}$ gives the lowest AUC 68.6 and lowest c@1 54.8. Together with the stylometric attribution results in Appendix~\ref{appendix-same-genre-stylo-attri}, these results provide transfer evidence that AAST's privacy gains are not limited to neural authorship models, but persist under independent classical stylometric attacks. Full verification results are in Appendix~\ref{appendix-same-genre-stylo-verifi}.

AAST$_{\Phi}$ and AAST$_{\mathrm{M}}$ show complementary strengths. AAST$_{\Phi}$ is often stronger on attribution at larger \(K\), while AAST$_{\mathrm{M}}$ is usually more stable for verification and cross-genre settings. The stylometric attribution and verification transfer evaluations follow the same pattern. Since the selection objective is the same, these differences likely come from the candidate distributions produced by the two generators, including differences in semantic preservation, style shift, and candidate diversity. We leave a controlled analysis of generator behavior for future work.

\subsubsection{Cross-genre authorship attribution}
\label{subsubsec:crossgenre_attribution}

Table~\ref{tab:crossgenre-attribution} shows that AAST reduces attribution risk when the attacker links across genres. Across both genre pairs and both SELMA prompt settings, AAST$_{\mathrm{M}}$ gives the lowest top-one Accuracy and R@8. In Tweet--Article, AAST$_{\mathrm{M}}$ gives the lowest Accuracy and R@8 under both SELMA+TaskOnly and SELMA+LIP. Under SELMA+LIP, it reduces R@8 to 5.0 and pushes the true author to an average rank of 192. AAST$_{\Phi}$ is usually the second strongest method.

The same pattern appears in Article--Tweet. AAST$_{\mathrm{M}}$ achieves the lowest Accuracy and R@8 under both attacks, and gives the highest average rank of the true author. These results show that AAST's protection is not limited to same-genre aggregation, it also weakens author signals that transfer across genres. This comes from identifier masking removing explicit markers, candidate preselection reducing stylistic cues, and bundle-level selection reducing consistent account-level signals.

\subsubsection{Cross-genre authorship verification}
\label{subsubsec:crossgenre_verification}
Appendix Table~\ref{tab:crossgenre-verification} shows that cross-genre verification follows the attribution results. AAST lowers the verifier's ability to link synthetic and reference texts across both genre datasets, and AAST$_{\mathrm{M}}$ usually gives the lowest verification risk. Complete results are provided in Appendix~\ref{appendix-cross-genre-verification}. 


\noindent \textbf{Note on Article--Tweet results.} Article--Tweet is the harder cross-genre setting because tweet inputs often contain little text beyond handles, URLs, hashtags, and brief reactions. After these surface markers are normalized, the remaining text can be short or generic, which gives bundle-level selection less semantic and stylistic variation to use. Even in this harder setting, AAST$_{\mathrm{M}}$ gives the strongest cross-genre privacy results. Further discussion is provided in Appendix~\ref{appendix-discuss}.

\begin{table}
    \centering  
    \fontsize{6.5}{7.5}\selectfont
    \begin{tabular}{
        >{\centering\arraybackslash}p{1.5cm}
        >{\centering\arraybackslash}p{0.8cm}
        >{\centering\arraybackslash}p{0.8cm}
        |
        >{\centering\arraybackslash}p{0.5cm}
        >{\centering\arraybackslash}p{0.5cm}
        >{\centering\arraybackslash}p{0.8cm}
    }
        \toprule
        \textbf{Genre Pair} & \textbf{Attack} & \textbf{Method}
        & \textbf{Acc.}($\downarrow$)
        & \textbf{R@8}($\downarrow$)
        & \textbf{Avg.Rank}($\uparrow$) \\
        \midrule

         \multirow{8}{*}{\shortstack{Tweet--Article}}
        & \multirow{4}{*}{\shortstack{SELMA +\\TaskOnly}}
        & AAST$_{\mathrm{M}}$ & \textbf{1.2} & \textbf{7.5} & \uline{173} \\
        & & AAST$_{\Phi}$ & \uline{1.9} & \uline{7.6} & \textbf{175} \\
        & & KiP        & 2.7 & 9.8 & 157 \\
        & & JAMDEC     & 2.2 & 11.3 & 121 \\
        \cmidrule{2-6}

        & \multirow{4}{*}{\shortstack{SELMA +\\LIP}}
        & AAST$_{\mathrm{M}}$ & \textbf{0.9} & \textbf{5.0} & \textbf{192} \\
        & & AAST$_{\Phi}$ & \uline{1.6} & \uline{6.6} & \uline{185} \\
        & & KiP        & 2.2 & 8.4 & 165 \\
        & & JAMDEC     & 2.2 & 10.8 & 125 \\       
        \midrule
        \midrule

        \multirow{8}{*}{\shortstack{Article--Tweet}}
        & \multirow{4}{*}{\shortstack{SELMA +\\TaskOnly}}
        & AAST$_{\mathrm{M}}$ & \textbf{1.3} & \textbf{5.7} & \textbf{222} \\
        & & AAST$_{\Phi}$ & 2.2 & \uline{7.5} & \uline{210} \\
        & & KiP        & 5.1 & 14.7 & 160 \\
        & & JAMDEC     & \uline{1.6} & 8.1 & 167 \\
        \cmidrule{2-6}

        & \multirow{4}{*}{\shortstack{SELMA +\\LIP}}
        & AAST$_{\mathrm{M}}$ & \textbf{1.2} & \textbf{5.4} & \textbf{224} \\
        & & AAST$_{\Phi}$ & 1.9 & \uline{6.9} & \uline{213} \\
        & & KiP        & 4.0 & 12.6 & 166 \\
        & & JAMDEC     & \uline{1.5} & 7.3 & 169 \\

        \bottomrule
    \end{tabular}
    \caption{Cross-genre authorship attribution risk results.}
    \label{tab:crossgenre-attribution}
\end{table}

\subsection{Utility Results}
\label{subsec:utility_results}

\paragraph{Reddit.}
\label{subsubsec:utility_reddit}

On Reddit, AAST maintains strong semantic quality (NLI) and linguistic quality (CoLA) while reducing account-level authorship linkability (See Figure~\ref{fig:reddit_utility} and Appendix Table~\ref{tab:utility-reddit-blog}). AAST$_{\Phi}$ gives the best CoLA score at every $K$, and its NLI improves as $K$ grows, reaching 72.9 at $K$=16. AAST$_{\mathrm{M}}$ follows closely, with 72.6 NLI at $K$=16. KiP has strong NLI, but its privacy risks remain high. JAMDEC gives lower attribution and verification risk than KiP in some settings, but its NLI is much lower than AAST.

AAST gives a stronger privacy and utility balance, preserving both content and linguistic quality while reducing linkability under aggregation. We note that CoLA scores for AAST are consistently high across bundle sizes, so CoLA mainly confirms linguistic acceptability for our method. In contrast, lower CoLA for KiP reflects the token fragmentation artifacts observed in some outputs (Appendix Table~\ref{tab:qualitative-2}). We still retain CoLA for comparability with baselines.

AAST also conditions generation on sentiment, so we measure whether each synthetic text preserves the sentiment of its corresponding private text. We report this metric only for AAST, since KiP and JAMDEC do not include sentiment control. Appendix Figure~\ref{fig:utili_senti} and Table~\ref{tab:aast-sentiment-alignment} show that AAST maintains stable sentiment alignment on Reddit, staying between 83\% and 85.5\%. This suggests that AAST largely preserves the sentiment of the original posts while reducing authorship linkability.

\paragraph{Blog and cross-genre utility.}
The Blog utility results follow the same pattern, with AAST preserving strong linguistic quality and stable sentiment alignment while maintaining lower authorship linkability (see Figure~\ref{fig:blog_utility}). Blog utility results are provided in Appendix~\ref{appendix-blog-utility}. 

Cross-genre utility results are provided in Appendix~\ref{appendix-cross-genre-utility}, with detailed values in Table~\ref{tab:crossgenre-utility}. In both Tweet--Article and Article--Tweet, AAST$_{\mathrm{M}}$ achieves the highest CoLA score and AAST$_{\Phi}$ is consistently second best, indicating stronger linguistic quality than the baselines.

\subsection{LLM Rewriting Baseline Comparison}
\label{subsec:llm-baseline-results}

Appendix Table~\ref{tab:rupta-comparison} compares AAST with RUPTA, a recent LLM-based text anonymization baseline, on Reddit and Blog. The results show that strong LLM-based rewriting alone is not sufficient for aggregation-level authorship privacy. Although some RUPTA variants obtain higher NLI, they also show much higher attribution and verification risk as $K$ grows. This suggests that staying closer to the private text can preserve semantic overlap, but can also preserve authorial cues. In contrast, AAST uses bundle-level selection to reduce account-level linkability while maintaining reasonable NLI and strong CoLA, giving a stronger privacy--utility balance under aggregation. Full results are in Appendix~\ref{appendix-llm-baseline-comparison}.

\subsection{Ablation Analysis}

\paragraph{Ablation design.}
We use ablations to clarify which modules drive AAST's privacy and utility tradeoff. The component ablation studies Identifier Masking (\S\ref{subsec:preprocessing}), Candidate Preselection (\S\ref{subsec:candidate_preselect}), Candidate Refinement (\S\ref{subsubsec:candidate_refinement}), and Bundle-Level Selection (\S\ref{subsec:bundle_level_selection}). We further add NoBundleSel, a controlled baseline that keeps the same settings as AAST but removes Bundle-Level Candidate Scoring and Candidate Refinement. This comparison isolates the role of bundle-level selection under our same-genre setting.

\paragraph{Component effects.}
Appendix Table~\ref{tab:aast-ablation} shows that different modules address different risks. Candidate Preselection, Refinement, and Bundle-Level Selection are always enabled in all experiments, while Identifier Masking is enabled only for cross-genre settings. Identifier Masking improves privacy but lowers NLI, so we use it only when surface markers are frequent. Candidate Preselection targets direct private--synthetic carryover, disabling it gives Bundle-Level Selection a larger candidate pool and can lower synthetic bundle linkability. Bundle-Level Selection is the main module for bundle-level privacy at larger $K$. Appendix~\ref{appendix-ablation-components} provides more details on component effects.

\paragraph{Effect of removing bundle-level selection.}

Appendix Figure~\ref{fig:abla_no_bundle} compares AAST$_{\Phi}$ with NoBundleSel$_{\Phi}$ on Reddit and Blog. NoBundleSel$_{\Phi}$ keeps the same generator, prompt, abstraction model, and candidate pool as AAST$_{\Phi}$, but removes Bundle-Level Selection. NoBundleSel$_{\Phi}$ shows higher attribution and verification risk as $K$ grows, while NLI and CoLA remain close to AAST$_{\Phi}$. This shows that bundle-level selection is the main source of the privacy gain, and that this gain does not come from a meaningful loss in semantic or linguistic utility. Full results are provided in Appendix~\ref{appendix-ablation-bundlesel}.

\subsection{Effect of Noisy Selection}

We also study a noisy abstraction selection variant, where Gaussian noise is added to abstraction selection scores before choosing the abstraction used for generation. As shown in Appendix~\ref{appendix-dp-evaluation} and Table~\ref{tab:dp-epsilon-results}, stronger noise generally reduces attribution and verification risk at larger bundle sizes, while introducing a small utility tradeoff.

\subsection{Qualitative Analysis}

We qualitatively inspect Reddit $K$=2 examples, focusing on whether each method preserves main events and core meaning. AAST keeps the central events and core meaning, such as the disease name, need for medical resources, loss of faith, and mental health concern, while changing the surface form. KiP preserves many local details, but often stays close to private wording and structure, and sometimes produces fragmented words or control artifacts. JAMDEC moves farther from the private text, but often loses core information through repetitive outputs and semantic drift. Appendix~\ref{appendix-qualitative-evaluation} provides complete examples and summary of observed strengths and weaknesses.

\subsection{Generation Cost}

\paragraph{Computational efficiency analysis.}
\label{subsubsec:computational-eval}

AAST is the most efficient method across all bundle sizes (Appendix Figure~\ref{fig:time_cost}). KiP requires about $1.9\times$ longer runtime on average, while JAMDEC requires about $18.5\times$ longer runtime. Appendix~\ref{appendix-computational-results} reports experimental settings, relative runtime, and output length analysis for interpreting the comparison.

\paragraph{Token cost analysis.}
Token usage grows almost linearly with bundle size because larger $K$ requires generating more synthetic texts. At $K$=16, 250 users \(\times\) 2\ bundles per user \(\times\) 16 texts per bundle requires generation for 8,000 texts. 
Token usage results are provided in Appendix~\ref{appendix-token-cost}.

%% file: latex/section/6_conclusion.tex
\section{Conclusion}

We have introduced AAST, an aggregation-aware synthetic text generation framework for reducing account-level authorship linkability. AAST shifts the privacy target from isolated text rewriting to bundle-level selection over a released account history. Across same-genre, cross-genre, neural, and independent non-neural stylometric attacks, AAST reduces account-level linkability as bundle size grows while maintaining semantic quality, linguistic acceptability, and sentiment alignment.

%% file: latex/section/10_limitation_ethical.tex
\section*{Limitations}

AAST reduces account-level authorship linkability, but it is not a complete anonymization mechanism. Because AAST preserves meaning, sentiment, and linguistic quality, some private topics, events, and preferences can remain in the synthetic text. Our utility evaluation also relies on automatic metrics and qualitative examples. While NLI, CoLA and sentiment alignment capture important dimensions of text quality, they may not fully reflect human judgments of fluency or downstream utility, which remain important directions for future work. 

AAST is less effective when released texts are very short and metadata heavy, as in some Article--Tweet cases. AAST still gives the strongest Article--Tweet privacy results, but tweet-like inputs often contain limited free text beyond handles, URLs, hashtags, and brief reactions (see Appendix Table~\ref{tab:masked-tweet-examples}). This gives bundle-level selection less semantic and stylistic material to use. Future work should design generation and selection methods for tweet-like text.

AAST uses a LUAR-based episode encoder during bundle-level selection, so same-genre LUAR attribution is not fully independent of the selector. To test transfer beyond this selector, we also evaluate independent stylometric attribution and verification, PAN22 verification, and SELMA cross-genre attacks, none of which are used during selection. These results suggest that AAST's gains transfer beyond the LUAR-based evaluator. Future work should evaluate stronger adaptive attackers under the same aggregation protocol.

\section*{Ethical Considerations}

Our work uses Reddit posts that were publicly accessible at the time of collection and gathered through Reddit's official API. Because the Reddit corpus concerns financial hardship and poverty, we treat it as sensitive even though it is public. We follow a data minimization principle: we collect only the text and metadata needed for authorship privacy evaluation, do not collect real names or external profile information, and do not combine Reddit content with outside sources to identify individuals.

We do not contact users, infer real world identities, or report results about individual accounts. Examples in this paper are used only when needed to illustrate method behavior. We avoid usernames, direct links, and unnecessary verbatim quotation in examples, and we mask identifying surface details when possible.

Access to the collected dataset will be limited to verified researchers upon request and under conditions that restrict use to privacy and synthetic text research. Researchers must agree not to attempt re-identification, contact users, redistribute the data, or link the posts with external sources. We will also honor reasonable removal requests when a post or author can be identified in the released research data.

AAST's bundle-level objective uses crowd similarity to reduce account-level linkability. While this improves privacy against attribution and verification attacks, it may also increase the chance that a synthetic bundle is mistakenly associated with nearby users. We do not intend AAST to imitate or impersonate specific individuals. Future deployments should monitor false attribution concentration, avoid outputs that closely mimic identifiable users, and use additional safeguards when the surrounding user population is small or sensitive.

AAST is intended as a defensive risk reduction tool, not as a guarantee of anonymity. As discussed in the Limitations section, synthetic outputs may still support some forms of linkage. Deployments should therefore avoid presenting AAST as complete anonymization and should use it only with appropriate privacy review and safeguards.

With respect to dual use, AAST is designed as a defensive tool to help users protect their privacy online. However, methods that reduce authorship linkability could also be misused to obscure responsibility for harmful or policy violating content, or to create misleading impressions about authorship. We recommend that AAST be used only for legitimate privacy preservation, privacy research, and controlled synthetic text release, with safeguards against impersonation, deceptive use, and attempts to evade accountability.

\section*{Acknowledgments}
This work was partially supported by the National Science Foundation  under Award No.~2247723.

\newpage

%% file: latex/section/appendix.tex
\clearpage
\appendix

\begin{figure*}[t]
    \centering

    \begin{minipage}[c]{0.87\linewidth}
        \centering

        \begin{subfigure}[t]{0.32\linewidth}
            \centering
            \includegraphics[width=\linewidth]{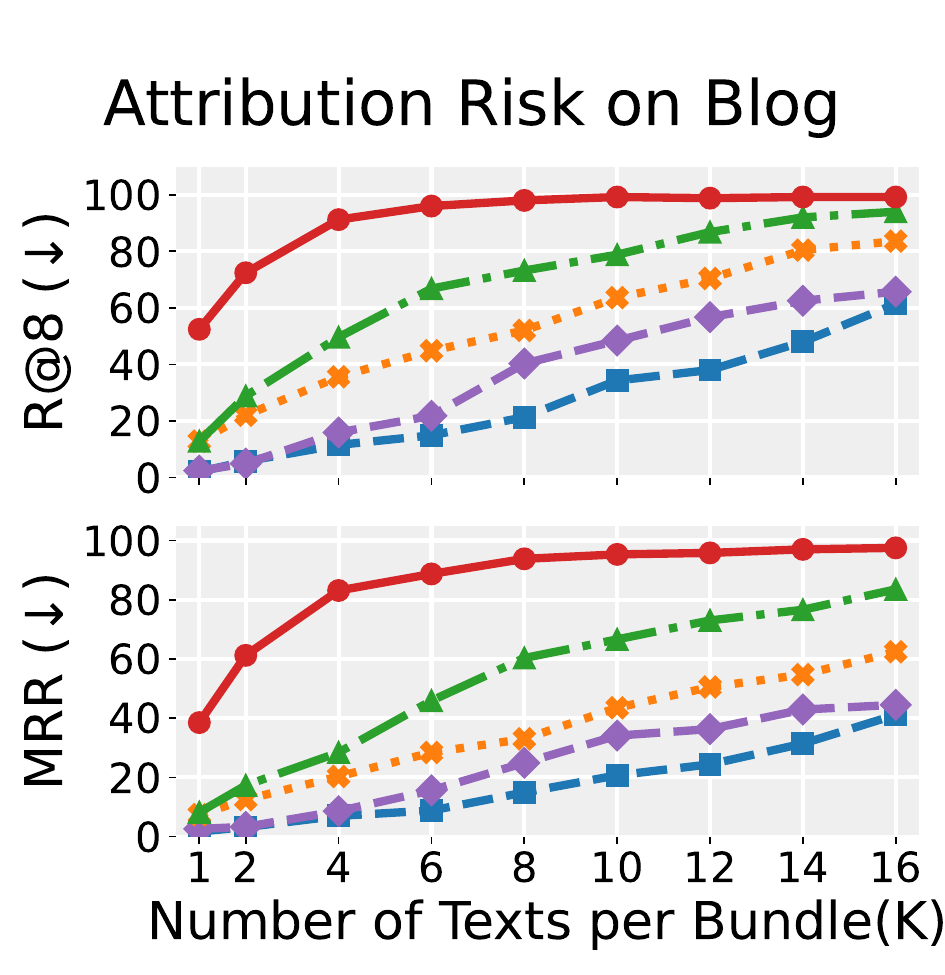}
            \caption{Attribution risk}
            \label{fig:blog_attr}
        \end{subfigure}
        \hfill
        \begin{subfigure}[t]{0.32\linewidth}
            \centering
            \includegraphics[width=\linewidth]{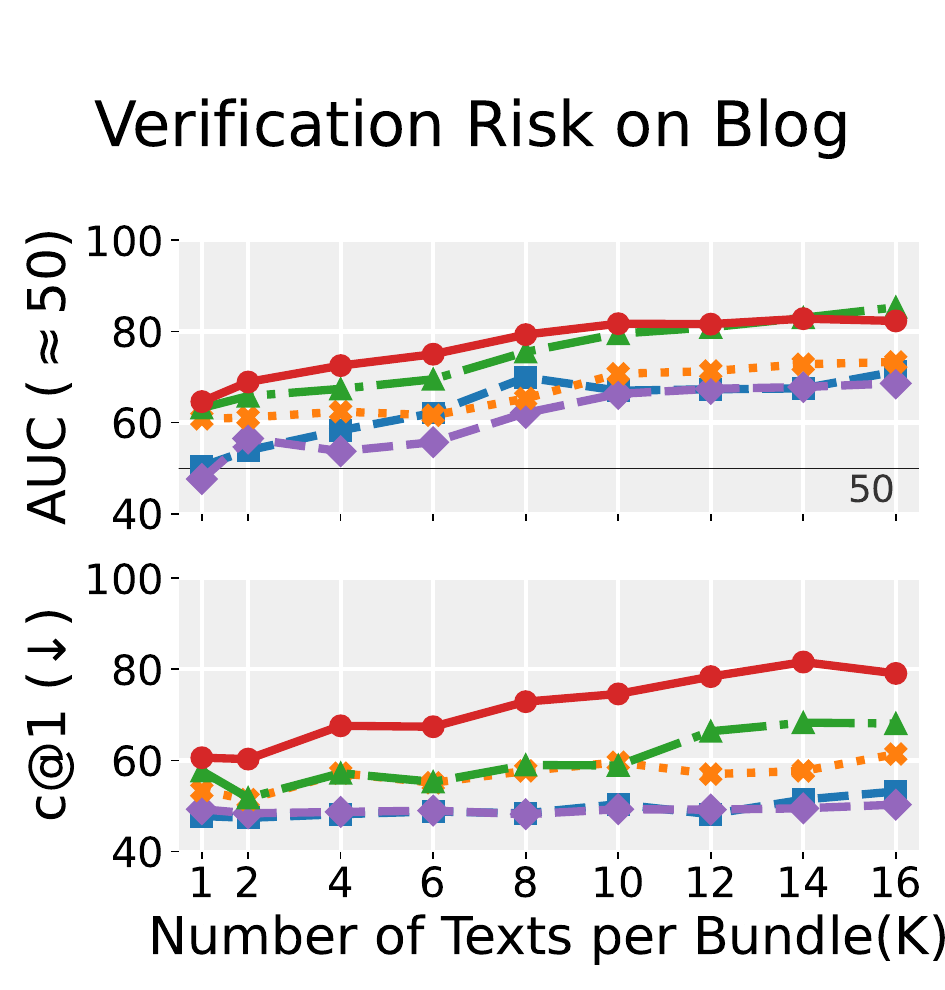}
            \caption{Verification risk}
            \label{fig:blog_veri}
        \end{subfigure}
        \hfill
        \begin{subfigure}[t]{0.32\linewidth}
            \centering
            \includegraphics[width=\linewidth]{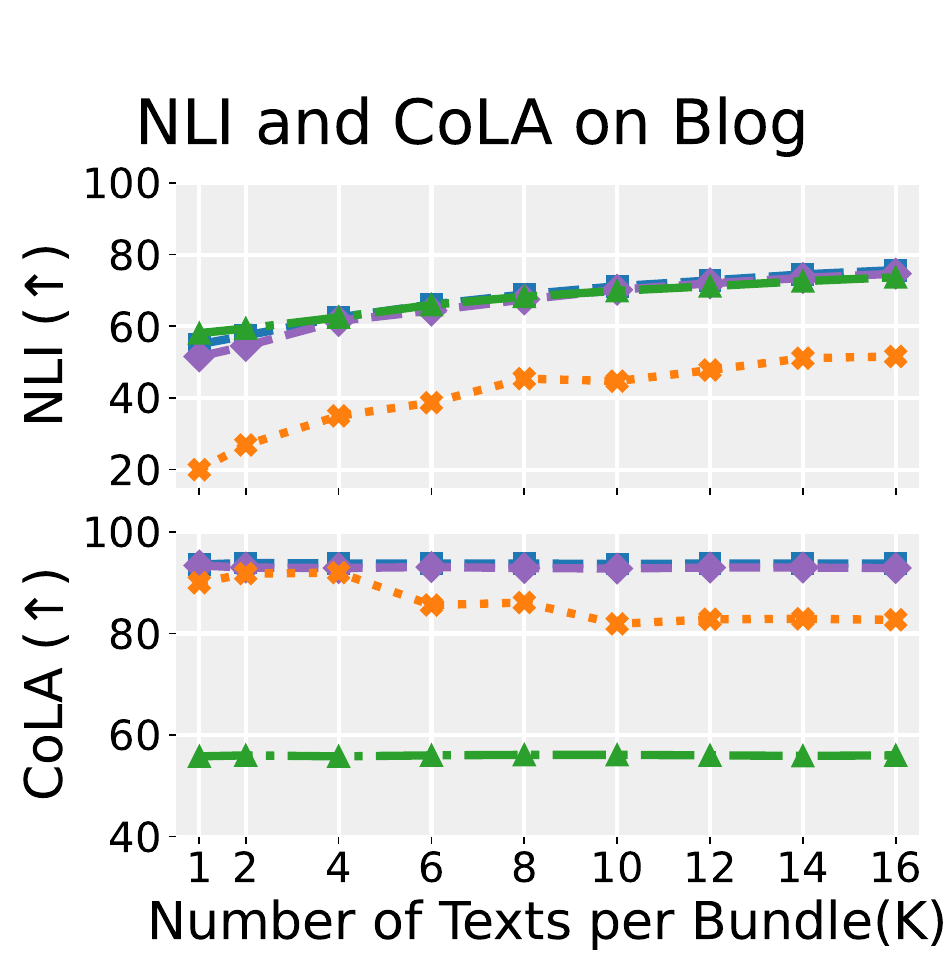}
            \caption{NLI and CoLA}
            \label{fig:blog_utility}
        \end{subfigure}

    \end{minipage}
    \hfill
    \begin{minipage}[c]{0.11\linewidth}
        \centering
        \raisebox{1.6cm}{\includegraphics[width=\linewidth]{latex/figure/k_legend.pdf}}
    \end{minipage}

    \caption{Attribution risk, verification risk, NLI, and CoLA across bundle sizes on Blog. In (c), each synthetic output is evaluated against its corresponding private text, so the private text serves as the reference and is not plotted.}
    \label{fig:blog_cross_k}
\end{figure*}

\section{Additional Details for AAST}
\label{appendix-additional-mthodology-details}

\subsection{Abstraction Models and Sentiment Alignment}
\label{appendix-abstraction-alignment}

For each private sample \(x\) and candidate \(a_j \in \mathcal{A}(x)\), we compute the composite score \(r(x,a_j)\), which combines semantic similarity, sentiment alignment, and mismatch penalty, and the highest scoring candidate is selected before generation. We use facebook/bart-large-cnn~\cite{DBLP:conf/acl/LewisLGGMLSZ20} to generate an initial abstraction candidate pool. We compute sentence embeddings with sentence-t5-base~\cite{ni2021sentencet5scalablesentenceencoders} and measure semantic similarity by cosine similarity. A sentiment model, siebert/sentiment-roberta-large-en~\cite{hartmann2023}, provides polarity \(y(a)\in\{0,1\}\). Following the abstraction setting~\cite{ma-rajtmajer-2026-private}, we set a sentiment control parameter for abstraction. The top $5$ candidates by \(\mathrm{score}(\cdot)\) form \(\mathcal{A}(x)\), which is the input to the noisy abstraction selection step in \S\ref{subsec:dp_selection}.

\subsection{Noisy Abstraction Selection}
\label{appendix-dp-aast}

Differential privacy (DP) provides a formal guarantee for randomized mechanisms~\cite{DBLP:journals/fttcs/DworkR14}. A mechanism \(M\) satisfies \((\epsilon,\delta)\)-DP if, for any neighboring datasets \(\mathcal{X}\) and \(\mathcal{X}'\) that differ in one element, and any event \(S \subseteq \mathrm{Range}(M)\),
\[
\Pr[M(\mathcal{X}) \in S]
\le
e^{\epsilon}\Pr[M(\mathcal{X}') \in S] + \delta .
\]

AAST can add Gaussian noise in the abstraction score selection step. The scope of the Gaussian noise analysis is limited: it applies only to the noisy score selection rule, conditioned on the candidate set used by the selector. Since the candidates are generated from the private input, we do not claim end-to-end DP for the selected abstraction text or the final synthetic corpus.

Because the raw composite score \(r(x,a_j)\) is not necessarily bounded, we first normalize scores within the candidate set \(\mathcal{A}(x)\):
\begingroup
\begin{equation}
\small
u(x,a_j) = \frac{
r(x,a_j)-\min\limits_{a_\ell \in \mathcal{A}(x)} r(x,a_\ell)}
{\max\limits_{a_\ell \in \mathcal{A}(x)} r(x,a_\ell)-
\min\limits_{a_\ell \in \mathcal{A}(x)} r(x,a_\ell)},
\label{eq:aast_norm_utility}
\end{equation}
\endgroup
when the denominator is nonzero, and set \(u(x,a_j)=0\) otherwise. This gives \(u(x,a_j)\in[0,1]\). We use \(\Delta u=1\) as the per candidate score sensitivity after normalization~\cite{xie2024differentially}.

Given privacy budget \(\epsilon\), number of private samples \(N\), and
$\delta = \frac{1}{N\log N}$,
we set the Gaussian noise scale as
\begingroup
\begin{equation}
\small
\sigma
=\frac{\sqrt{2\log(1.25/\delta)}\,\Delta u}{\epsilon}.
\label{eq:gaussian_sigma}
\end{equation}
\endgroup

For each private sample \(x \in \mathcal{X}\) and candidate \(a_j \in \mathcal{A}(x)\), we sample
$\tilde{u}_j = u(x,a_j) + \eta_j, \eta_j \sim \mathcal{N}(0,\sigma^2)$.
The noisy selected abstraction is
\begingroup
\begin{equation}
\small
a_{\mathrm{noisy}}(x)=\arg\max_{a_j \in \mathcal{A}(x)}\tilde{u}_j .
\label{eq:dp_selection}
\end{equation}
\endgroup

\subsection{Analysis for Noisy Score Selection}
\label{appendix-dp-privacy-guarantee}

\noindent\textbf{Proposition 1. Analysis of noisy score selection.}
For a private sample \(x\), let \(\mathcal{A}(x)=\{a_1,\dots,a_m\}\) be the candidate set used by the abstraction selector. Conditioned on this candidate set, define the normalized utility vector
$\mathbf{u}(x)=\bigl(u(x,a_1),\dots,u(x,a_m)\bigr)$,
where each \(u(x,a_j)\in[0,1]\). Assume that \(\mathbf{u}\) has global \(\ell_2\)-sensitivity at most \(\Delta u\) with respect to neighboring private inputs. If Gaussian noise with scale \(\sigma\) from Eq.~\ref{eq:gaussian_sigma} is added to \(\mathbf{u}(x)\), then the noisy score vector is \((\epsilon,\delta)\)-DP. Selecting the index of the largest noisy score is also \((\epsilon,\delta)\)-DP by post processing. This statement is conditional on the candidate set and does not cover candidate generation, the selected abstraction text, or the final synthetic corpus.\\

\noindent\textit{Proof.}
Consider the vector valued query \(\mathbf{u}(x)\), conditioned on the candidate set used by the selector. The mechanism releases
$\tilde{\mathbf{u}}(x)=\mathbf{u}(x)+\mathbf{z},\mathbf{z}\sim\mathcal{N}(0,\sigma^2 I)$,
where \(\sigma\) is calibrated to sensitivity \(\Delta u\) as in Eq.~\ref{eq:gaussian_sigma}. By the Gaussian mechanism~\cite{DBLP:journals/fttcs/DworkR14}, \(\tilde{\mathbf{u}}(x)\) satisfies \((\epsilon,\delta)\)-DP under this conditional score selection view. The selected index
$
j^*=\arg\max_j \tilde{u}_j
$
is a deterministic function of the noisy scores, and also satisfies \((\epsilon,\delta)\)-DP by post processing. 

\subsection{Complexity Analysis of AAST}
\label{appendix_complexity_analysis}

A direct search over all combinations in a \(K\) bundle is exponential in \(K\). For example, jointly searching over a Query bundle and a Target bundle requires evaluating \(C^{2K}\) combinations per user. We instead use coordinate descent, which updates one slot at a time while keeping all other slots fixed, and use one pass in our experiments.

For each slot update, AAST evaluates at most \(C\) candidate bundles through the episode embedding function. Since each user has \(K\) slots in Query and \(K\) slots in Target, one pass requires at most \(2K \cdot C\) episode embeddings per user. Across all users, the dominant episode embedding cost is
\[
\mathcal{O}\big(|\mathcal{U}| \cdot 2K \cdot C\big),
\]
where \(\mathcal{U}\) denotes the set of users.

The scoring objective also computes partner similarity and crowd similarity for each candidate. Partner similarity is constant cost once bundle embeddings are available. Crowd similarity compares the candidate bundle embedding with \(\mathcal{V}_{-u}\), adding similarity computations against other users' current bundle embeddings. Candidate Refinement adds text similarity computations within the near optimal set \(\mathcal{R}\), whose size is bounded by the candidate pool. These terms add similarity costs, but they do not change the dominant number of episode embedding evaluations. Thus, coordinate descent reduces the dominant search factor from exponential in \(K\) to linear in \(K\) and \(C\).

\section{Additional Experimental Details}
\label{appendix-additional-experimental-details}

\subsection{Prompt Design}
\label{appendix-prompt-design}

Because the input to the generator is abstracted private text, we use the following prompt for LLM generation: 

\noindent``\textit{Rewrite the abstracted text while preserving its core meaning. Vary sentence structure and word choice to reduce stylistic similarity.}'' 

This prompt makes clear that the input is already an abstraction of the private text, while guiding the generator to preserve the main meaning and reduce stylistic similarity. We use the same prompt for both Mistral-Small and Phi-4 generators.

\subsection{Baseline Selection and Setup}
\label{appendix-baseline-setup}

We choose KiP and JAMDEC as the main baselines because they are recent methods designed for authorship privacy and evaluated with authorship attacks. Many privacy rewriting methods focus on removing sensitive attributes, named entities, or private disclosures, which is related but not the same as reducing authorship attribution and verification risk. We focus on KiP and JAMDEC as the main baselines because they directly target author style obfuscation. In addition, we include RUPTA as an expanded LLM-based rewriting comparison, since it provides a recent text anonymization baseline without aggregation-aware bundle-level selection. This allows us to test whether strong LLM rewriting alone is sufficient under account-level aggregation.

To approximately match generation budgets, AAST and JAMDEC both use \texttt{max\_new\_tokens} = 128. KiP uses \texttt{token\_max\_length} instead. On the Reddit dataset, the average text length is 219.6 tokens, which gives 219.6 + 128 = 347.6. We therefore set \texttt{token\_max\_length} = 345. On the Blog dataset, the average text length is 243.4 tokens, which gives 243.4 + 128 = 371.4. Thus, we set \texttt{token\_max\_length} = 370. We also set both \texttt{lex\_diversity} and \texttt{order\_diversity} to 60 for KiP to encourage more diverse outputs.

\subsection{Models and Implementation Settings}
\label{appendix-implementation-settings}

Model choices and fixed implementation settings are summarized in Table~\ref{tab:aast_implementation_settings}. These values are held constant across all experiments.

\begin{table*}[t]
\centering
\fontsize{7.5}{8.0}\selectfont
\setlength{\tabcolsep}{3pt}
\renewcommand{\arraystretch}{1.08}
\begin{tabular}{
>{\raggedright\arraybackslash}p{3.8cm}
>{\raggedright\arraybackslash}p{4.5cm}
>{\raggedright\arraybackslash}p{5.4cm}}
\toprule
\textbf{Model or Parameter} & \textbf{Value} & \textbf{Purpose} \\
\midrule
Abstraction model (\S\ref{subsec:data_abstraction})
& \texttt{facebook/bart-large-cnn}~\cite{DBLP:conf/acl/LewisLGGMLSZ20}
& Generates abstraction candidates from each private text. \\
\midrule
Sentiment classifier model (\S\ref{subsec:data_abstraction})
& \texttt{sentiment-roberta-large-en}~\cite{hartmann2023}
& Checks whether abstraction candidates preserve the private text sentiment. \\
\midrule
Semantic scoring model  (\S\ref{subsec:dp_selection})
& \texttt{sentence-t5-base}~\cite{ni2021sentencet5scalablesentenceencoders}
& Computes semantic similarity between the private text and abstraction candidates. \\
\midrule
Generator (\S\ref{subsec:syn_generation})
&  \texttt{Mistral-Small}~\cite{mistralai2025mistralsmall32} and \texttt{Phi-4}~\cite{DBLP:journals/corr/abs-2412-08905}
& Generates synthetic candidate rewrites from selected abstractions. \\
\midrule
Style embedding model (\S\ref{subsec:candidate_preselect})
& \texttt{intfloat/e5-base-v2}~\cite{wang2022text} 
& Filters candidates that remain too close to the private input in stylistic space. \\
\midrule
Episode encoder model (\S\ref{subsubsec:candidate_scoring})
& \texttt{Luar-mud}~\cite{rivera2021learning} 
& Computes bundle embeddings for partner and crowd similarity scoring. \\
\midrule \midrule
Text similarity surrogate $n$
& \(\in \{3,4,5\}\)
& Character term frequency-inverse document frequency. Refines among near optimal candidates using bundle-level textual similarity. \\

\midrule
Initial abstraction candidates $h$
& 10
& Number of candidates generated before abstraction candidate selection. \\
\midrule
Abstraction Candidate set size
& 5
& Number of abstraction candidates scored for noisy selection. \\

\midrule
Abstraction length
& Maximum 150 tokens; minimum 50 tokens
& Controls the length of abstraction candidates. \\
\midrule
Synthetic candidates per text \(C\)
& 4
& Number of generated candidates by generator for each selected abstraction. \\
\midrule
Generator settings
& Temperature 0.5~\cite{mistralai2025mistralsmall32} for Mistral-Small; temperature 1.0 for Phi-4; repetition penalty 1.0
& Fixed generator settings used across experiments. \\

\midrule
Partner/crowd weight $\alpha$
& 0.6
& Balances partner similarity reduction and crowd similarity increase. \\
\midrule
Crowd neighborhood size $m$
& Top 8 other-user bundles
& Computes crowd similarity using the nearest other-user bundle embeddings. \\
\midrule
Near optimal margin \(\tau\)
& 0.002
& Defines the near optimal candidate set after bundle-level scoring. \\
\bottomrule
\end{tabular}
\caption{Models and fixed implementation settings for AAST. These values are held constant across all experiments.}
\label{tab:aast_implementation_settings}
\end{table*}

\begin{figure}[t]
    \centering
    \begin{minipage}{0.49\linewidth} 
        \centering
        \includegraphics[width=\linewidth]{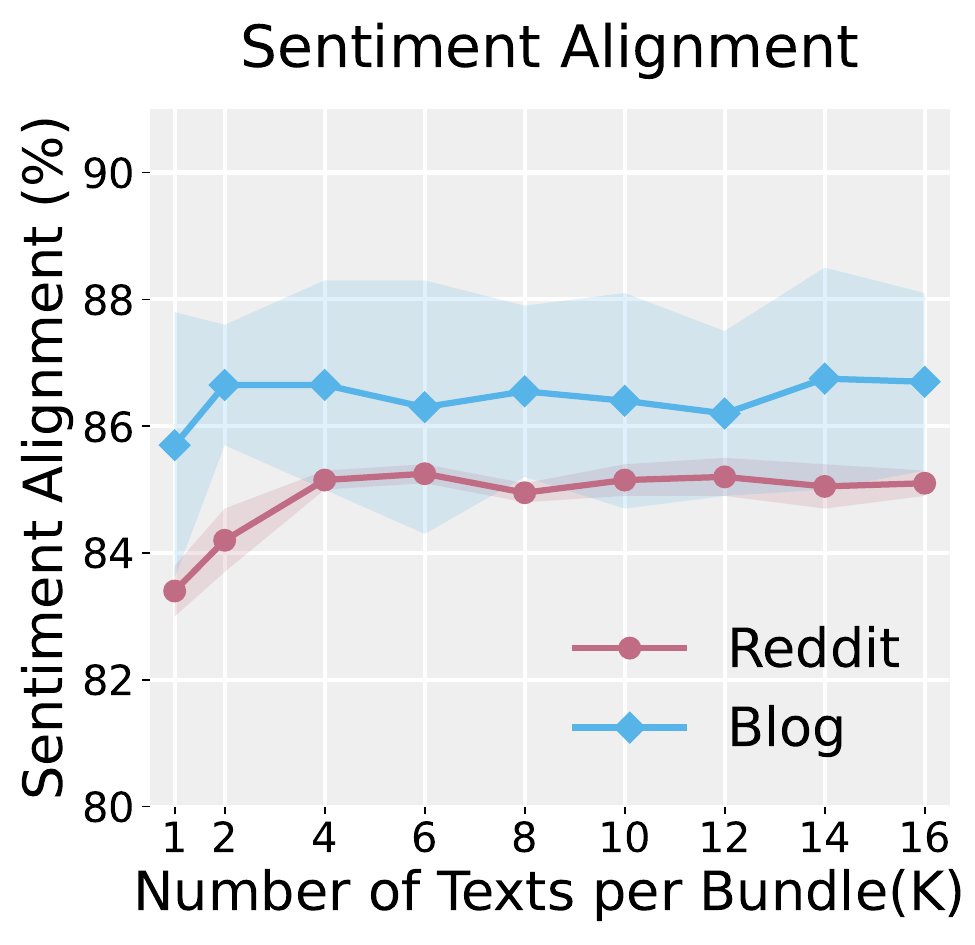}
        \caption{Sentiment alignment across bundle sizes and datasets.}
        \label{fig:utili_senti}
    \end{minipage}
    \hfill
    \begin{minipage}{0.49\linewidth}
        \centering
        \includegraphics[width=\linewidth]{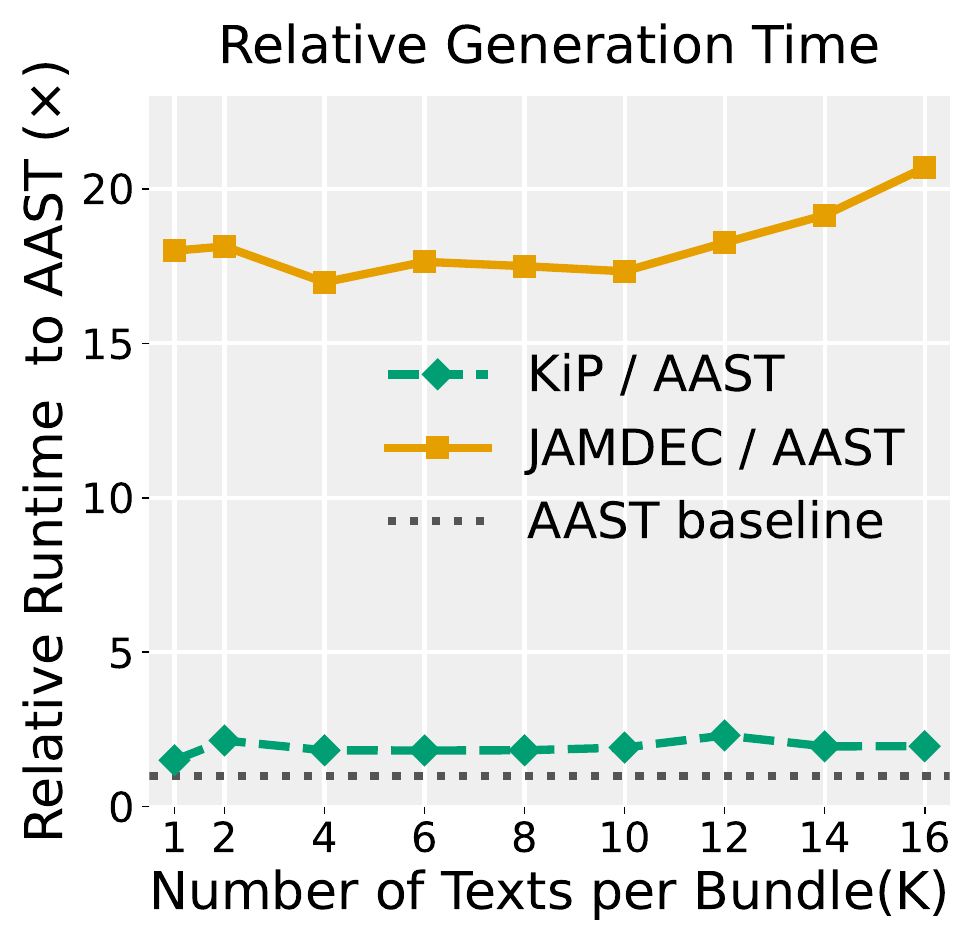}
        \caption{Relative generation time compared with AAST$_{\mathrm{M}}$.}
        \label{fig:time_cost}
    \end{minipage}
\end{figure}

\section{Additional Metrics Details}
\label{appendix-metrics-detail}

\subsection{Privacy Metrics Details}
\label{appendix-privacy-metrics}

For same-genre evaluation, we use authorship attribution and verification attacks. LUAR attribution ranks the true user among a candidate set for each bundle~\cite{rivera2021learning}. We report Mean Reciprocal Rank (MRR)~\cite{rivera2021learning}, which averages the reciprocal rank of the true user, and Recall at 8 (R@8)~\cite{bao2024keep}, which measures whether the true user appears among the top eight candidates. Lower values indicate lower attribution risk. To test transfer beyond LUAR, we also include an independent classical stylometric attribution attack~\cite{basile2017n,pedregosa2011scikit}. This attacker uses character 2--5-grams, distance-character 1--3-grams, and word 1--3-grams, and ranks candidate users for each synthetic bundle. This evaluation tests whether AAST's attribution gains transfer from the LUAR-based attacker-aware setting to a non-neural stylometric attacker.

PAN22 verification decides whether two bundles are written by the same user~\cite{stamatatos2022overview}. We report Area Under the Curve (AUC)~\cite{stamatatos2022overview} and Correctness at One (c@1)~\cite{penas-rodrigo-2011-simple}. For privacy, AUC closer to random chance 50\% and lower c@1 indicate weaker verification. We also include a classical stylometric transfer attack to test whether the same synthetic outputs reduce verification risk under a different attack family~\cite{basile2017n,pedregosa2011scikit}. The verifier represents each bundle with character  3--5-gram term frequency--inverse document frequency features and trains a logistic regression classifier on same author and different author bundle pairs. We split users into two disjoint halves, train on one half, and evaluate on the held-out half.

Following KiP~\cite{bao2024keep}, we evaluate bundle sizes \(K \in \{1,2,4,6,8,10,12,14,16\}\), where $K$ is the number of texts from the same user grouped into each bundle. Each text is one post or blog entry. This sweep measures how privacy risk changes as the attacker observes more texts jointly.

For cross-genre evaluation, we use SELMA, the embedding authorship model introduced with CROSSNEWS~\cite{ma2025crossnews}. Following CROSSNEWS, we use the strongest SELMA prompt variants for each task. For attribution, we use SELMA+TaskOnly and SELMA+LIP; for verification, we use SELMA+TaskOnly and SELMA+PromptAV~\cite{hung2023wrote,huang2024can,ma2025crossnews}. We report the same metrics as CROSSNEWS: top one Accuracy, R@8, and Average Rank for attribution, and Accuracy and F1 for verification.

\subsection{Utility Metrics Details}
\label{appendix-utility-metrics}

We evaluate synthetic text quality along three dimensions. Semantic quality is measured by Natural Language Inference (NLI)~\cite{liu-etal-2022-wanli}. Given a private text and its synthetic counterpart, NLI measures whether the synthetic text preserves the core meaning of the private text.

Linguistic acceptability is measured by Corpus of Linguistic Acceptability (CoLA)~\cite{warstadt2019neural}. CoLA serves as an indicator of grammaticality and fluency.

Sentiment alignment measures whether the synthetic text preserves the sentiment polarity of the private text. We use the sentiment classifier from \S\ref{subsec:models} to assign sentiment labels to private and synthetic texts~\cite{hartmann2023}, and report the proportion of matched labels.

\section{Additional Same-genre Authorship Results}
\label{appendix-same-genre-authorship}

Tables~\ref{tab:attribution-reddit-blog} and~\ref{tab:verification-reddit-blog} report the full same-genre authorship attribution and verification risk results on Reddit and Blog across bundle sizes.

\begin{table}[t]
    \centering
    {
    \fontsize{6.5}{7.5}\selectfont
    \begin{tabular}{
        >{\centering\arraybackslash}p{0.4cm}
        >{\centering\arraybackslash}p{1.0cm}
        |
        >{\centering\arraybackslash}p{0.7cm}
        >{\centering\arraybackslash}p{0.7cm}
        ||
        >{\centering\arraybackslash}p{0.7cm}
        >{\centering\arraybackslash}p{0.7cm}
    }
        \toprule
        \multirow{2}{*}{\textbf{$K$}} 
        & \multirow{2}{*}{\textbf{Method}}
        & \multicolumn{2}{c|}{\textbf{Reddit}}
        & \multicolumn{2}{c}{\textbf{Blog}} \\
        \cmidrule(lr){3-4}\cmidrule(lr){5-6}
        & & \textbf{MRR}($\downarrow$) & \textbf{R@8}($\downarrow$)
          & \textbf{MRR}($\downarrow$) & \textbf{R@8}($\downarrow$) \\
        \midrule

        \multirow{5}{*}{1}
        & Private          & 32.2 & 45.6 & 38.5 & 52.4 \\
        & AAST$_{\Phi}$    & \uline{2.4} & \uline{3.6} & \textbf{1.7} & \textbf{2.0} \\
        & AAST$_{\mathrm{M}}$ & \textbf{2.2} & \textbf{2.9} & \uline{2.6} & \uline{2.4} \\
        & JAMDEC           & 4.5 & 8.4 & 7.4 & 12.8 \\
        & KiP              & 9.4 & 17.6 & 8.2 & 12.8 \\
        \cmidrule{1-6}

        \multirow{5}{*}{2}
        & Private          & 67.1 & 81.2 & 61.2 & 72.4 \\
        & AAST$_{\Phi}$    & \uline{5.5} & \uline{7.3} & \textbf{3.2} & \uline{5.7} \\
        & AAST$_{\mathrm{M}}$ & \textbf{3.2} & \textbf{3.8} & \uline{3.3} & \textbf{4.9} \\
        & JAMDEC           & 9.9 & 14.0 & 12.5 & 22.0 \\
        & KiP              & 20.2 & 38.4 & 17.3 & 28.8 \\
        \cmidrule{1-6}

        \multirow{5}{*}{4}
        & Private          & 89.8 & 98.0 & 83.1 & 91.2 \\
        & AAST$_{\Phi}$    & \uline{6.7} & \uline{11.8} & \textbf{7.0} & \textbf{11.5} \\
        & AAST$_{\mathrm{M}}$ & \textbf{5.6} & \textbf{9.1} & \uline{8.6} & \uline{15.9} \\
        & JAMDEC           & 18.1 & 27.2 & 20.3 & 35.6 \\
        & KiP              & 42.8 & 63.6 & 28.4 & 49.6 \\
        \cmidrule{1-6}

        \multirow{5}{*}{6}
        & Private          & 96.0 & 99.2 & 88.7 & 96.0 \\
        & AAST$_{\Phi}$    & \uline{11.8} & \textbf{19.3} & \textbf{8.8} & \textbf{14.9} \\
        & AAST$_{\mathrm{M}}$ & \textbf{11.2} & \uline{19.7} & \uline{15.6} & \uline{21.8} \\
        & JAMDEC           & 24.5 & 38.0 & 28.4 & 44.8 \\
        & KiP              & 53.8 & 72.0 & 45.9 & 66.8 \\
        \cmidrule{1-6}

        \multirow{5}{*}{8}
        & Private          & 97.8 & 100 & 93.8 & 98.0 \\
        & AAST$_{\Phi}$    & \textbf{17.0} & \uline{31.8} & \textbf{14.9} & \textbf{21.3} \\
        & AAST$_{\mathrm{M}}$ & \uline{18.4} & \textbf{28.4} & \uline{24.9} & \uline{40.3} \\
        & JAMDEC           & 24.4 & 41.6 & 33.0 & 52.0 \\
        & KiP              & 68.0 & 86.4 & 60.3 & 73.2 \\
        \cmidrule{1-6}

        \multirow{5}{*}{10}
        & Private          & 99.3 & 100 & 95.3 & 99.2 \\
        & AAST$_{\Phi}$    & \textbf{25.5} & \textbf{39.7} & \textbf{20.7} & \textbf{34.3} \\
        & AAST$_{\mathrm{M}}$ & \uline{25.9} & \uline{43.9} & \uline{34.1} & \uline{48.4} \\
        & JAMDEC           & 36.0 & 56.4 & 43.5 & 63.6 \\
        & KiP              & 73.3 & 90.0 & 66.6 & 78.8 \\
        \cmidrule{1-6}

        \multirow{5}{*}{12}
        & Private          & 99.5 & 100 & 95.8 & 98.8 \\
        & AAST$_{\Phi}$    & \uline{33.9} & \textbf{49.8} & \textbf{24.4} & \textbf{38.0} \\
        & AAST$_{\mathrm{M}}$ & \textbf{31.6} & \uline{50.5} & \uline{36.3} & \uline{56.7} \\
        & JAMDEC           & 39.6 & 60.0 & 50.5 & 70.4 \\
        & KiP              & 79.1 & 91.2 & 73.0 & 86.8 \\
        \cmidrule{1-6}

        \multirow{5}{*}{14}
        & Private          & 99.6 & 100 & 97.0 & 99.2 \\
        & AAST$_{\Phi}$    & \textbf{35.5} & \textbf{54.2} & \textbf{31.4} & \textbf{48.0} \\
        & AAST$_{\mathrm{M}}$ & \uline{37.4} & \uline{55.1} & \uline{42.9} & \uline{62.5} \\
        & JAMDEC           & 52.6 & 72.4 & 54.7 & 80.4 \\
        & KiP              & 82.7 & 93.6 & 76.6 & 92.0 \\
        \cmidrule{1-6}

        \multirow{5}{*}{16}
        & Private          & 99.8 & 100 & 97.5 & 99.2 \\
        & AAST$_{\Phi}$    & \textbf{39.8} & \textbf{60.7} & \textbf{41.1} & \textbf{61.4} \\
        & AAST$_{\mathrm{M}}$ & \uline{43.5} & \uline{63.7} & \uline{44.5} & \uline{65.7} \\
        & JAMDEC           & 54.7 & 74.8 & 62.4 & 83.6 \\
        & KiP              & 92.7 & 97.8 & 83.5 & 94.0 \\

        \bottomrule
    \end{tabular}
    }
    \caption{Attribution risk results on Reddit and Blog across bundle sizes.}
    \label{tab:attribution-reddit-blog}
\end{table}

\begin{table}[t]
    \centering
    {
    \fontsize{6.5}{7.5}\selectfont
    \begin{tabular}{
        >{\centering\arraybackslash}p{0.4cm}
        >{\centering\arraybackslash}p{1.0cm}
        |
        >{\centering\arraybackslash}p{0.8cm}
        >{\centering\arraybackslash}p{0.8cm}
        ||
        >{\centering\arraybackslash}p{0.8cm}
        >{\centering\arraybackslash}p{0.8cm}
    }
        \toprule
        \multirow{2}{*}{\textbf{$K$}} 
        & \multirow{2}{*}{\textbf{Method}}
        & \multicolumn{2}{c||}{\textbf{Reddit}}
        & \multicolumn{2}{c}{\textbf{Blog}} \\
        \cmidrule(lr){3-4}\cmidrule(lr){5-6}
        & & \textbf{AUC}($\approx$50) & \textbf{c@1}($\downarrow$)
          & \textbf{AUC}($\approx$50) & \textbf{c@1}($\downarrow$) \\
        \midrule

        \multirow{5}{*}{1}
        & Private              & 51.9 & 52.9 & 64.6 & 60.6 \\
        & AAST$_{\Phi}$        & \textbf{50.6} & \uline{51.7} & \uline{48.3} & \textbf{48.6} \\
        & AAST$_{\mathrm{M}}$  & \uline{48.9} & \textbf{44.6} & \textbf{47.6} & \uline{49.3} \\
        & JAMDEC               & 51.2 & 52.0 & 60.8 & 53.4 \\
        & KiP                  & 51.3 & 52.6 & 63.2 & 57.7 \\
        \cmidrule{1-6}

        \multirow{5}{*}{2}
        & Private              & 60.9 & 58.9 & 68.9 & 60.3 \\
        & AAST$_{\Phi}$        & 57.4 & \uline{50.3} & \textbf{53.9} & \textbf{47.4} \\
        & AAST$_{\mathrm{M}}$  & \textbf{47.9} & \textbf{49.4} & \uline{56.5} & \uline{48.4} \\
        & JAMDEC               & \uline{54.3} & 51.6 & 61.1 & 51.4 \\
        & KiP                  & 62.5 & 52.8 & 65.8 & 51.8 \\
        \cmidrule{1-6}

        \multirow{5}{*}{4}
        & Private              & 72.6 & 66.3 & 72.5 & 67.6 \\
        & AAST$_{\Phi}$        & \uline{61.7} & \uline{48.3} & \uline{58.3} & \textbf{48.2} \\
        & AAST$_{\mathrm{M}}$  & \textbf{55.5} & \textbf{48.2} & \textbf{53.7} & \uline{48.7} \\
        & JAMDEC               & 69.8 & 55.6 & 62.4 & 57.2 \\
        & KiP                  & 70.9 & 60.7 & 67.4 & 57.2 \\
        \cmidrule{1-6}

        \multirow{5}{*}{6}
        & Private              & 73.7 & 70.0 & 75.0 & 67.4 \\
        & AAST$_{\Phi}$        & \uline{67.7} & \uline{49.4} & 62.1 & \textbf{48.8} \\
        & AAST$_{\mathrm{M}}$  & \textbf{65.8} & \textbf{48.4} & \textbf{55.7} & \uline{49.0} \\
        & JAMDEC               & 69.7 & 52.6 & \uline{61.6} & 55.1 \\
        & KiP                  & 72.3 & 66.1 & 69.5 & 55.3 \\
        \cmidrule{1-6}

        \multirow{5}{*}{8}
        & Private              & 78.8 & 71.7 & 79.3 & 72.9 \\
        & AAST$_{\Phi}$        & \uline{65.1} & \uline{48.3} & 69.9 & \uline{48.4} \\
        & AAST$_{\mathrm{M}}$  & \textbf{64.2} & \textbf{48.2} & \textbf{62.1} & \textbf{48.2} \\
        & JAMDEC               & 77.0 & 52.0 & \uline{65.3} & 57.7 \\
        & KiP                  & 77.4 & 64.2 & 75.5 & 59.0 \\
        \cmidrule{1-6}

        \multirow{5}{*}{10}
        & Private              & 77.0 & 69.3 & 81.7 & 74.6 \\
        & AAST$_{\Phi}$        & \uline{69.3} & \uline{49.7} & \uline{67.0} & \uline{50.4} \\
        & AAST$_{\mathrm{M}}$  & \textbf{61.2} & \textbf{47.5} & \textbf{66.3} & \textbf{49.3} \\
        & JAMDEC               & 74.7 & 55.9 & 70.7 & 59.6 \\
        & KiP                  & 73.6 & 67.6 & 79.5 & 58.9 \\
        \cmidrule{1-6}

        \multirow{5}{*}{12}
        & Private              & 82.3 & 76.9 & 81.6 & 78.4 \\
        & AAST$_{\Phi}$        & \textbf{64.6} & \uline{48.2} & \textbf{67.3} & \textbf{48.2} \\
        & AAST$_{\mathrm{M}}$  & \uline{66.9} & \textbf{48.0} & \uline{67.4} & \uline{49.2} \\
        & JAMDEC               & 79.3 & 64.2 & 71.3 & 57.0 \\
        & KiP                  & 80.5 & 71.7 & 80.9 & 66.4 \\
        \cmidrule{1-6}

        \multirow{5}{*}{14}
        & Private              & 86.6 & 82.8 & 82.8 & 81.6 \\
        & AAST$_{\Phi}$        & \uline{69.0} & \uline{48.8} & \textbf{67.5} & \uline{51.4} \\
        & AAST$_{\mathrm{M}}$  & \textbf{66.0} & \textbf{48.0} & \uline{67.8} & \textbf{49.5} \\
        & JAMDEC               & 81.1 & 69.6 & 72.8 & 57.7 \\
        & KiP                  & 84.1 & 79.2 & 83.0 & 68.3 \\
        \cmidrule{1-6}

        \multirow{5}{*}{16}
        & Private              & 89.7 & 83.6 & 82.3 & 79.1 \\
        & AAST$_{\Phi}$        & \uline{66.7} & \uline{49.8} & \uline{71.2} & \uline{53.2} \\
        & AAST$_{\mathrm{M}}$  & \textbf{64.1} & \textbf{48.0} & \textbf{68.6} & \textbf{50.3} \\
        & JAMDEC               & 81.4 & 73.1 & 73.3 & 61.4 \\
        & KiP                  & 92.6 & 86.1 & 85.3 & 68.1 \\

        \bottomrule
    \end{tabular}
    }
    \caption{Verification risk results on Reddit and Blog across bundle sizes.}
    \label{tab:verification-reddit-blog}
\end{table}

\section{Additional Same-genre Stylometric Attribution Results}
\label{appendix-same-genre-stylo-attri}

Table~\ref{tab:stylometric-attribution} reports same-genre attribution results under a classical stylometric transfer attack, with 95\% bootstrap confidence intervals computed over users. This unseen stylometric attacker evaluates whether AAST's privacy gains transfer beyond the LUAR attribution model used in bundle-level selection.

The results show that AAST remains effective under this independent attacker. Both AAST variants reduce attribution risk compared with KiP and JAMDEC, especially at larger bundle sizes. At $K$=16, AAST gives the lowest and second-lowest MRR and R@8 on both datasets, while KiP and JAMDEC remain substantially more linkable. These results suggest that AAST's protection is not specific to LUAR and transfers to a classical stylometric attribution setting.

\begin{table*}[t]
    \centering
    {
    \fontsize{7}{8}\selectfont
    \begin{tabular}{
        >{\centering\arraybackslash}p{0.6cm}
        |
        >{\centering\arraybackslash}p{1.5cm}
        |
        >{\centering\arraybackslash}p{2.4cm}
        >{\centering\arraybackslash}p{2.4cm}
        ||
        >{\centering\arraybackslash}p{2.4cm}
        >{\centering\arraybackslash}p{2.4cm}
    }
        \toprule
        \multirow{2}{*}{\textbf{$K$}} 
        & \multirow{2}{*}{\textbf{Method}}
        & \multicolumn{2}{c|}{\textbf{Reddit}}
        & \multicolumn{2}{c}{\textbf{Blog}} \\
        \cmidrule(lr){3-4}\cmidrule(lr){5-6}
        & & \textbf{MRR}($\downarrow$) & \textbf{R@8}($\downarrow$)
          & \textbf{MRR}($\downarrow$) & \textbf{R@8}($\downarrow$) \\
        \midrule

        \multirow{5}{*}{1}
        & Private          & 17.5 [14.1, 20.9] & 28.0 [23.6, 32.8] & 14.2 [11.2, 17.4] & 25.6 [21.2, 30.4] \\
        & AAST$_{\Phi}$    & \textbf{3.9} [2.6, 5.4] & \uline{6.4} [4.6, 8.6] & \uline{3.4} [2.2, 4.8] & \textbf{3.6} [2.4, 5.2] \\
        & AAST$_{\mathrm{M}}$ & 5.3 [3.8, 6.8] & 8.9 [6.6, 11.6] & \textbf{2.4} [1.6, 3.4] & \textbf{3.6} [2.4, 5.0] \\
        & JAMDEC           & \uline{4.2} [2.8, 6.0] & \textbf{5.6} [3.8, 7.8] & 7.9 [5.8, 10.4] & 16.0 [12.6, 19.8] \\
        & KiP              & 7.8 [5.6, 10.2] & 11.6 [8.6, 14.8] & 5.1 [3.6, 6.8] & 8.8 [6.4, 11.6] \\
        \cmidrule{1-6}

        \multirow{5}{*}{8}
        & Private          & 52.0 [46.0, 58.2] & 70.4 [64.0, 76.8] & 45.7 [40.2, 51.6] & 66.8 [60.8, 72.8] \\
        & AAST$_{\Phi}$    & \textbf{8.0} [6.0, 10.4] & \textbf{15.6} [12.0, 19.6] & \uline{11.3} [8.6, 14.4] & \uline{19.2} [15.0, 23.8] \\
        & AAST$_{\mathrm{M}}$ & \uline{10.9} [8.4, 13.6] & \uline{18.4} [14.6, 22.6] & \textbf{7.6} [5.6, 9.8] & \textbf{14.4} [11.0, 18.0] \\
        & JAMDEC           & 13.8 [10.6, 17.4] & 24.8 [20.0, 29.8] & 25.8 [21.0, 30.8] & 44.4 [38.6, 50.4] \\
        & KiP              & 21.1 [16.8, 25.6] & 38.0 [32.0, 44.4] & 25.9 [21.0, 31.0] & 42.4 [36.4, 48.6] \\
        \cmidrule{1-6}

        \multirow{5}{*}{16}
        & Private          & 59.8 [53.0, 66.6] & 79.2 [73.0, 85.0] & 62.6 [56.0, 69.4] & 84.0 [78.0, 89.2] \\
        & AAST$_{\Phi}$    & \uline{13.8} [10.6, 17.6] & \textbf{22.4} [17.8, 27.4] & \uline{18.1} [14.2, 22.4] & \uline{30.4} [25.0, 36.2] \\
        & AAST$_{\mathrm{M}}$ & \textbf{12.6} [9.8, 15.6] & \uline{25.6} [20.6, 30.8] & \textbf{11.8} [9.0, 14.8] & \textbf{21.2} [17.0, 25.6] \\
        & JAMDEC           & 21.0 [17.0, 25.4] & 39.8 [34.2, 45.6] & 30.1 [25.0, 35.6] & 52.0 [46.0, 58.2] \\
        & KiP              & 44.6 [38.0, 51.6] & 64.3 [57.0, 71.6] & 42.8 [36.4, 49.6] & 66.0 [59.0, 72.8] \\

        \bottomrule
    \end{tabular}
    }
    \caption{Same-genre attribution results on Reddit and Blog under an independent classical n-gram stylometric attacker. The attributor uses character 2--5-grams, distance-character 1--3-grams, and word 1--3-grams. Values are percentages with 95\% bootstrap confidence intervals computed over users.}
    \label{tab:stylometric-attribution}
\end{table*}

\section{Additional Same-genre Stylometric Verification Results}
\label{appendix-same-genre-stylo-verifi}

Table~\ref{tab:stylometric-verifi} reports same-genre verification results under the classical stylometric transfer attack. Figure~\ref{fig:stylo_verifi} shows the corresponding trends. The results are consistent with the main PAN22 verification results: AAST keeps verification risk lower than JAMDEC and KiP across most bundle sizes on both Reddit and Blog.

At larger $K$, the gap becomes clearer. On Reddit at $K$=16, AAST$_{\mathrm{M}}$ gives 71.6 AUC and 52.4 c@1, while JAMDEC gives 90.1 AUC and 75.3 c@1, and KiP gives 84.2 AUC and 67.0 c@1. On Blog at $K$=16, AAST$_{\Phi}$ gives the lowest AUC 68.6, while AAST$_{\mathrm{M}}$ gives 72.1 AUC and 55.2 c@1. In contrast, JAMDEC gives 94.0 AUC and 84.0 c@1, and KiP gives 96.1 AUC and 74.4 c@1.

These results provide additional transfer evidence that AAST's verification gains are not limited to neural authorship models, but persist under a classical stylometric attack. Together with the stylometric attribution results in Appendix~\ref{appendix-same-genre-stylo-attri}, they support our design goal: AAST reduces account-level authorship linkability while preserving useful text content instead of trying to remove all information from the synthetic text.

\begin{table*}[t]
\centering
\fontsize{7.0}{8.0}\selectfont
\setlength{\tabcolsep}{2.0pt}
\begin{tabular}{
>{\centering\arraybackslash}p{0.8cm}|
>{\raggedright\arraybackslash}p{1.5cm}|
>{\centering\arraybackslash}p{2.35cm}
>{\centering\arraybackslash}p{2.35cm}|
>{\centering\arraybackslash}p{2.35cm}
>{\centering\arraybackslash}p{2.35cm}}
\toprule
\multirow{2}{*}{\textbf{$K$}} & \multirow{2}{*}{\textbf{Method}} 
& \multicolumn{2}{c|}{\textbf{Reddit}} 
& \multicolumn{2}{c}{\textbf{Blog}} \\
\cmidrule(lr){3-4}\cmidrule(lr){5-6}
& & \textbf{AUC $\approx$50} & \textbf{c@1 $\downarrow$} & \textbf{AUC $\approx$50} & \textbf{c@1 $\downarrow$} \\
\midrule
\multirow{5}{*}{1}
& Private & 54.2 [47.2, 61.0] & 50.8 [44.4, 56.8] & 60.0 [52.6, 66.7] & 51.6 [45.6, 57.6] \\
& AAST$_{\mathrm{M}}$ & 52.5 [45.5, 59.9] & 49.2 [43.0, 55.4] & \textbf{50.2} [43.2, 57.3] & 50.8 [44.8, 56.9] \\
& AAST$_{\Phi}$ & \textbf{49.8} [42.6, 57.5] & 49.4 [43.0, 55.4] & \underline{50.3} [43.1, 57.9] & \textbf{49.2} [42.8, 55.2] \\
& JAMDEC & \underline{50.5} [43.3, 57.8] & \textbf{48.4} [42.4, 54.8] & 64.0 [57.2, 70.4] & 59.2 [52.8, 65.6] \\
& KiP & 53.7 [46.2, 61.2] & \underline{48.8} [42.0, 54.8] & \underline{50.3} [42.8, 57.7] & \underline{50.0} [43.6, 56.0] \\
\midrule
\multirow{5}{*}{2}
& Private & 69.8 [63.1, 76.3] & 55.2 [48.4, 61.2] & 75.9 [69.3, 81.5] & 64.4 [58.0, 70.0] \\
& AAST$_{\mathrm{M}}$ & \underline{53.7} [46.3, 60.8] & \textbf{49.6} [43.2, 55.6] & \underline{49.0} [41.6, 56.1] & \underline{50.8} [44.0, 56.4] \\
& AAST$_{\Phi}$ & \textbf{50.6} [43.8, 58.3] & \textbf{49.6} [43.2, 55.6] & \textbf{49.2} [42.1, 56.5] & \textbf{49.6} [43.2, 55.6] \\
& JAMDEC & 58.1 [50.7, 65.0] & 52.0 [46.0, 57.6] & 73.3 [66.7, 79.3] & 63.6 [57.6, 69.2] \\
& KiP & 57.4 [50.1, 64.4] & 53.6 [47.2, 59.6] & 57.9 [50.7, 65.3] & 52.0 [45.2, 57.6] \\
\midrule
\multirow{5}{*}{4}
& Private & 80.9 [75.4, 85.9] & 62.4 [56.0, 68.4] & 86.5 [81.9, 90.9] & 76.4 [70.4, 81.6] \\
& AAST$_{\mathrm{M}}$ & \textbf{50.9} [43.5, 58.2] & \underline{52.8} [46.4, 58.8] & \underline{49.0} [41.7, 56.2] & \underline{51.2} [44.4, 57.2] \\
& AAST$_{\Phi}$ & \underline{55.1} [47.8, 62.1] & \textbf{50.4} [44.0, 56.4] & \textbf{49.5} [42.1, 56.6] & \textbf{48.8} [42.4, 54.4] \\
& JAMDEC & 69.8 [62.8, 76.3] & 58.8 [52.4, 64.8] & 83.1 [78.2, 87.4] & 70.8 [65.2, 76.4] \\
& KiP & 59.9 [52.7, 66.7] & 53.0 [47.0, 59.0] & 66.2 [59.6, 72.5] & 54.0 [47.6, 59.6] \\
\midrule
\multirow{5}{*}{6}
& Private & 92.2 [89.1, 95.1] & 74.0 [68.4, 79.6] & 90.8 [87.5, 94.1] & 80.0 [74.4, 84.8] \\
& AAST$_{\mathrm{M}}$ & \textbf{58.2} [51.1, 65.5] & \textbf{51.2} [44.8, 56.8] & \underline{53.2} [46.2, 60.0] & \textbf{50.0} [43.6, 56.0] \\
& AAST$_{\Phi}$ & \underline{63.3} [56.2, 69.7] & \underline{53.2} [47.2, 59.2] & \textbf{52.8} [45.2, 59.8] & \textbf{50.0} [43.6, 55.6] \\
& JAMDEC & 73.9 [66.9, 80.4] & 55.3 [48.4, 62.1] & 84.2 [79.6, 88.7] & 74.4 [68.8, 79.6] \\
& KiP & 64.4 [56.2, 72.3] & 54.5 [47.7, 61.2] & 75.1 [68.8, 81.1] & 58.4 [52.0, 64.0] \\
\midrule
\multirow{5}{*}{8}
& Private & 93.3 [90.3, 96.1] & 78.0 [72.8, 82.8] & 93.8 [91.0, 96.2] & 83.6 [78.4, 88.0] \\
& AAST$_{\mathrm{M}}$ & \textbf{61.4} [54.4, 68.5] & \textbf{50.8} [44.4, 56.8] & \textbf{55.0} [47.7, 62.3] & \underline{52.4} [46.0, 58.0] \\
& AAST$_{\Phi}$ & \underline{67.1} [60.4, 74.1] & \underline{53.6} [47.6, 59.6] & \underline{58.6} [51.4, 65.1] & \textbf{50.8} [44.0, 56.8] \\
& JAMDEC & 79.8 [74.4, 85.1] & 68.8 [63.2, 74.4] & 87.5 [83.6, 91.5] & 76.0 [70.8, 81.6] \\
& KiP & 75.3 [69.6, 81.3] & 57.2 [50.8, 63.2] & 82.4 [77.3, 87.3] & 61.2 [54.8, 67.2] \\
\midrule
\multirow{5}{*}{10}
& Private & 94.1 [91.2, 96.6] & 78.4 [73.2, 83.2] & 95.8 [93.5, 97.6] & 84.8 [80.4, 89.2] \\
& AAST$_{\mathrm{M}}$ & \underline{66.8} [60.4, 73.4] & \textbf{51.6} [45.2, 57.6] & \underline{59.8} [52.4, 66.9] & \underline{51.6} [45.2, 57.6] \\
& AAST$_{\Phi}$ & \textbf{66.0} [59.4, 73.2] & \underline{52.8} [46.8, 58.8] & \textbf{58.3} [50.6, 65.2] & \textbf{51.2} [44.8, 56.8] \\
& JAMDEC & 80.6 [74.5, 86.7] & 65.3 [58.4, 72.1] & 89.5 [85.6, 93.1] & 74.4 [68.4, 80.0] \\
& KiP & 76.5 [70.0, 82.7] & 60.6 [53.8, 67.5] & 89.7 [85.9, 93.1] & 62.4 [55.6, 68.4] \\
\midrule
\multirow{5}{*}{12}
& Private & 94.5 [91.7, 97.0] & 82.8 [78.4, 87.2] & 97.2 [95.5, 98.6] & 87.2 [82.8, 90.8] \\
& AAST$_{\mathrm{M}}$ & \textbf{64.7} [58.3, 71.5] & \underline{53.6} [47.2, 59.6] & \underline{65.3} [58.8, 72.3] & \underline{53.2} [46.4, 59.2] \\
& AAST$_{\Phi}$ & \underline{67.0} [60.7, 73.8] & \textbf{52.8} [47.2, 58.8] & \textbf{63.2} [55.6, 69.9] & \textbf{51.2} [44.4, 56.8] \\
& JAMDEC & 82.5 [75.8, 88.0] & 67.9 [61.0, 74.2] & 91.3 [87.8, 94.6] & 78.4 [73.2, 83.6] \\
& KiP & 78.1 [71.4, 84.7] & 62.1 [55.3, 68.9] & 93.0 [90.0, 95.5] & 68.0 [61.6, 73.6] \\
\midrule
\multirow{5}{*}{14}
& Private & 95.3 [92.8, 97.6] & 85.6 [81.2, 90.0] & 98.0 [96.6, 99.1] & 89.6 [85.6, 93.2] \\
& AAST$_{\mathrm{M}}$ & \textbf{68.5} [62.2, 75.1] & \textbf{51.6} [45.2, 57.6] & \underline{64.4} [57.2, 71.0] & \textbf{52.8} [46.4, 58.8] \\
& AAST$_{\Phi}$ & \underline{70.8} [64.5, 77.2] & \underline{55.6} [48.8, 61.6] & \textbf{64.0} [56.7, 70.8] & \underline{53.6} [47.2, 59.6] \\
& JAMDEC & 88.8 [83.6, 93.3] & 73.2 [66.3, 78.9] & 92.8 [89.8, 95.8] & 79.6 [74.4, 84.4] \\
& KiP & 90.6 [86.3, 94.8] & 66.3 [58.9, 73.2] & 96.4 [94.4, 98.1] & 70.8 [64.8, 76.4] \\
\midrule
\multirow{5}{*}{16}
& Private & 96.1 [93.7, 98.2] & 87.2 [83.2, 91.2] & 98.5 [97.5, 99.4] & 90.0 [86.0, 93.6] \\
& AAST$_{\mathrm{M}}$ & \textbf{71.6} [65.0, 77.6] & \textbf{52.4} [46.0, 58.4] & \underline{72.1} [65.4, 77.9] & \underline{55.2} [48.4, 61.2] \\
& AAST$_{\Phi}$ & \underline{75.6} [69.2, 81.1] & \underline{56.3} [50.0, 62.4] & \textbf{68.6} [61.7, 75.1] & \textbf{54.8} [48.4, 60.4] \\
& JAMDEC & 90.1 [85.3, 94.2] & 75.3 [68.7, 81.3] & 94.0 [91.2, 96.6] & 84.0 [79.2, 88.4] \\
& KiP & 84.2 [72.9, 93.3] & 67.0 [55.6, 78.4] & 96.1 [93.8, 98.0] & 74.4 [68.8, 79.6] \\
\bottomrule
\end{tabular}
\caption{Same-genre verification results under a classical stylometric transfer attack on Reddit and Blog. The verifier uses character 3--5-gram term frequency--inverse document frequency features with logistic regression. Values are percentages with 95\% bootstrap confidence intervals.}
\label{tab:stylometric-verifi}
\end{table*}

\section{Additional Same-genre Utility Results}
\label{appendix-same-genre-utility}

\subsection{Additional Reddit Utility Results}
\label{appendix-reddit-utility}

Table~\ref{tab:utility-reddit-blog} reports the NLI and CoLA results on Reddit, while Table~\ref{tab:aast-sentiment-alignment} provides the sentiment alignment results.

\begin{table}[t]
    \centering
    {
    \fontsize{6.5}{7.5}\selectfont
    \begin{tabular}{
        >{\centering\arraybackslash}p{0.4cm}
        >{\centering\arraybackslash}p{1.0cm}
        |
        >{\centering\arraybackslash}p{0.7cm}
        >{\centering\arraybackslash}p{0.7cm}
        ||
        >{\centering\arraybackslash}p{0.7cm}
        >{\centering\arraybackslash}p{0.7cm}
    }
        \toprule
        \multirow{2}{*}{\textbf{$K$}} 
        & \multirow{2}{*}{\textbf{Method}}
        & \multicolumn{2}{c|}{\textbf{Reddit}}
        & \multicolumn{2}{c}{\textbf{Blog}} \\
        \cmidrule(lr){3-4}\cmidrule(lr){5-6}
        & & \textbf{NLI}($\uparrow$) & \textbf{CoLA}($\uparrow$)
          & \textbf{NLI}($\uparrow$) & \textbf{CoLA}($\uparrow$) \\
        \midrule

        \multirow{4}{*}{1}
        & AAST$_{\Phi}$        & \uline{55.1} & \textbf{94.7} & \uline{55.0} & \textbf{93.6} \\
        & AAST$_{\mathrm{M}}$  & 53.4 & \uline{93.1} & 51.6 & \uline{93.4} \\
        & JAMDEC               & 20.2 & 91.4 & 20.0 & 90.0 \\
        & KiP                  & \textbf{65.4} & 63.2 & \textbf{58.1} & 55.8 \\
        \cmidrule{1-6}

        \multirow{4}{*}{2}
        & AAST$_{\Phi}$        & \uline{57.2} & \textbf{94.9} & \uline{57.4} & \textbf{93.9} \\
        & AAST$_{\mathrm{M}}$  & 56.4 & \uline{94.1} & 54.5 & \uline{93.0} \\
        & JAMDEC               & 22.7 & 92.3 & 26.9 & 91.8 \\
        & KiP                  & \textbf{65.5} & 61.9 & \textbf{59.4} & 56.0 \\
        \cmidrule{1-6}

        \multirow{4}{*}{4}
        & AAST$_{\Phi}$        & \uline{61.3} & \textbf{94.8} & \textbf{62.5} & \textbf{93.8} \\
        & AAST$_{\mathrm{M}}$  & 60.5 & \uline{94.2} & \uline{61.5} & \uline{92.9} \\
        & JAMDEC               & 28.8 & 92.2 & 35.0 & 92.0 \\
        & KiP                  & \textbf{65.9} & 61.7 & \textbf{62.5} & 55.8 \\
        \cmidrule{1-6}

        \multirow{4}{*}{6}
        & AAST$_{\Phi}$        & \uline{64.2} & \textbf{94.9} & \textbf{66.1} & \textbf{93.8} \\
        & AAST$_{\mathrm{M}}$  & 62.9 & \uline{94.2} & 64.4 & \uline{93.1} \\
        & JAMDEC               & 34.3 & 92.2 & 38.7 & 85.6 \\
        & KiP                  & \textbf{68.2} & 61.2 & \uline{66.0} & 56.0 \\
        \cmidrule{1-6}

        \multirow{4}{*}{8}
        & AAST$_{\Phi}$        & \uline{65.9} & \textbf{94.8} & \textbf{68.9} & \textbf{93.8} \\
        & AAST$_{\mathrm{M}}$  & 65.7 & \uline{94.1} & 67.5 & \uline{92.9} \\
        & JAMDEC               & 37.1 & 92.2 & 45.4 & 86.1 \\
        & KiP                  & \textbf{69.1} & 61.1 & \uline{68.3} & 56.1 \\
        \cmidrule{1-6}

        \multirow{4}{*}{10}
        & AAST$_{\Phi}$        & \uline{68.6} & \textbf{94.7} & \textbf{71.2} & \textbf{93.7} \\
        & AAST$_{\mathrm{M}}$  & 67.7 & \uline{94.1} & \uline{70.2} & \uline{92.8} \\
        & JAMDEC               & 40.6 & 91.9 & 44.7 & 81.9 \\
        & KiP                  & \textbf{70.9} & 61.3 & 69.9 & 56.1 \\
        \cmidrule{1-6}

        \multirow{4}{*}{12}
        & AAST$_{\Phi}$        & \uline{70.0} & \textbf{94.7} & \textbf{72.8} & \textbf{93.8} \\
        & AAST$_{\mathrm{M}}$  & 69.7 & \uline{94.1} & \uline{72.0} & \uline{93.0} \\
        & JAMDEC               & 43.3 & 89.9 & 47.8 & 82.8 \\
        & KiP                  & \textbf{72.3} & 61.0 & 71.2 & 56.0 \\
        \cmidrule{1-6}

        \multirow{4}{*}{14}
        & AAST$_{\Phi}$        & \uline{71.4} & \textbf{94.8} & \textbf{74.5} & \textbf{93.8} \\
        & AAST$_{\mathrm{M}}$  & 71.1 & \uline{94.0} & \uline{73.5} & \uline{93.0} \\
        & JAMDEC               & 43.3 & 90.4 & 51.1 & 82.9 \\
        & KiP                  & \textbf{73.4} & 60.1 & 72.6 & 55.9 \\
        \cmidrule{1-6}

        \multirow{4}{*}{16}
        & AAST$_{\Phi}$        & \textbf{72.9} & \textbf{94.7} & \textbf{75.7} & \textbf{93.8} \\
        & AAST$_{\mathrm{M}}$  & 72.6 & \uline{94.0} & \uline{74.7} & \uline{92.9} \\
        & JAMDEC               & 45.8 & 86.8 & 51.6 & 82.7 \\
        & KiP                  & \uline{72.8} & 60.9 & 73.7 & 56.0 \\

        \bottomrule
    \end{tabular}
    }
    \caption{NLI and CoLA results on Reddit and Blog across bundle sizes.}
    \label{tab:utility-reddit-blog}
\end{table}

\begin{table}[t]
    \centering
    {
    \fontsize{6.5}{7.5}\selectfont
    \begin{tabular}{
        >{\centering\arraybackslash}p{0.4cm}
        >{\centering\arraybackslash}p{1.0cm}
        |
        >{\centering\arraybackslash}p{1.0cm}
        >{\centering\arraybackslash}p{1.0cm}
    }
        \toprule
        \multirow{2}{*}{\textbf{$K$}}
        & \multirow{2}{*}{\textbf{Method}}
        & \multicolumn{2}{c}{\textbf{Sentiment Alignment($\%$)}($\uparrow$)}\\
        \cmidrule(lr){3-4}
        & & \textbf{Reddit} & \textbf{Blog} \\
        \midrule

        \multirow{2}{*}{1}
        & AAST$_{\Phi}$       & 83.0 & 83.6 \\
        & AAST$_{\mathrm{M}}$ & 83.8 & 87.8 \\
        \cmidrule{1-4}

        \multirow{2}{*}{2}
        & AAST$_{\Phi}$       & 83.7 & 85.7 \\
        & AAST$_{\mathrm{M}}$ & 84.7 & 87.6 \\
        \cmidrule{1-4}

        \multirow{2}{*}{4}
        & AAST$_{\Phi}$       & 85.3 & 85.0 \\
        & AAST$_{\mathrm{M}}$ & 85.0 & 88.3 \\
        \cmidrule{1-4}

        \multirow{2}{*}{6}
        & AAST$_{\Phi}$       & 85.1 & 84.3 \\
        & AAST$_{\mathrm{M}}$ & 85.4 & 88.3 \\
        \cmidrule{1-4}

        \multirow{2}{*}{8}
        & AAST$_{\Phi}$       & 84.8 & 85.2 \\
        & AAST$_{\mathrm{M}}$ & 85.1 & 87.9 \\
        \cmidrule{1-4}

        \multirow{2}{*}{10}
        & AAST$_{\Phi}$       & 84.9 & 84.7 \\
        & AAST$_{\mathrm{M}}$ & 85.4 & 88.1 \\
        \cmidrule{1-4}

        \multirow{2}{*}{12}
        & AAST$_{\Phi}$       & 84.9 & 84.9 \\
        & AAST$_{\mathrm{M}}$ & 85.5 & 87.5 \\
        \cmidrule{1-4}

        \multirow{2}{*}{14}
        & AAST$_{\Phi}$       & 85.4 & 85.0 \\
        & AAST$_{\mathrm{M}}$ & 84.7 & 88.5 \\
        \cmidrule{1-4}

        \multirow{2}{*}{16}
        & AAST$_{\Phi}$       & 85.3 & 85.3 \\
        & AAST$_{\mathrm{M}}$ & 84.9 & 88.1 \\

        \bottomrule
    \end{tabular}
    }
    \caption{Sentiment alignment results on Reddit and Blog across bundle sizes. Higher sentiment alignment indicates better preservation of private text sentiment.}
    \label{tab:aast-sentiment-alignment}
\end{table}

\subsection{Blog Utility Results}
\label{appendix-blog-utility}

AAST also maintains strong NLI and CoLA on Blog, as reported in Figure~\ref{fig:blog_utility} and Table~\ref{tab:utility-reddit-blog}. AAST$_{\Phi}$ gives the best CoLA score across all bundle sizes, and its NLI becomes the strongest or tied for the strongest from $K$=4 onward. KiP has competitive NLI at smaller bundle sizes, but its CoLA scores are much lower and its linkability remains high. JAMDEC has lower NLI than AAST across all bundle sizes. Together with the privacy results, these utility scores show that AAST preserves content and linguistic quality while keeping authorship risk lower under aggregation.

AAST also preserves sentiment well on Blog. Figure~\ref{fig:utili_senti} and Table~\ref{tab:aast-sentiment-alignment} show that AAST$_{\mathrm{M}}$ and AAST$_{\Phi}$ maintain stable sentiment alignment across bundle sizes. AAST$_{\mathrm{M}}$ is especially stable, staying around 88\%, while AAST$_{\Phi}$ stays around 85\% on average. This suggests that AAST largely preserves the sentiment of the original blog entries while reducing authorship linkability.

\section{Additional LLM Baseline Comparison}
\label{appendix-llm-baseline-comparison}

Table~\ref{tab:rupta-comparison} compares AAST with RUPTA, an additional LLM-based text anonymization baseline, on Reddit and Blog across $K$=1,8,16. RUPTA$_{\Phi}$ and RUPTA$_{\mathrm{M}}$ denote RUPTA using Phi-4 and Mistral-Small as the generator, respectively. Utility is not reported for Private.

The results show that RUPTA can preserve semantic utility, but it does not consistently reduce account-level authorship risk under aggregation. 
The higher NLI of some RUPTA variants should be interpreted together with their privacy scores: staying closer to the private text can preserve semantic overlap, but can also preserve authorial cues and increase linkability, similar to the pattern observed for KiP. AAST does not simply maximize NLI; it provides a stronger privacy–utility balance by keeping attribution and verification risk much lower while maintaining strong linguistic quality and reasonable semantic preservation. As $K$ grows, both RUPTA variants show high attribution risk, often approaching Private or KiP. In contrast, AAST maintains substantially lower R@8 and MRR across both datasets, especially at $K=8$ and $K=16$. A similar pattern appears in verification. AAST keeps AUC closer to 50 and c@1 lower than RUPTA in most large-bundle settings. 

These results suggest that strong LLM-based rewriting alone is not sufficient for aggregation-level privacy. The bundle-level design of AAST is important for reducing linkability across a released account history.

\begin{table*}[t]
    \centering
    {
    \fontsize{7.0}{8.5}\selectfont
    \setlength{\tabcolsep}{1.4pt}
    \begin{tabular}{
        >{\centering\arraybackslash}p{0.8cm}
        >{\raggedright\arraybackslash}p{1.1cm}
        |
        >{\centering\arraybackslash}p{0.9cm}
        >{\centering\arraybackslash}p{1.0cm}
        |
        >{\centering\arraybackslash}p{1.1cm}
        >{\centering\arraybackslash}p{1.0cm}
        |
        >{\centering\arraybackslash}p{0.9cm}
        >{\centering\arraybackslash}p{1.0cm}
        ||
        >{\centering\arraybackslash}p{0.9cm}
        >{\centering\arraybackslash}p{1.0cm}
        |
        >{\centering\arraybackslash}p{1.1cm}
        >{\centering\arraybackslash}p{1.0cm}
        |
        >{\centering\arraybackslash}p{0.9cm}
        >{\centering\arraybackslash}p{1.0cm}
    }
        \toprule
        \multirow{3}{*}{\textbf{$K$}}
        & \multirow{3}{*}{\textbf{Method}}
        & \multicolumn{6}{c||}{\textbf{Reddit}}
        & \multicolumn{6}{c}{\textbf{Blog}} \\
        \cmidrule(lr){3-8}\cmidrule(lr){9-14}
        & & \multicolumn{2}{c|}{\textbf{Attribution}}
          & \multicolumn{2}{c|}{\textbf{Verification}}
          & \multicolumn{2}{c||}{\textbf{Utility}}
          & \multicolumn{2}{c|}{\textbf{Attribution}}
          & \multicolumn{2}{c|}{\textbf{Verification}}
          & \multicolumn{2}{c}{\textbf{Utility}} \\
        \cmidrule(lr){3-4}\cmidrule(lr){5-6}\cmidrule(lr){7-8}
        \cmidrule(lr){9-10}\cmidrule(lr){11-12}\cmidrule(lr){13-14}
        & & \textbf{R@8}($\downarrow$) & \textbf{MRR}($\downarrow$)
          & \textbf{AUC}($\approx$50) & \textbf{c@1}($\downarrow$)
          & \textbf{NLI}($\uparrow$) & \textbf{CoLA}($\uparrow$)
          & \textbf{R@8}($\downarrow$) & \textbf{MRR}($\downarrow$)
          & \textbf{AUC}($\approx$50) & \textbf{c@1}($\downarrow$)
          & \textbf{NLI}($\uparrow$) & \textbf{CoLA}($\uparrow$) \\
        \midrule

        \multirow{7}{*}{1}
        & Private & 45.6 & 32.2 & 51.9 & 52.9 & -- & -- & 52.4 & 38.5 & 64.6 & 60.6 & -- & -- \\
        & AAST$_{\Phi}$ & \uline{3.6} & \uline{2.4} & \textbf{50.6} & 51.7 & 55.1 & \textbf{94.7} & \textbf{2.0} & \textbf{1.7} & \textbf{48.3} & \textbf{48.6} & 55.0 & \textbf{93.6} \\
        & AAST$_{\mathrm{M}}$ & \textbf{2.9} & \textbf{2.2} & 48.9 & \textbf{44.6} & 53.4 & \uline{93.1} & \uline{2.4} & \uline{2.6} & \uline{47.6} & \uline{49.3} & 51.6 & \uline{93.4} \\
        & RUPTA$_{\Phi}$ & 8.4 & 14.5 & \uline{51.0} & \uline{47.9} & 62.1 & 85.8 & 26.4 & 16.8 & 59.6 & 57.6 & \uline{60.5} & 84.5 \\
        & RUPTA$_{\mathrm{M}}$ & 40.6 & 25.0 & 52.2 & 52.3 & \textbf{71.1} & 79.7 & 48.8 & 34.9 & 62.4 & 59.6 & \textbf{76.0} & 77.8 \\
        & KiP & 17.6 & 9.4 & 51.3 & 52.6 & \uline{65.4} & 63.2 & 12.8 & 8.2 & 63.2 & 57.7 & 58.1 & 55.8 \\
        & JAMDEC & 8.4 & 4.5 & 51.2 & 52.0 & 20.2 & 91.4 & 12.8 & 7.4 & 60.8 & 53.4 & 20.0 & 90.0 \\
        \cmidrule{1-14}

        \multirow{7}{*}{8}
        & Private & 100 & 97.8 & 78.8 & 71.7 & -- & -- & 98.0 & 93.8 & 79.3 & 72.9 & -- & -- \\
        & AAST$_{\Phi}$ & \uline{31.8} & \textbf{17.0} & \uline{65.1} & \uline{48.3} & 65.9 & \textbf{94.8} & \textbf{21.3} & \textbf{14.9} & 69.9 & \uline{48.4} & 68.9 & \textbf{93.8} \\
        & AAST$_{\mathrm{M}}$ & \textbf{28.4} & \uline{18.4} & \textbf{64.2} & \textbf{48.2} & 65.7 & \uline{94.1} & \uline{40.3} & \uline{24.9} & \textbf{62.1} & \textbf{48.2} & 67.5 & \uline{92.9} \\
        & RUPTA$_{\Phi}$ & 93.6 & 81.1 & 76.2 & 68.9 & \uline{70.1} & 85.6 & 91.6 & 79.2 & 74.9 & 67.3 & \uline{70.5} & 84.8 \\
        & RUPTA$_{\mathrm{M}}$ & 99.2 & 94.6 & 71.8 & 65.5 & \textbf{82.3} & 79.4 & 98.0 & 93.1 & 77.5 & 69.1 & \textbf{83.3} & 78.1 \\
        & KiP & 86.4 & 68.0 & 77.4 & 64.2 & 69.1 & 61.1 & 73.2 & 60.3 & 75.5 & 59.0 & 68.3 & 56.1 \\
        & JAMDEC & 41.6 & 24.4 & 77.0 & 52.0 & 37.1 & 92.2 & 52.0 & 33.0 & \uline{65.3} & 57.7 & 45.4 & 86.1 \\
        \cmidrule{1-14}

        \multirow{7}{*}{16}
        & Private & 100 & 99.8 & 89.7 & 83.6 & -- & -- & 99.2 & 97.5 & 82.3 & 79.1 & -- & -- \\
        & AAST$_{\Phi}$ & \textbf{60.7} & \textbf{39.8} & \uline{66.7} & \uline{49.8} & 72.9 & \textbf{94.7} & \textbf{61.4} & \textbf{41.1} & \uline{71.2} & \uline{53.2} & \uline{75.7} & \textbf{93.8} \\
        & AAST$_{\mathrm{M}}$ & \uline{63.7} & \uline{43.5} & \textbf{64.1} & \textbf{48.0} & 72.6 & \uline{94.0} & \uline{65.7} & \uline{44.5} & \textbf{68.6} & \textbf{50.3} & 74.7 & \uline{92.9} \\
        & RUPTA$_{\Phi}$ & 98.0 & 94.3 & 86.3 & 79.5 & \uline{74.9} & 85.7 & 98.0 & 95.4 & 82.4 & 78.1 & 74.8 & 85.0 \\
        & RUPTA$_{\mathrm{M}}$ & 100 & 98.0 & 81.9 & 74.5 & \textbf{86.1} & 79.2 & 99.4 & 98.2 & 86.2 & 83.0 & \textbf{83.1} & 78.3 \\
        & KiP & 97.8 & 92.7 & 92.6 & 86.1 & 72.8 & 60.9 & 94.0 & 83.5 & 85.3 & 68.1 & 73.7 & 56.0 \\
        & JAMDEC & 74.8 & 54.7 & 81.4 & 73.1 & 45.8 & 86.8 & 83.6 & 62.4 & 73.3 & 61.4 & 51.6 & 82.7 \\

        \bottomrule
    \end{tabular}
    }
    \caption{Comparison with the LLM-based RUPTA baseline on Reddit and Blog across $K$=1,8,16. Utility is not reported for Private.}
    \label{tab:rupta-comparison}
\end{table*}

\section{Additional Cross-genre Verification Results}
\label{appendix-cross-genre-verification}

The verification results in Table~\ref{tab:crossgenre-verification} follow the same cross-genre trend, with AAST lowering the verifier's ability to link synthetic and reference texts. In Tweet--Article, AAST$_{\mathrm{M}}$ gives the lowest F1 under both SELMA + TaskOnly and SELMA + PromptAV. Its binary verification Accuracy is also the best or second best across the two attacks. The reduction is clearer in Article--Tweet. AAST$_{\mathrm{M}}$ gives the lowest Accuracy and F1 under both attacks, with especially low F1 under SELMA + PromptAV. These results show that AAST reduces not only cross-genre attribution risk, but also the verifier's ability to decide whether synthetic text and reference text come from the same author.

The Article--Tweet setting is harder for AAST because tweets often contain short reactions, handles, hashtags, and URLs. We discuss this case in Appendix~\ref{appendix-discuss}.

\begin{table}
    \centering  
    \fontsize{6.5}{7.5}\selectfont
        \begin{tabular}{
        >{\centering\arraybackslash}p{1.5cm}
        >{\centering\arraybackslash}p{0.9cm}
        >{\centering\arraybackslash}p{0.9cm}
        |
        >{\centering\arraybackslash}p{0.9cm}
        >{\centering\arraybackslash}p{0.9cm}
    }
        \toprule
        \textbf{Genre Pair} & \textbf{Attack} & \textbf{Method}
        & \textbf{Acc.}($\downarrow$)
        & \textbf{F1}($\downarrow$) \\
        \midrule

        \multirow{8}{*}{\shortstack{Tweet--Article}}
        & \multirow{4}{*}{\shortstack{SELMA +\\TaskOnly}}
        & AAST$_{\mathrm{M}}$ & \textbf{54.8} & \textbf{54.3} \\
        & & AAST$_{\Phi}$ & 55.6 & \uline{53.6} \\
        & & KiP & 58.2 & 56.8 \\
        & & JAMDEC & \uline{55.0} & 62.9 \\
        \cmidrule{2-5}

        & \multirow{4}{*}{\shortstack{SELMA +\\PromptAV}}
        & AAST$_{\mathrm{M}}$ & \uline{55.2} & \textbf{53.9} \\
        & & AAST$_{\Phi}$ & 55.9 & \uline{56.2} \\
        & & KiP & 58.4 & 59.2 \\
        & & JAMDEC & \textbf{55.0} & 62.5 \\        
        \midrule \midrule
        \multirow{8}{*}{\shortstack{Article--Tweet}}
        & \multirow{4}{*}{\shortstack{SELMA +\\TaskOnly}}
        & AAST$_{\mathrm{M}}$ & \textbf{51.7} & \textbf{41.7} \\
        & & AAST$_{\Phi}$ & 53.7 & \uline{45.5} \\
        & & KiP & 56.3 & 47.8 \\
        & & JAMDEC & \uline{53.3} & 50.8 \\
        \cmidrule{2-5}

        & \multirow{4}{*}{\shortstack{SELMA +\\PromptAV}}
        & AAST$_{\mathrm{M}}$ & \textbf{51.9} & \textbf{34.1} \\
        & & AAST$_{\Phi}$ & 53.6 & \uline{42.5} \\
        & & KiP & 56.3 & 47.5 \\
        & & JAMDEC & \uline{53.1} & 50.2 \\

        \bottomrule
    \end{tabular}
    \caption{Cross-genre authorship verification risk results.}
    \label{tab:crossgenre-verification}
\end{table}

\section{Cross-genre Utility Results}
\label{appendix-cross-genre-utility}

Table~\ref{tab:crossgenre-utility} shows that in both Tweet--Article and Article--Tweet, AAST$_{\mathrm{M}}$ achieves the highest CoLA score, indicating stronger linguistic quality than the baselines. AAST$_{\Phi}$ is consistently second best on CoLA, while KiP has much lower CoLA despite its higher NLI.

The cross-genre setting also shows a clear privacy utility tradeoff. KiP preserves more semantic content according to NLI, but its privacy results are weaker in both attribution and verification. JAMDEC has lower NLI and lower CoLA than AAST in both genre pairs. AAST$_{\mathrm{M}}$ gives the strongest balance in this setting. It substantially reduces cross-genre linkability while maintaining the best linguistic quality. Its lower NLI suggests that preserving exact semantic entailment across genre rewriting remains challenging, especially when converting between short tweets and longer articles.

\begin{table}[t]
    \centering
    {
    \fontsize{6.5}{8}\selectfont
    \begin{tabular}{
        >{\centering\arraybackslash}p{1.2cm}
        >{\centering\arraybackslash}p{0.9cm}
        |
        >{\centering\arraybackslash}p{0.9cm}
        >{\centering\arraybackslash}p{0.9cm}
    }
        \toprule
        \textbf{Genre Pair} & \textbf{Method}
        & \textbf{NLI}($\uparrow$)
        & \textbf{CoLA}($\uparrow$) \\
        \midrule

        \multirow{4}{*}{\shortstack{Tweet\\--\\Article}}
        & AAST$_{\mathrm{M}}$ & 27.6 & \textbf{89.6} \\
        & AAST$_{\Phi}$ & \uline{30.9} & \uline{83.2} \\
        & KiP & \textbf{53.8} & 53.9 \\
        & JAMDEC & 18.3 & 71.4 \\        
        \midrule \midrule
        \multirow{4}{*}{\shortstack{Article\\--\\Tweet}}
        & AAST$_{\mathrm{M}}$ & 14.9 & \textbf{86} \\
        & AAST$_{\Phi}$ & \uline{18.0} & \uline{80.3} \\
        & KiP & \textbf{32.8} & 47.7 \\
        & JAMDEC & 10.8 & 59.5 \\

        \bottomrule
    \end{tabular}
    }
    \caption{Cross-genre NLI and CoLA results.}
    \label{tab:crossgenre-utility}
\end{table}

\section{Discussion of the Article--Tweet Setting}
\label{appendix-discuss}

Table~\ref{tab:masked-tweet-examples} shows masked tweet inputs from one author. These examples illustrate a difficult case for bundle-level selection. Many tweet inputs contain little text beyond handles, URLs, hashtags, and other surface markers. After these markers are removed or normalized, the remaining text can be very short or generic, such as brief reactions or replies with only one word. Thus, there is less semantic and stylistic content for AAST to preserve and less variation for the bundle-level objective to exploit.

This pattern also helps explain why Article--Tweet differs from Tweet--Article in authorship attribution and verification risk. In Tweet--Article, AAST$_{\mathrm{M}}$ and AAST$_{\Phi}$ are usually the best two methods, showing that AAST has more room to preserve meaning and reduce author signals when generating longer article outputs. In Article--Tweet, AAST$_{\mathrm{M}}$ remains the strongest method, but AAST$_{\Phi}$ is less consistently the second best. This suggests that short tweet inputs with heavy metadata leave less room for AAST's bundle-level selection to balance privacy and utility. After handles, URLs, hashtags, and brief reactions are normalized, many tweet inputs contain limited text. Thus, the bundle-level signal can become weak or noisy, and AAST then relies more heavily on Candidate Preselection. The point is not that AAST fails in Article--Tweet. AAST$_{\mathrm{M}}$ still gives the strongest cross-genre privacy results in this setting. 

\begin{table}[t]
\centering
\fontsize{9}{11.5}\selectfont
\begin{tabular}{p{0.3cm}p{6.4cm}}
\toprule
\textbf{ID} & \textbf{Masked Tweet Example} \\
\midrule
1 & \texttt{@[USER1] @[USER2] @[USER3] @[USER4] @[USER5] @[USER6] Congratulations to all fellow winners} \\
2 & \texttt{\#NewProfilePic [URL]} \\
3 & \texttt{@[USER7] No} \\
4 & \texttt{@[USER8] haha} \\
\bottomrule
\end{tabular}
\caption{Masked examples of tweet inputs from one author. Handles and URLs are replaced with placeholders.}
\label{tab:masked-tweet-examples}
\end{table}

\section{Ablation Analysis}
\label{appendix-ablation-analysis}

Our ablation reports two analyses. First, Appendix~\ref{appendix-ablation-components} studies the effects of Identifier Masking, Candidate Preselection, Candidate Refinement, and Bundle-Level Selection using AAST$_{\mathrm{M}}$ on Reddit across \(K=1,8,16\). Second, Appendix~\ref{appendix-ablation-bundlesel} adds a controlled NoBundleSel$_{\Phi}$ comparison on Reddit and Blog to isolate the role of bundle-level selection under the main same-genre setting.

\subsection{Component Effects}
\label{appendix-ablation-components}
\subsubsection{Effect of Identifier Masking}

Identifier Masking (\S\ref{subsec:preprocessing}) normalizes structured surface markers such as handles, URLs, hashtags, emails, numbers, emojis, and repeated punctuation. These markers do not always identify an author by themselves, but they can create direct lexical shortcuts or platform artifacts that affect generation and evaluation. In Table~\ref{tab:aast-ablation}, enabling Identifier Masking lowers privacy risk on Reddit, however, it also comes with lower utility. Since AAST is designed to generate useful synthetic text, not only to remove all privacy signals, this utility drop matters for downstream use where preserving meaning is important.

Based on this tradeoff, we apply Identifier Masking selectively. In cross-genre settings involving tweet-like text, structured surface markers are frequent and often carry limited semantic content, so we enable Identifier Masking. For same-genre Reddit and Blog, these surface forms may carry useful content, and masking them can reduce semantic preservation. Therefore, we disable Identifier Masking in the reported same-genre settings and treat it as a conditional preprocessing module for cross-genre tweet settings.

\begin{table}
    \centering  
    {
    \fontsize{7}{7.5}\selectfont 
    \begin{tabular}{
        >{\centering\arraybackslash}p{0.8cm} |
        >{\centering\arraybackslash}p{1.4cm} |
        >{\centering\arraybackslash}p{1.4cm} |
        >{\centering\arraybackslash}p{1.4cm}}
        \toprule

        \multirow{3}{*}{\textbf{$K$}} &  
        \multicolumn{3}{c}{\textbf{ $\sigma$}} \\
        \cmidrule(lr){2-4}
         & \textbf{$\epsilon$ = 4} & \textbf{$\epsilon$ = 2} & \textbf{$\epsilon$ = 1} \\

        \midrule       
        1 & 1.02 & 2.03 & 4.07 \\
        \midrule
        8 & 1.15 & 2.31 & 4.61 \\
        \midrule
        16 & 1.19 & 2.39 & 4.78 \\
        \bottomrule
    \end{tabular}
    \caption{Gaussian noise standard deviations $\sigma$ used for noisy abstraction selection.}
    \label{tab:noise_multipliers}
    }   
\end{table}

\begin{table}[t]
    \centering
    {
    \fontsize{8}{8.5}\selectfont
    \begin{tabular}{
        >{\centering\arraybackslash}p{2.0cm}
        >{\centering\arraybackslash}p{3.0cm}
    }
        \toprule
        \textbf{Bundle Size $K$}
        & \textbf{Total Tokens} \((\times 10^6)\) \\
        \midrule
        1  & 0.50 \\
        2  & 1.01 \\
        4  & 2.00 \\
        6  & 3.00 \\
        8  & 3.99 \\
        10 & 4.99 \\
        12 & 6.00 \\
        14 & 6.98 \\
        16 & 7.97 \\
        \bottomrule
    \end{tabular}
    }
    \caption{Token usage of AAST$_{\mathrm{M}}$ on Blog across bundle sizes. Counts include both input and output tokens for Mistral generation.}
    \label{tab:token-usage-blog}
\end{table}

\subsubsection{Effect of Candidate Preselection}

Candidate Preselection (\S\ref{subsec:candidate_preselect}) targets direct private--synthetic carryover. It compares each generated candidate with its corresponding private input and retains the candidates that are less similar to the private input before bundle-level selection.

The Reddit ablation in Table~\ref{tab:aast-ablation} measures synthetic bundle linkability, not direct private--synthetic similarity. Under this metric, disabling Candidate Preselection can lower attribution and verification risk. This happens because disabling Candidate Preselection leaves a larger candidate pool for Bundle-Level Selection, giving the bundle-level objective more freedom to choose candidates that reduce synthetic bundle linkability.

Thus, Candidate Preselection should not be interpreted as the main source of same-genre bundle-level privacy gains. Its role is different: it reduces the chance that a generated candidate remains too close to its private input. We keep Candidate Preselection in the main AAST pipeline because this protection is important for stronger direct comparison settings and for cross-genre evaluations, where private or reference texts may be compared against released synthetic texts.

\subsubsection{Effect of Candidate Refinement and Bundle-Level Selection}

Candidate Refinement (\S\ref{subsubsec:candidate_refinement}) and Bundle-Level Selection (\S\ref{subsec:bundle_level_selection}) are the main components for controlling authorship leakage. The setting without Candidate Refinement removes the refinement step after Bundle-level Candidate Scoring (\S\ref{subsubsec:candidate_scoring}), while the setting without Bundle-Level Selection removes both Bundle-level Candidate Scoring and Candidate Refinement. Removing Candidate Refinement sharply increases verification risk. At $K$=16, AUC rises from 51.3 to 78.5, and c@1 rises from 46.5 to 55.9. Removing Bundle-Level Selection causes an even larger degradation. At $K$=16, R@8 rises to 75.9, MRR rises to 55.9, AUC rises to 83.6, and c@1 rises to 75.7. Although these settings obtain higher utility scores, they leave strong bundle-level author signals. These results show that AAST needs Bundle-Level Selection over the generated pool, not only independent rewriting for each text.

In summary, the ablation shows that the modules address different risks. Identifier Masking normalizes structured surface markers when they are frequent and low in semantic content. Candidate Preselection reduces direct private--synthetic carryover. Bundle-Level Selection provides the main protection against bundle-level linkability as $K$ increases.

\begin{table*}[t]
    \centering
    {
    \fontsize{8}{9}\selectfont
    \begin{tabular}{
        >{\centering\arraybackslash}p{5.2cm}
        >{\centering\arraybackslash}p{0.8cm}
        |
        >{\centering\arraybackslash}p{1.0cm}
        >{\centering\arraybackslash}p{1.0cm}
        |
        >{\centering\arraybackslash}p{1.0cm}
        >{\centering\arraybackslash}p{1.0cm}
        |
        >{\centering\arraybackslash}p{1.0cm}
        >{\centering\arraybackslash}p{1.0cm}
    }
        \toprule
        \multirow{2}{*}{\textbf{Ablation configuration}}
        & \multirow{2}{*}{\textbf{$K$}}
        & \multicolumn{2}{c|}{\textbf{Attribution}}
        & \multicolumn{2}{c|}{\textbf{Verification}}
        & \multicolumn{2}{c}{\textbf{Utility}} \\
        \cmidrule(lr){3-4}\cmidrule(lr){5-6}\cmidrule(lr){7-8}
        & & \textbf{R@8}($\downarrow$) & \textbf{MRR}($\downarrow$)
          & \textbf{AUC}($\approx$50) & \textbf{c@1}($\downarrow$)
          & \textbf{NLI}($\uparrow$) & \textbf{CoLA}($\uparrow$) \\
        \midrule

        \multirow{3}{*}{\makecell{Identifier Masking(\S\ref{subsec:preprocessing}) enabled}}
        & 1  & 0  & 1.2  & 55.6 & 52.6 & 18.8 & 92.5 \\
        & 8  & 3.3  & 2.0  & 55.8 & 46.4 & 38.4 & 92.8 \\
        & 16 & 13.5 & 8.2  & 51.3 & 46.5 & 50.6 & 92.7 \\
        \cmidrule{1-8}

        \multirow{3}{*}{\makecell{Identifier Masking disabled}}
        & 1  & 2.9  & 2.2  & 48.9 & 44.6 & 53.4 & 93.1 \\
        & 8  & 28.4 & 18.4 & 64.2 & 48.2 & 65.7 & 94.1 \\
        & 16 & 63.7 & 43.5 & 64.1 & 48.0 & 72.6 & 94.0 \\
        \midrule \midrule

        \multirow{3}{*}{
        \makecell{Candidate Preselection(\S\ref{subsec:candidate_preselect}) disabled}}
        & 1  & 0  & 0.9 & 48.5 & 46.6 & 18.9 & 92.1 \\
        & 8  & 0.8  & 1.3 & 42.5 & 47.1 & 38.6 & 92.5 \\
        & 16 & 6.3  & 4.6 & 41.6 & 47.5 & 50.8 & 92.5 \\
        \cmidrule{1-8}

        \multirow{3}{*}{
        \makecell{Candidate Refinement(\S\ref{subsubsec:candidate_refinement}) disabled}}
        & 1  & 2.5  & 2.3  & 52.7 & 47.7 & 53.9 & 93.0 \\
        & 8  & 20.7 & 11.6 & 75.2 & 52.5 & 65.5 & 94.2 \\
        & 16 & 24.1 & 14.8 & 78.5 & 55.9 & 72.1 & 94.1 \\
        \cmidrule{1-8}

        \multirow{3}{*}{
        \makecell{Bundle-Level Selection(\S\ref{subsec:bundle_level_selection}) disabled}
        }
        & 1  & 10.7 & 7.5  & 53.4 & 50.7 & 52.2 & 93.1 \\
        & 8  & 57.9 & 39.1 & 75.8 & 69.4 & 74.6 & 94.2 \\
        & 16 & 75.9 & 55.9 & 83.6 & 75.7 & 71.7 & 94.1 \\

        \bottomrule
    \end{tabular}
    }
    \caption{Ablation study of AAST components on Reddit across \(K=1,8,16\). The first two row groups compare Identifier Masking configurations. The last three row groups disable one component relative to the Identifier Masking enabled configuration. In the reported main experiments, same-genre Reddit and Blog use Identifier Masking disabled, while cross-genre Article--Tweet and Tweet--Article use Identifier Masking enabled. Candidate Preselection, Refinement, and Bundle-Level Selection are enabled in all reported main AAST settings. Bundle-Level Selection consists of Bundle-level Candidate Scoring followed by Candidate Refinement.}
    \label{tab:aast-ablation}
\end{table*}

\subsection{Effect of Removing Bundle-Level Selection}
\label{appendix-ablation-bundlesel}

\begin{figure*}[t]
    \centering

    \begin{minipage}{0.9\linewidth}
        \centering
        \includegraphics[width=\linewidth]{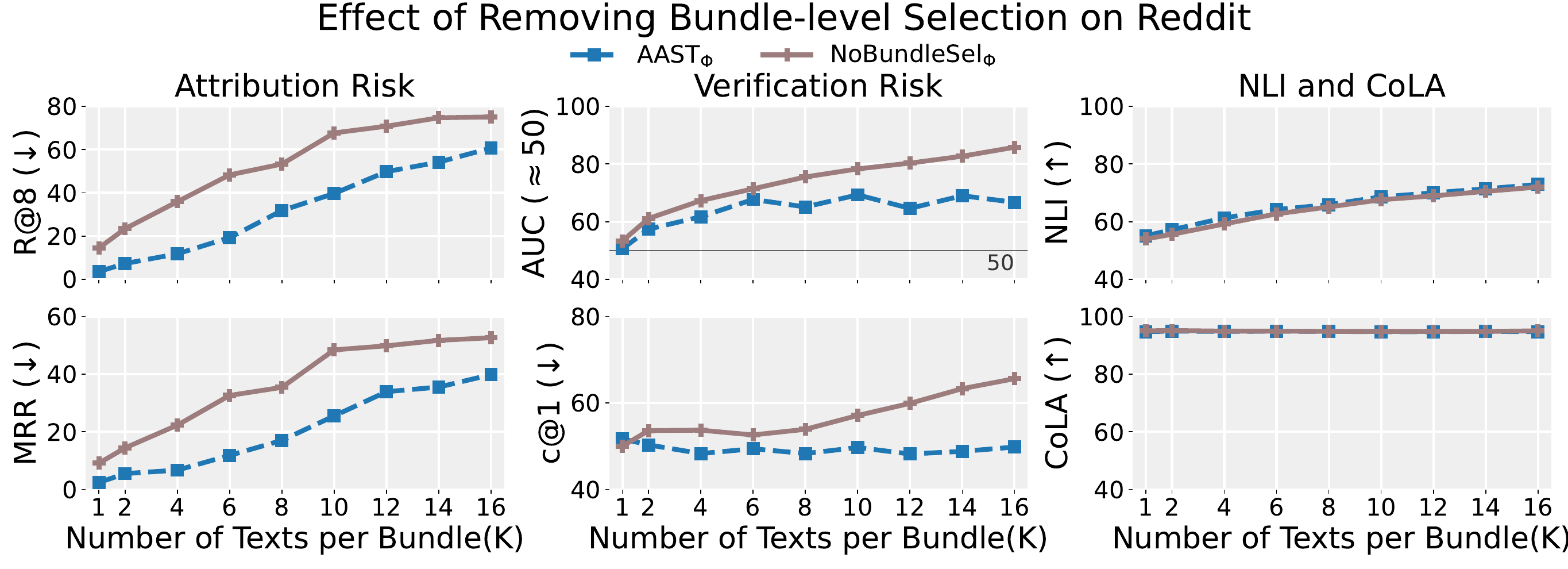}
    \end{minipage}

    \vspace{0.8em}

    \begin{minipage}{0.9\linewidth}
        \centering
        \includegraphics[width=\linewidth]{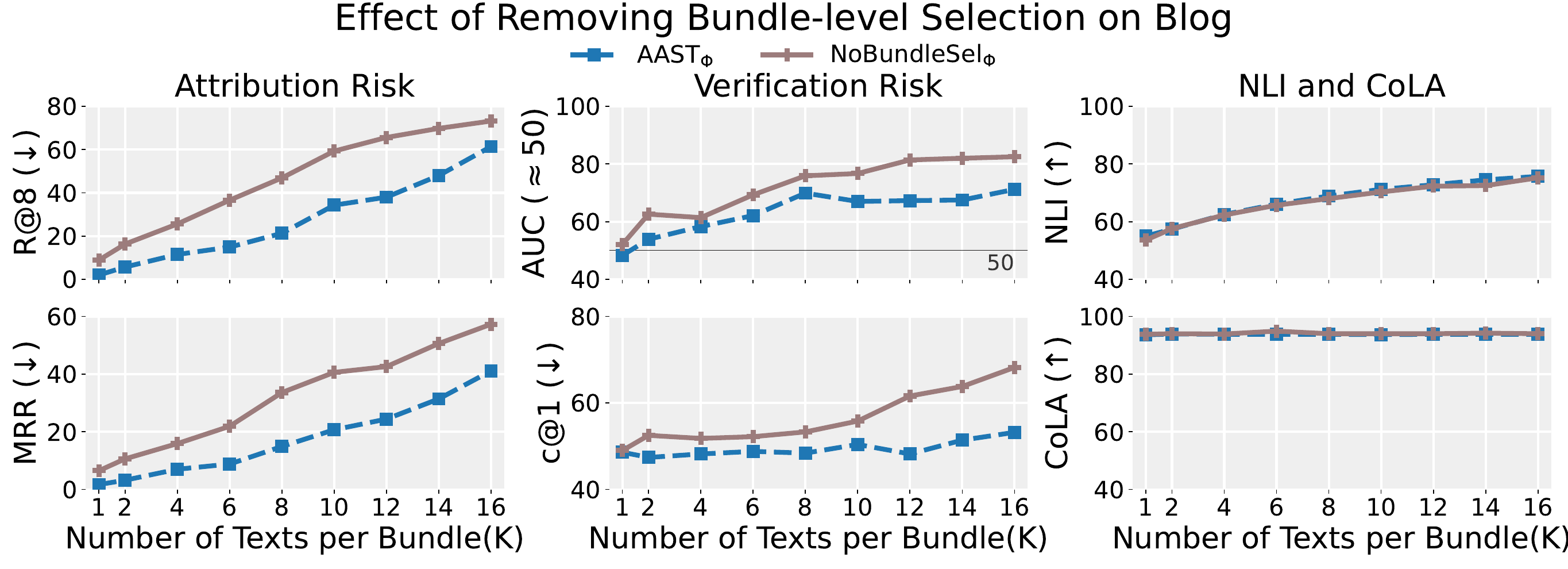}
    \end{minipage}

    \caption{Effect of removing bundle-level selection on Reddit and Blog. NoBundleSel$_{\Phi}$ keeps the same settings as AAST$_{\Phi}$, but removes Bundle-level Candidate Scoring and Candidate Refinement.}
    \label{fig:abla_no_bundle}
\end{figure*}

Figure~\ref{fig:abla_no_bundle} and Table~\ref{tab:abla_no_bundle} compare AAST$_{\Phi}$ with NoBundleSel$_{\Phi}$ on Reddit and Blog. NoBundleSel$_{\Phi}$ keeps the same settings as AAST$_{\Phi}$, but removes Bundle-level Candidate Scoring and Candidate Refinement. This controlled comparison tests whether the privacy gains come from bundle-level selection, not only from the LLM generator, prompt, candidate pool, or abstraction model.

Removing bundle-level selection largely increases attribution risk as $K$ grows. On Reddit, NoBundleSel$_{\Phi}$ increases R@8 from 14.5 at $K$=1 to 75.1 at $K$=16, while AAST$_{\Phi}$ stays lower, from 3.6 to 60.7. The same pattern appears on Blog, where NoBundleSel$_{\Phi}$ reaches 73.2 R@8 at $K$=16, compared with 61.4 for AAST$_{\Phi}$. Verification risk also increases after removing bundle-level selection. At $K$=16, NoBundleSel$_{\Phi}$ reaches 85.8 AUC and 65.6 c@1 on Reddit, while AAST$_{\Phi}$ gives 66.7 AUC and 49.8 c@1. On Blog, NoBundleSel$_{\Phi}$ reaches 82.5 AUC and 68.2 c@1, compared with 71.2 AUC and 53.2 c@1 for AAST$_{\Phi}$.

In contrast, the utility scores remain close. Across both datasets, NLI and CoLA are similar between AAST$_{\Phi}$ and NoBundleSel$_{\Phi}$. This shows that bundle-level selection is the main source of the privacy gain, and that this gain does not come from a meaningful loss in semantic or linguistic utility.

The gap between AAST and NoBundleSel is not meant to show that bundle-level selection removes all aggregation risk, especially at the largest bundle sizes. Instead, the result should be read as a privacy--utility tradeoff. At $K$=16, AAST$_{\Phi}$ still preserves high utility while reducing attribution and verification risk relative to NoBundleSel$_{\Phi}$. This supports our design goal: AAST reduces account-level linkability without sacrificing semantic and linguistic quality.

\begin{table*}[t]
\centering
\scriptsize
\setlength{\tabcolsep}{3.0pt}
\renewcommand{\arraystretch}{1.05}
\resizebox{\textwidth}{!}{
\begin{tabular}{
>{\centering\arraybackslash}p{0.5cm}
>{\centering\arraybackslash}p{1.7cm}|
>{\centering\arraybackslash}p{0.9cm}
>{\centering\arraybackslash}p{0.9cm}|
>{\centering\arraybackslash}p{0.9cm}
>{\centering\arraybackslash}p{0.9cm}|
>{\centering\arraybackslash}p{0.9cm}
>{\centering\arraybackslash}p{0.9cm}|
>{\centering\arraybackslash}p{0.9cm}
>{\centering\arraybackslash}p{0.9cm}|
>{\centering\arraybackslash}p{0.9cm}
>{\centering\arraybackslash}p{0.9cm}|
>{\centering\arraybackslash}p{0.9cm}
>{\centering\arraybackslash}p{0.9cm}}
\toprule
\multirow{3}{*}{$K$} & \multirow{3}{*}{Method}
& \multicolumn{4}{c|}{Attribution}
& \multicolumn{4}{c|}{Verification}
& \multicolumn{4}{c}{NLI and CoLA} \\
\cmidrule(lr){3-6}\cmidrule(lr){7-10}\cmidrule(lr){11-14}
& & \multicolumn{2}{c|}{Reddit} & \multicolumn{2}{c|}{Blog}
& \multicolumn{2}{c|}{Reddit} & \multicolumn{2}{c|}{Blog}
& \multicolumn{2}{c|}{Reddit} & \multicolumn{2}{c}{Blog} \\
\cmidrule(lr){3-4}\cmidrule(lr){5-6}\cmidrule(lr){7-8}\cmidrule(lr){9-10}\cmidrule(lr){11-12}\cmidrule(lr){13-14}
& & MRR$\downarrow$ & R@8$\downarrow$ & MRR$\downarrow$ & R@8$\downarrow$
& AUC$\approx$50 & c@1$\downarrow$ & AUC$\approx$50 & c@1$\downarrow$
& NLI$\uparrow$ & CoLA$\uparrow$ & NLI$\uparrow$ & CoLA$\uparrow$ \\
\midrule

\multirow{2}{*}{1}
& AAST$_{\Phi}$ & 2.4 & 3.6 & 1.7 & 2.0 & 50.6 & 51.7 & 48.3 & 48.6 & 55.1 & 94.7 & 55.0 & 93.6 \\
& NoBundleSel$_{\Phi}$ & 9.1 & 14.5 & 6.5 & 8.9 & 53.3 & 50.0 & 52.1 & 49.0 & 54.0 & 95.0 & 53.6 & 93.8 \\

\midrule
\multirow{2}{*}{2}
& AAST$_{\Phi}$ & 5.5 & 7.3 & 3.2 & 5.7 & 57.4 & 50.3 & 53.9 & 47.4 & 57.2 & 94.9 & 57.4 & 93.9 \\
& NoBundleSel$_{\Phi}$ & 14.4 & 23.4 & 10.6 & 16.3 & 61.1 & 53.6 & 62.6 & 52.5 & 55.6 & 95.1 & 57.6 & 94.0 \\

\midrule
\multirow{2}{*}{4}
& AAST$_{\Phi}$ & 6.7 & 11.8 & 7.0 & 11.5 & 61.7 & 48.3 & 58.3 & 48.2 & 61.3 & 94.8 & 62.5 & 93.8 \\
& NoBundleSel$_{\Phi}$ & 22.3 & 35.9 & 15.9 & 25.6 & 67.3 & 53.7 & 61.4 & 51.8 & 59.2 & 94.9 & 62.3 & 93.9 \\

\midrule
\multirow{2}{*}{6}
& AAST$_{\Phi}$ & 11.8 & 19.3 & 8.8 & 14.9 & 67.7 & 49.4 & 62.1 & 48.8 & 64.2 & 94.9 & 66.1 & 93.8 \\
& NoBundleSel$_{\Phi}$ & 32.6 & 48.3 & 21.9 & 36.6 & 71.4 & 52.6 & 69.3 & 52.2 & 62.7 & 94.9 & 65.7 & 94.9 \\

\midrule
\multirow{2}{*}{8}
& AAST$_{\Phi}$ & 17.0 & 31.8 & 14.9 & 21.3 & 65.1 & 48.3 & 69.9 & 48.4 & 65.9 & 94.8 & 68.9 & 93.8 \\
& NoBundleSel$_{\Phi}$ & 35.4 & 53.3 & 33.6 & 46.9 & 75.5 & 53.9 & 75.9 & 53.3 & 65.1 & 94.8 & 68.0 & 94.0 \\

\midrule
\multirow{2}{*}{10}
& AAST$_{\Phi}$ & 25.5 & 39.7 & 20.7 & 34.3 & 69.3 & 49.7 & 67.0 & 50.4 & 68.6 & 94.7 & 71.2 & 93.7 \\
& NoBundleSel$_{\Phi}$ & 48.4 & 67.7 & 40.6 & 59.3 & 78.3 & 57.1 & 76.7 & 55.8 & 67.6 & 94.8 & 70.3 & 94.0 \\

\midrule
\multirow{2}{*}{12}
& AAST$_{\Phi}$ & 33.9 & 49.8 & 24.4 & 38.0 & 64.6 & 48.2 & 67.3 & 48.2 & 70.0 & 94.7 & 72.8 & 93.8 \\
& NoBundleSel$_{\Phi}$ & 49.8 & 70.8 & 42.6 & 65.6 & 80.3 & 59.9 & 81.4 & 61.6 & 68.9 & 94.8 & 72.3 & 94.0 \\

\midrule
\multirow{2}{*}{14}
& AAST$_{\Phi}$ & 35.5 & 54.2 & 31.4 & 48.0 & 69.0 & 48.8 & 67.5 & 51.4 & 71.4 & 94.8 & 74.5 & 93.8 \\
& NoBundleSel$_{\Phi}$ & 51.7 & 74.7 & 50.6 & 69.8 & 82.7 & 63.3 & 82.0 & 63.8 & 70.5 & 94.8 & 72.5 & 94.2 \\

\midrule
\multirow{2}{*}{16}
& AAST$_{\Phi}$ & 39.8 & 60.7 & 41.1 & 61.4 & 66.7 & 49.8 & 71.2 & 53.2 & 72.9 & 94.7 & 75.7 & 93.8 \\
& NoBundleSel$_{\Phi}$ & 52.6 & 75.1 & 57.3 & 73.2 & 85.8 & 65.6 & 82.5 & 68.2 & 72.0 & 95.0 & 75.2 & 94.0 \\

\bottomrule
\end{tabular}
}
\caption{Results for the effect of removing bundle-level selection on Reddit and Blog. NoBundleSel$_{\Phi}$ keeps the same settings as AAST$_{\Phi}$, but removes Bundle-level Candidate Scoring and Candidate Refinement.}
\label{tab:abla_no_bundle}
\end{table*}

\section{Effect of Noisy Selection}
\label{appendix-dp-evaluation}

We also study AAST$_{\Phi}$ with Gaussian noise added to the abstraction score selection step under \(\epsilon \in \{4,2,1\}\). The noise perturbs the normalized candidate scores before one abstraction candidate is selected. For each bundle size \(K\), we set \(\delta = 1/(N\log N)\), where \(N\) is the number of private texts entering the noisy selection step. In our setting, each experiment uses 250 users with two bundles per user and \(K\) texts per bundle, so \(N=250\times2\times K\). This gives \(N=500\), \(4000\), and \(8000\) for \(K=1,8,16\), respectively. Table~\ref{tab:noise_multipliers} reports the resulting Gaussian noise standard deviations \(\sigma\).

Table~\ref{tab:dp-epsilon-results} shows how noisy abstraction selection affects AAST$_{\Phi}$. The effect becomes clearer as aggregation size increases. At $K$=8 and $K$=16, stronger noise generally lowers attribution risk compared with \(\epsilon=\infty\), which denotes no noise. On Reddit, R@8 decreases from 31.8 to 22.7 at $K$=8 and from 60.7 to 53.4 at $K$=16. On Blog, R@8 decreases from 21.3 to 19.6 at $K$=8 and from 61.4 to 54.2 at $K$=16. MRR follows a similar pattern. Verification risk also tends to move closer to the desired range.

The noise introduces a utility tradeoff. The setting with no noise usually gives the highest NLI because it selects the abstraction candidate with the strongest semantic and sentiment score. As noise increases, the selected abstraction can move away from this top candidate.

These results show that noisy abstraction selection can reduce bundle-level linkability, especially at larger \(K\), with a small drop in semantic preservation. This analysis is limited to the abstraction score selection step and does not claim end to end DP for candidate generation or the final synthetic corpus.

\begin{table*}[t]
    \centering
    {
    \fontsize{7.5}{8.5}\selectfont
    \begin{tabular}{
        >{\centering\arraybackslash}p{1.0cm}
        >{\centering\arraybackslash}p{0.6cm}
        >{\centering\arraybackslash}p{1.2cm}
        >{\centering\arraybackslash}p{1.6cm}
        |
        >{\centering\arraybackslash}p{1.2cm}
        >{\centering\arraybackslash}p{1.2cm}
        >{\centering\arraybackslash}p{1.2cm}
        >{\centering\arraybackslash}p{1.2cm}
    }
        \toprule
        \textbf{Dataset} & \textbf{$K$} & \textbf{Evaluation} & \textbf{Metric}
        & \textbf{$\epsilon=\infty$}
        & \textbf{$\epsilon=4$}
        & \textbf{$\epsilon=2$}
        & \textbf{$\epsilon=1$} \\
        \midrule

        \multirow{25}{*}{Reddit}
        & \multirow{7}{*}{1}
        & \multirow{2}{*}{Attribution} & MRR($\downarrow$) & 2.4 & 3.1 & 2.7 & 3.0 \\
        & & & R@8($\downarrow$) & 3.6 & 4.0 & 2.8 & 3.2 \\
        \cmidrule(lr){3-8}
        & & \multirow{2}{*}{Verification} & AUC($\approx$50) & 50.6 & 56.0 & 58.5 & 54.4 \\
        & & & c@1($\downarrow$) & 51.7 & 50.8 & 53.4 & 48.7 \\
        \cmidrule(lr){3-8}
        & & \multirow{3}{*}{Utility} & NLI($\uparrow$) & 55.1 & 53.4 & 53.7 & 52.3 \\
        & & & CoLA($\uparrow$) & 94.7 & 94.9 & 94.9 & 94.8 \\
        & & & Sentiment($\uparrow$) & 83.0 & 78.6 & 81.0 & 79.6 \\
        \cmidrule(lr){2-8}

        & \multirow{7}{*}{8}
        & \multirow{2}{*}{Attribution} & MRR($\downarrow$) & 17.0 & 16.7 & 15.9 & 13.2 \\
        & & & R@8($\downarrow$) & 31.8 & 30.0 & 25.9 & 22.7 \\
        \cmidrule(lr){3-8}
        & & \multirow{2}{*}{Verification} & AUC($\approx$50) & 65.1 & 66.7 & 65.6 & 65.0 \\
        & & & c@1($\downarrow$) & 48.3 & 55.1 & 48.4 & 48.8 \\
        \cmidrule(lr){3-8}
        & & \multirow{3}{*}{Utility} & NLI($\uparrow$) & 65.9 & 64.9 & 65.1 & 64.8 \\
        & & & CoLA($\uparrow$) & 94.8 & 94.8 & 94.9 & 94.8 \\
        & & & Sentiment($\uparrow$) & 84.8 & 81.0 & 81.0 & 81.2 \\
        \cmidrule(lr){2-8}

        & \multirow{7}{*}{16}
        & \multirow{2}{*}{Attribution} & MRR($\downarrow$) & 39.8 & 37.9 & 35.4 & 35.0 \\
        & & & R@8($\downarrow$) & 60.7 & 58.3 & 55.7 & 53.4 \\
        \cmidrule(lr){3-8}
        & & \multirow{2}{*}{Verification} & AUC($\approx$50) & 66.7 & 67.6 & 67.0 & 65.5 \\
        & & & c@1($\downarrow$) & 49.8 & 50.2 & 47.8 & 48.4 \\
        \cmidrule(lr){3-8}
        & & \multirow{3}{*}{Utility} & NLI($\uparrow$) & 72.9 & 71.9 & 71.6 & 71.6 \\
        & & & CoLA($\uparrow$) & 94.7 & 94.8 & 94.8 & 94.8 \\
        & & & Sentiment($\uparrow$) & 85.3 & 81.6 & 81.9 & 81.6 \\
        \midrule
        \midrule

        \multirow{25}{*}{Blog}
        & \multirow{7}{*}{1}
        & \multirow{2}{*}{Attribution} & MRR($\downarrow$) & 1.7 & 2.2 & 2.5 & 2.2 \\
        & & & R@8($\downarrow$) & 2.0 & 3.2 & 3.2 & 2.4 \\
        \cmidrule(lr){3-8}
        & & \multirow{2}{*}{Verification} & AUC($\approx$50) & 48.3 & 50.4 & 50.8 & 51.5 \\
        & & & c@1($\downarrow$) & 48.6 & 51.6 & 51.8 & 49.5 \\
        \cmidrule(lr){3-8}
        & & \multirow{3}{*}{Utility} & NLI($\uparrow$) & 55.0 & 52.8 & 52.9 & 53.0 \\
        & & & CoLA($\uparrow$) & 93.6 & 94.2 & 94.0 & 94.4 \\
        & & & Sentiment($\uparrow$) & 83.6 & 84.6 & 86.0 & 87.2 \\
        \cmidrule(lr){2-8}

        & \multirow{7}{*}{8}
        & \multirow{2}{*}{Attribution} & MRR($\downarrow$) & 14.9 & 13.0 & 13.5 & 12.0 \\
        & & & R@8($\downarrow$) & 21.3 & 22.6 & 20.9 & 19.6 \\
        \cmidrule(lr){3-8}
        & & \multirow{2}{*}{Verification} & AUC($\approx$50) & 69.9 & 60.0 & 66.3 & 62.3 \\
        & & & c@1($\downarrow$) & 48.4 & 49.6 & 50.8 & 48.2 \\
        \cmidrule(lr){3-8}
        & & \multirow{3}{*}{Utility} & NLI($\uparrow$) & 68.9 & 67.9 & 67.5 & 67.4 \\
        & & & CoLA($\uparrow$) & 93.8 & 93.8 & 94.0 & 93.9 \\
        & & & Sentiment($\uparrow$) & 85.2 & 84.2 & 84.1 & 84.4 \\
        \cmidrule(lr){2-8}

        & \multirow{7}{*}{16}
        & \multirow{2}{*}{Attribution} & MRR($\downarrow$) & 41.1 & 34.3 & 37.4 & 35.2 \\
        & & & R@8($\downarrow$) & 61.4 & 51.1 & 53.3 & 54.2 \\
        \cmidrule(lr){3-8}
        & & \multirow{2}{*}{Verification} & AUC($\approx$50) & 71.2 & 70.7 & 70.0 & 70.6 \\
        & & & c@1($\downarrow$) & 53.2 & 51.0 & 49.6 & 50.3 \\
        \cmidrule(lr){3-8}
        & & \multirow{3}{*}{Utility} & NLI($\uparrow$) & 75.7 & 75.2 & 75.2 & 74.9 \\
        & & & CoLA($\uparrow$) & 93.8 & 93.9 & 94.0 & 94.0 \\
        & & & Sentiment($\uparrow$) & 85.3 & 84.6 & 85.3 & 84.8 \\

        \bottomrule
    \end{tabular}
    }
    \caption{Attribution risk, verification risk, and utility of AAST$_{\Phi}$ on Reddit and Blog under different privacy budgets. Results are reported for $K=1, 8, 16$.}
    \label{tab:dp-epsilon-results}
\end{table*}

\section{Qualitative Evaluation}
\label{appendix-qualitative-evaluation}

\subsection{Observed Strengths and Weaknesses}
\label{appendix-strengths-weaknesses}

Table~\ref{tab:qualitative-strengths-weaknesses} summarizes the main strengths and weaknesses observed from our quantitative results and qualitative examples. AAST is designed for aggregation-aware privacy and gives the strongest overall balance. KiP often preserves local meaning but can retain private wording and author signals. JAMDEC can move farther from the private text, but this often comes with semantic drift and higher generation cost.

\begin{table*}[t]
\centering
\small
\begin{tabular}{p{1.5cm}p{3.8cm}p{5.0cm}p{4.0cm}}
\toprule
\textbf{Method} & \textbf{Observed Strengths} & \textbf{Observed Weaknesses} & \textbf{Main Pattern} \\
\midrule

AAST &
Reduces aggregation linkability and preserves main events in qualitative examples. Does not require training generator. &
May still reduce exact semantic entailment, especially in cross-genre rewriting. Some private topics and events can remain because utility is preserved. &
Best privacy--utility balance under aggregation, with bundle-level selection reducing author signals across multiple released texts. \\
\midrule
KiP &
Often preserves local meaning and many details of private text. &
Shows weaker privacy under aggregation. Qualitative examples suggest that outputs can stay close to private wording and structure, and may contain fragmented words or control artifacts. Requires model adaptation through fine tuning. &
Strong meaning preservation, but higher authorship carryover and less stable output form. \\
\midrule
JAMDEC &
Can generate longer synthetic texts and does not require task specific generator fine tuning. &
Longer outputs often contain repetitive or weakly grounded content, with substantial semantic drift in qualitative examples. It also has much higher generation time due to over generation and filtering. &
Distance from the private text can help privacy, but often at the cost of meaning preservation and computational efficiency. \\

\bottomrule
\end{tabular}
\caption{Qualitative summary of observed strengths and weaknesses. The table summarizes patterns from the quantitative results and qualitative examples.}
\label{tab:qualitative-strengths-weaknesses}
\end{table*}

\subsection{Example Comparison}
\label{appendix-qualitative-example}

Table~\ref{tab:qualitative-1} and Table~\ref{tab:qualitative-2} provide qualitative examples on Reddit at $K$=2. The private bundle contains two query texts and two target texts from the same user. We bold key semantic elements in the private texts and the corresponding preserved or rewritten content in the synthetic outputs.

The examples show that AAST preserves the main meaning of the private texts while changing the surface form. For instance, both AAST variants retain the mother's Parkinson's diagnosis, the need for medical resources, the loss of faith, the fear of death, the work and tuition constraint, and the emotional harm caused by early return from a mission. These outputs do not copy the private texts directly, but they keep the central events, sentiment, and support seeking intent.

KiP keeps many details from the private texts, which explains why its utility can appear strong in a local semantic comparison. However, it often remains close to the original wording and structure, so it does not sufficiently reduce authorship signals under aggregation. This helps explain why KiP remains much closer to the private text in the privacy metrics. Its low CoLA score is also visible in the examples, where words are frequently split into unnatural fragments such as ``griev ing'', ``s an ity'', and ``in sensitive'', making the text less fluent.

JAMDEC shows the opposite failure pattern. It can reduce linkability because the generated text often drifts away from the original meaning. In several examples, the core private information is lost, such as the Parkinson's diagnosis, the religious grief, the tuition repayment constraint, and the mission related mental health context. This semantic drift may help privacy scores, but it weakens the utility of the synthetic text for downstream applications. Thus, the qualitative examples support the quantitative trend that AAST better balances useful meaning preservation with reduced aggregation based authorship risk.

\begin{table*}[t]
\centering
\fontsize{8.5}{7.5}\selectfont
\renewcommand{\arraystretch}{1.2}
\begin{tabular}{
>{\centering\arraybackslash}m{0.9cm} |
>{\centering\arraybackslash}m{0.8cm}
|
>{\arraybackslash}m{13.2cm}
}
\toprule
\textbf{Method} & \textbf{Bundle} & \multicolumn{1}{c}{\textbf{Text}} \\
\midrule

\multirow{25}{*}{Private}
& \makecell{Query\\ \\ID:1} & My \textbf{mom} just got diagnosed with \textbf{Parkinson's} this week and we are pretty \textbf{devastated}. She just turned \textbf{60}, and I didn't expect to have to confront \textbf{major health problems} so soon. We don't know much about \textbf{this disease} and are \textbf{scared and devastated}, to say the least. She asked me to help her find \textbf{resources} so that she can \textbf{understand her disease better}, as she's \textbf{not internet-savvy}. Can anyone please help me with this? Where can we read more about this disease? Are there any types of \textbf{support} that might be available to her? Anything I need to know about \textbf{how to support her}? \\
\cmidrule(lr){2-3}
& \makecell{Query\\ \\ID:2} & Have any of you gone through a \textbf{grieving} process after \textbf{leaving the church}? I have been a \textbf{jack Mormon} for many years, and just in the past couple of years have I finally fully \textbf{lost my faith} and left completely. I was already living the "worldly" life, so there hasn't been much to gain since I left, but there has been a lot to lose. I am \textbf{terrified of death} now, I feel much more \textbf{pessimistic} about the future and cynical about life, and sometimes I miss that feeling of being part of a community. To make matters worse, my \textbf{mom} was just diagnosed with a \textbf{degenerative disease} that will probably \textbf{kill her}. I find myself actually wishing that I had some \textbf{belief system} and community {support} to sustain me through the \textbf{shock and grief} I feel from her diagnosis, but I don't know that I can ever believe in any religion again. I am \textbf{grieving for my mom}, and grieving for all I feel that I lost when I lost my faith. I feel like I will be \textbf{crushed under the weight} of all of this grief. \\
\cmidrule(lr){2-3}
& \makecell{Target\\ \\ID:1} & I'm a \textbf{student taking evening classes} and have also been \textbf{working full time} to pay the bills. My job is getting steadily more \textbf{stressful and demanding}, and I'm struggling to \textbf{mentally cope} and to perform my duties. But I took some \textbf{tuition assistance from my employer} to pay for school, and their stipulation is that I have to \textbf{stay at this job} until at least Jan of 2021 or I have to \textbf{pay back all the money} (which I can't do). I am losing my mind because of how much I \textbf{hate this job}, and I need some advice on how to \textbf{keep my sanity}. What keeps you sane when you're \textbf{stuck at a job you hate but can't afford to leave}?  \\
\cmidrule(lr){2-3}
& \makecell{Target\\ \\ID:2} & I just need to \textbf{vent}. I have a friend who has to \textbf{come home early from her mission} due to \textbf{mental health issues}. I just had dinner with her the other day and was saddened to see how much she's struggling with \textbf{feelings of failure} from being home, compounded by the incredibly \textbf{insensitive comments} some members have made. It was so hard for me to keep my mouth shut while I was with her, because I was just boiling inside with \textbf{rage at the church} for making young folks feel this way. I had to go home early from my mission also, and although I returned eventually I never fully got over the guilt until I left the church. Missions are \textbf{emotionally} (and sometimes physically) \textbf{traumatic} for many people, and the church does NOT take care of those kids the way it should protect its \textbf{UNPAID VOLUNTEERS}. I could go on and on about everything that is fucked up about the way \textbf{missions are run}.  \\
\cmidrule(lr){1-3}

\multirow{19}{*}{AAST$_{\mathrm{M}}$}
& \makecell{Query\\ \\ID:1} & The news that my \textbf{mother} was recently diagnosed with \textbf{Parkinson's} has left our family \textbf{deeply shaken}. She was looking forward to enjoying her \textbf{retirement}, and this unexpected diagnosis has been incredibly difficult to process. The reality of facing such a \textbf{serious health challenge} so soon is overwhelming. We're still learning about the disease and feel a mix of \textbf{fear and uncertainty} about what lies ahead. My mom, who \textbf{isn't comfortable navigating the internet}, has asked for my help in finding \textbf{reliable information} so she can better understand her condition. If anyone has recommendations for \textbf{trustworthy resources}, I'd be so grateful for your guidance. Where should we turn for \textbf{accurate and helpful information}? \\
\cmidrule(lr){2-3}
& \makecell{Query\\ \\ID:2} & The weight of my \textbf{grief} is twofold: I mourn the absence of my \textbf{mother}, and I mourn the dissolution of the \textbf{beliefs} that once shaped my world. For years, I identified as a \textbf{jack Mormon}, but only recently did I fully sever my ties to the \textbf{faith}. The transition hasn't been a revelation or a release; it's been a quiet, aching \textbf{loss}. I was already living outside the church expectations, so the departure didn't bring new freedoms, only a deepening sense of \textbf{emptiness}. The absence of \textbf{faith} hasn't lifted me; it's left me \textbf{adrift}. \\
\cmidrule(lr){2-3}
& \makecell{Target\\ \\ID:1} & This job is driving me absolutely crazy, and I'm desperate for ways to \textbf{maintain my mental well-being}. How do you manage to stay grounded when \textbf{trapped in a job you despise but can't walk away from}? The situation is even more frustrating because I accepted \textbf{tuition assistance from my employer}, which comes with a catch. I'm obligated to \textbf{remain here} until I've fully \textbf{repaid the money}. \\
\cmidrule(lr){2-3}
& \makecell{Target\\ \\ID:2} & I feel the need to \textbf{express my frustration}. A close friend of mine is \textbf{returning home prematurely from her mission} because of \textbf{mental health challenges}. We recently shared a meal together, and it broke my heart to witness how deeply she's grappling with a \textbf{sense of inadequacy} over her early return. The situation is made even worse by the \textbf{thoughtless remarks} she's had to endure from others, which only pile onto her distress. \\
\cmidrule(lr){1-3}

\multirow{15}{*}{AAST$_{\Phi}$}
& \makecell{Query\\ \\ID:1} & Just last week, my \textbf{mom} found out she has \textbf{Parkinson's disease}. Dad finds technology a real maze, so we're \textbf{in the dark} about what this all means. I need to know right away \textbf{how to help mom} figure out her \textbf{next steps}. Feeling \textbf{in the dark and scared} is tough. Does anyone know of \textbf{specialists or services} that can help her see what she can do? Plus, any \textbf{advice} to make this journey smoother would be amazing. Thanks a ton for any \textbf{help} you can provide. \\
\cmidrule(lr){2-3}
& \makecell{Query\\ \\ID:2} & For years, I considered myself a \textbf{jack Mormon}, but recently, my \textbf{faith has faded} significantly, prompting my final decision to \textbf{leave}. This change has instilled in me an overwhelming \textbf{fear of death}, casting a shadow over my view of the \textbf{future}, filled with life's \textbf{cynicism}. Sometimes, I miss the sense of \textbf{community and connection} I once cherished. The thought of embracing any \textbf{religion} again appears unlikely. \\
\cmidrule(lr){2-3}
& \makecell{Target\\ \\ID:1} & In the whirlwind of daily life, juggling a \textbf{full-time education with evening classes} and a \textbf{challenging job} is an immense undertaking. The weight of my \textbf{workplace duties} has been steadily climbing, causing a \textbf{mental tug-of-war} as I strive to meet my commitments. In times of such difficulty, how does one stay rooted and \textbf{preserve their mental health} in a \textbf{job they find utterly detestable but from which there is no escape}? How do you navigate these turbulent waters? \\
\cmidrule(lr){2-3}
& \makecell{Target\\ \\ID:2} & It's quite disheartening that my friend had to \textbf{cut her mission short} due to \textbf{mental health issues}. Witnessing her struggle with \textbf{guilt} over something she wasn't responsible for is distressing. The way \textbf{missions are currently managed} is seriously lacking and warrants significant criticism. \\

\bottomrule
\end{tabular}
\caption{Qualitative examples on Reddit at $K$=2. This part shows the private bundle and the AAST outputs. Bold text marks key semantic elements in the private texts and corresponding preserved content in the synthetic texts.}
\label{tab:qualitative-1}
\end{table*}

\begin{table*}[t]
\centering
\fontsize{8.5}{6.5}\selectfont
\renewcommand{\arraystretch}{1.2}
\begin{tabular}{
>{\centering\arraybackslash}m{1.1cm} |
>{\centering\arraybackslash}m{0.8cm}
|
>{\arraybackslash}m{13cm}
}
\toprule
\textbf{Method} & \textbf{Bundle} & \multicolumn{1}{c}{\textbf{Text}} \\
\midrule

\multirow{25}{*}{KiP}
& \makecell{Query\\ \\ID:1} &  My \textbf{mother} was just diagnosed with p y lor i this week, and we are pretty \textbf{devastated}. She just turned \textbf{60}, and I didn't expect to have to confront \textbf{major health problems} so soon. We don't know much about \textbf{this disease} and are, to say the least, \textbf{scared and devastated}. She asked me to help her find \textbf{resources} so that she might better understand her disease, as she is \textbf{not web-savvy}. Can anyone please help me with this?. Where can we read more about this disease?. A re there any kinds of \textbf{support} that might be available to her?. Anything I need to know about \textbf{how to support her}?  \\
\cmidrule(lr){2-3}
& \makecell{Query\\ \\ID:2} & H a ve any of you gone through a \textbf{griev ing process} after \textbf{leaving the church}?. I have been a \textbf{jack-mor mon} for many years, and only in the last couple of years have I completely \textbf{lost my faith} and left completely. I was already living a life of the world ly life. So there has n't been much to gain since I left, but there has been a lot to lose. I am now more \textbf{f right ened of death}; I feel much more \textbf{pe s s im istic about the future} and \textbf{c y n ical about life}; and sometimes I miss that feeling of being part of a community. To make matters worse, my \textbf{mother} was just diagnosed with a \textbf{de generative disease} that will probably \textbf{kill her}. And really, I find myself wanting that I had some \textbf{belief system} and community support to sustain me through the \textbf{shock and grief} I feel from her diagnosis, but I don't know that I can ever again believe in any \textbf{religion}. I am \textbf{griev ing for my mother}, and griev ing for all that I feel that I have lost when I have \textbf{lost my faith}. I feel as if I shall be \textbf{crushed under the weight} of all these \textbf{grief s}  \\
\cmidrule(lr){2-3}
& \makecell{Target\\ \\ID:1} & I am a \textbf{student taking evening classes} and have been \textbf{working full time} to pay the bills. lexic al = 60. My job is steadi ly more \textbf{difficult and more demanding}; and I am struggling to \textbf{cope mentally} and perform my duties. But I took some \textbf{tuition assistance from my employer} to pay for school, and their s tipul ation is that I have to \textbf{stay at this job} until at least Jan of 20 21, or I must \textbf{pay back all the money}, which I cannot do. I am losing my mind because of how much I \textbf{hate this job}. I need some advice on how to \textbf{keep my s an ity}. What keeps you s an e when you are \textbf{stuck at a job you hate but can't afford to leave}? \\
\cmidrule(lr){2-3}
& \makecell{Target\\ \\ID:2} & I just need to \textbf{vent}. lexic al = 60. And order = 60. I have a friend who has to \textbf{come home early from her mission} because of \textbf{mental health problems}. I just had dinner with her the other day and was sad d ened to see how much she was struggling with \textbf{feelings of failure} from being at home, compound e d by the incredibly \textbf{in sensitive comments} some members made. I t was so hard for me to keep my mouth shut while I was with her, because I was boiling inside with \textbf{rage at the church} for making young people feel this way. ...(omitted) \\
\cmidrule(lr){1-3}

\multirow{21}{*}{JAMDEC}
& \makecell{Query\\ \\ID:1} & We are trying to figure out what we are going to do with the money, but we don't know if we will be able to \textbf{take care of her} for the rest of our lives or if she will need to be in a nursing home for the remainder of her life, so we are just trying to get as much money as we can to \textbf{help her out} as much as possible while we are still able to do so! Thank you so much for taking the time to read our story and if you are able to \textbf{help us out}, we would love to hear from you so we can thank you for your kindness! ! ...(omitted) \\
\cmidrule(lr){2-3}
& \makecell{Query\\ \\ID:2} & If so, what was it like for you, and what advice would you give to someone who is going through the same thing right now? Please share your thoughts and experiences in the comments section below, and if you'd like to learn more about this topic, check out my book, Becoming a Supporter (affiliate link). If so, what was it like for you, and what advice would you give to someone who is going through the same thing right now? I was raised in the \textbf{church}, went to seminary, and served a mission, but after a few years I realized that the \textbf{church} was full of lies and that I had been lied to for so long, that I just couldn't take it anymore. ...(omitted) \\
\cmidrule(lr){2-3}
& \makecell{Target\\ \\ID:1} & I'm not sure what I'm going to do in the future, but I don't know what I want to do with my life, or if I even want to have a future in the first place, so I'm trying to figure out what to do and how to do it while I still have the ability to do so, and while I'm still able to \textbf{pay my bills} and keep up with my obligations to my family, my friends, and the people who care about me, and who have supported me in the past and will continue to support me for as long as I'm able to continue to live and work in this country I don't want to do that, so I'm looking for a job that will allow me to work from home, ...(omitted) \\
\cmidrule(lr){2-3}
& \makecell{Target\\ \\ID:2} & I don't know what I'm going to do with my life, but I know I'm not going to be able to do what I want to do for the rest of my life if I don't do something about it now." That's what I've been trying to say to myself for the past few years, but it's hard to say it out loud when you're in a relationship with someone who doesn't seem to understand what you're going through, or if they do, they don't understand it as well as you do, so you end up saying it to yourself in a way that makes you feel like She has been in and out of the hospital for the past year and a half due to her \textbf{mental illness}, and I have been trying to help her get the help she needs, but she has been unable to do so due to the fact that she is a member of the \textbf{Church} of Jesus Christ of Latter Day Saints. ...(omitted) \\

\bottomrule
\end{tabular}
\caption{Qualitative examples on Reddit at $K$=2, continued. This part shows KiP and JAMDEC outputs for the same private bundle in Table~\ref{tab:qualitative-1}. Bold text marks retained content related to the private text.}
\label{tab:qualitative-2}
\end{table*}

\section{Generation Cost}
\label{appendix-generation-cost}

\subsection{Computational Efficiency Results}
\label{appendix-computational-results}

We evaluate computational cost by measuring the time required to generate synthetic text for 250 Blog users across $K$=1 to $K$=16. AAST uses Mistral-Small as the generator. For each method, synthetic text generation was distributed across the same three GPUs, including two NVIDIA A6000 GPUs with 48GB VRAM and one NVIDIA A100 GPU with 40GB VRAM. Because JAMDEC requires substantially longer generation time, reaching several days at larger $K$, we report runtime relative to AAST instead of only listing raw GPU hours.

Figure~\ref{fig:time_cost} shows that AAST is consistently the fastest method across all bundle sizes. KiP requires about $1.5\times$ to $2.3\times$ the runtime of AAST. JAMDEC is much more expensive, requiring about $18.5\times$ the runtime of AAST on average. At $K$=16, AAST completes generation in 5.8 hours, while KiP takes 11.3 hours and JAMDEC takes about 120 hours. These results show that AAST is more practical for downstream applications that require large scale synthetic text generation, as it scales efficiently with aggregation size while preserving the generation quality needed for privacy and utility evaluation.

We also report output length to make the runtime comparison easier to interpret. In the Reddit $K$=1 setting, each method generates 500 synthetic texts. The average private text length is 159.1 words, while the average synthetic lengths for AAST, JAMDEC, and KiP are 106.8, 118.5, and 139.2 words, respectively. KiP outputs are about \(1.3\times\) longer than AAST outputs, and JAMDEC outputs are about \(1.1\times\) longer. These length differences are much smaller than the runtime gaps, where KiP is about \(1.9\times\) slower than AAST and JAMDEC is about \(18.5\times\) slower. This suggests that AAST's efficiency comes mainly from its simpler generation pipeline. The AAST output length reflects the prompt design in Appendix~\ref{appendix-prompt-design}, which prioritizes preserving core meaning.

\subsection{Token Cost Analysis}
\label{appendix-token-cost}

Table~\ref{tab:token-usage-blog} reports the token usage of AAST$_{\mathrm{M}}$ on Blog. Token usage grows almost linearly with bundle size because larger $K$ means more private texts must be rewritten. At $K$=16, the setting contains 250 users with two bundles per user and 16 texts per bundle, resulting in $8{,}000$ private texts to generate. This produces about 7.97 million input and output tokens.

This cost should be interpreted in light of the corpus scale. The large token count at higher $K$ comes from generating a full synthetic corpus for thousands of private texts. Thus, the token cost mainly reflects the number of texts that must be rewritten under the aggregation setting.

\begin{table*}[t]
    \centering
    {
    \fontsize{7.5}{8.5}\selectfont
    \renewcommand{\arraystretch}{1.2}
    \begin{tabular}{
    >{\raggedright\arraybackslash}p{3.9cm}
    >{\raggedright\arraybackslash}p{3.9cm}
    >{\raggedright\arraybackslash}p{3.9cm}}
        \toprule
        \multicolumn{3}{l}{\textbf{Subreddits}} \\
        \midrule
        frugal & povertyfinance & poverty \\
        poor & assistance & homeless \\
        almosthomeless & tenanthelp & section8publichousing \\
        unemployment & foodstamps & povertykitchen \\
        debt & bankruptcy & borrow \\
        studentloans & medicaid & \\
        \midrule
        \multicolumn{3}{l}{\textbf{Filtering Keywords}} \\
        \midrule
        low income & financial burden & poverty \\
        poor & unemployment & underemployment \\
        wage stagnation & low wage & minimum wage \\
        underpaid & low pay & financial hardship \\
        money struggles & financially strained & financial stress \\
        affordability issues & living in poverty & living below the poverty line \\
        working poor & underprivileged & financially disadvantaged \\
        impoverished & economic inequality & trapped in a cycle of poverty \\
        living in low-income housing & experiencing financial instability & living with no savings \\
        completely broke & barely making ends meet & struggling to make ends meet \\
        paycheck to paycheck & living paycheck to paycheck & barely scraping by \\
        barely surviving & cutting back on basic necessities & can't afford \\
        cant afford & cannot afford & can't pay \\
        cant pay & cannot pay & can't pay rent \\
        cant pay rent & cannot pay rent & behind on rent \\
        late on rent & rent overdue & rent arrears \\
        eviction & eviction notice & getting evicted \\
        facing eviction & homeless & almost homeless \\
        about to be homeless & living in my car & sleeping in my car \\
        couch surfing & shutoff notice & utilities shut off \\
        power shut off & water shut off & gas shut off \\
        food insecurity & food bank & food pantry \\
        food stamps & snap & ebt \\
        lost my job & laid off & hours cut \\
        reduced hours & collections & in collections \\
        debt collector & payday loan & title loan \\
        bankrupt & bankruptcy & chapter 7 \\
        chapter 13 & medical debt & no health insurance \\
        section 8 & housing voucher & public housing \\
        medicaid & broke & behind on bills \\
        \bottomrule
    \end{tabular}
    \caption{Subreddits and filtering keywords used to construct the Reddit dataset.}
    \label{tab:reddit_sub_key}
    }
\end{table*}